\documentclass[lettersize,journal]{IEEEtran}
\usepackage{bm}
\usepackage[pdftex]{graphicx}
\usepackage{subfigure}
\usepackage{booktabs}
\usepackage{multicol}
\usepackage{multirow}
\usepackage{newfloat}
\usepackage{tabularx}
\usepackage[nocompress]{cite}
\usepackage{array}
\usepackage{amsmath,amssymb,amsfonts}
\usepackage{titlesec}
\usepackage{amsfonts}
\usepackage{balance}
\usepackage[colorlinks=true, citecolor=black, linkcolor=black, urlcolor=black]{hyperref}
\newcommand{\eg}{e.g.}

\begin{document}
\title{Deep Attentional Guided Image Filtering}
\author{Zhiwei~Zhong,
Xianming~Liu,~\IEEEmembership{Member,~IEEE,}
Junjun~Jiang,~\IEEEmembership{Member,~IEEE,}
Debin Zhao,~\IEEEmembership{Member,~IEEE,}
Xiangyang Ji,~\IEEEmembership{Member,~IEEE}

\IEEEcompsocitemizethanks{
\IEEEcompsocthanksitem Z. Zhong, X. Liu, J. Jiang and D. Zhao are with the School of Computer Science and Technology, Harbin Institute of Technology, Harbin 150001, China, and also with Peng Cheng Laboratory, Shenzhen 518052, China  E-mail: \{zhwzhong,csxm,jiangjunjun\}@hit.edu.cn.
\IEEEcompsocthanksitem X. Ji is with the Department of Automation, Tsinghua University, Beijing 100084, China.  E-mail: xyji@tsinghua.edu.cn.}
\thanks{}}

\markboth{Journal of \LaTeX\ Class Files,~Vol.~18, No.~9, September~2020}%
{How to Use the IEEEtran \LaTeX \ Templates}

\maketitle

\begin{abstract}
Guided filter is a fundamental tool in computer vision and computer graphics which aims to transfer structure information from guidance image to target image. Most existing methods construct filter kernels from the guidance itself without considering the mutual dependency between the guidance and target. However, since there typically exist significantly different edges in two images, simply transferring all structural information of the guidance to the target would result in various artifacts. To cope with this problem, we propose an effective framework named deep attentional guided image filtering, the filtering process of which can fully integrate the complementary information contained in both images. Specifically, we propose an attentional kernel learning module to generate dual sets of filter kernels from the guidance and the target, respectively, and then adaptively combine them by modeling the pixel-wise dependency between the two images. Meanwhile, we propose a multi-scale guided image filtering module to progressively generate the filtering result with the constructed kernels in a coarse-to-fine manner. Correspondingly, a multi-scale fusion strategy is introduced to reuse the intermediate results in the coarse-to-fine process. Extensive experiments show that the proposed framework compares favorably with the state-of-the-art methods in a wide range of guided image filtering applications, such as guided super-resolution, cross-modality restoration, texture removal, and semantic segmentation. Moreover, our scheme achieved the first place in real depth map super-resolution challenge held in ACM ICMR'2021\footnote{https://icmr21-realdsr-challenge.github.io/\#Leaderboard}.
\end{abstract}


\begin{figure*}[!tb]
	\begin{center}
		\subfigure[Guidance Image]{
		\begin{minipage}[b]{0.19\linewidth}
			\includegraphics[width=1\linewidth]{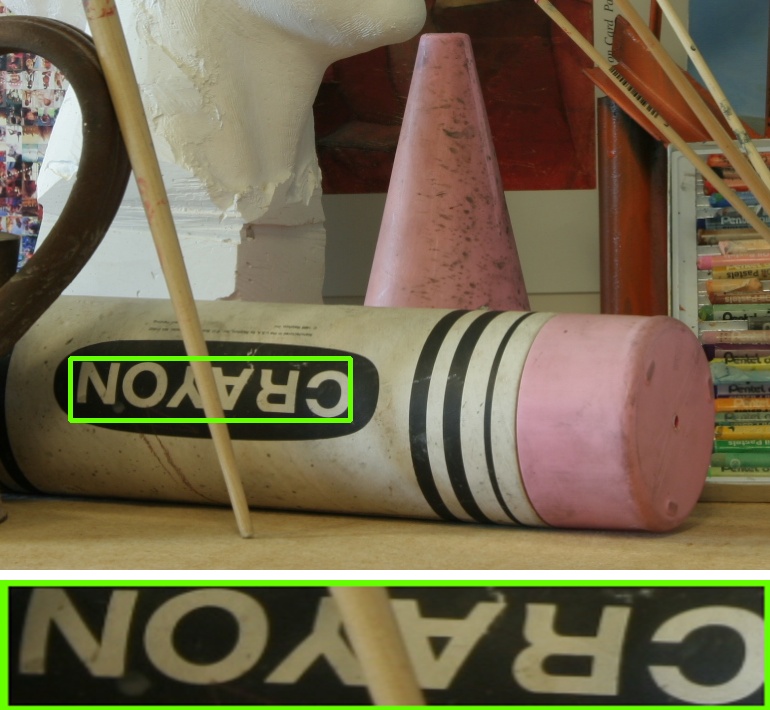} 
		\end{minipage}
		%\label{fig:grid_4figs_1cap_4subcap_1}
	}\subfigure[GF]{
    		\begin{minipage}[b]{0.19\linewidth}
  		 	\includegraphics[width=1\linewidth]{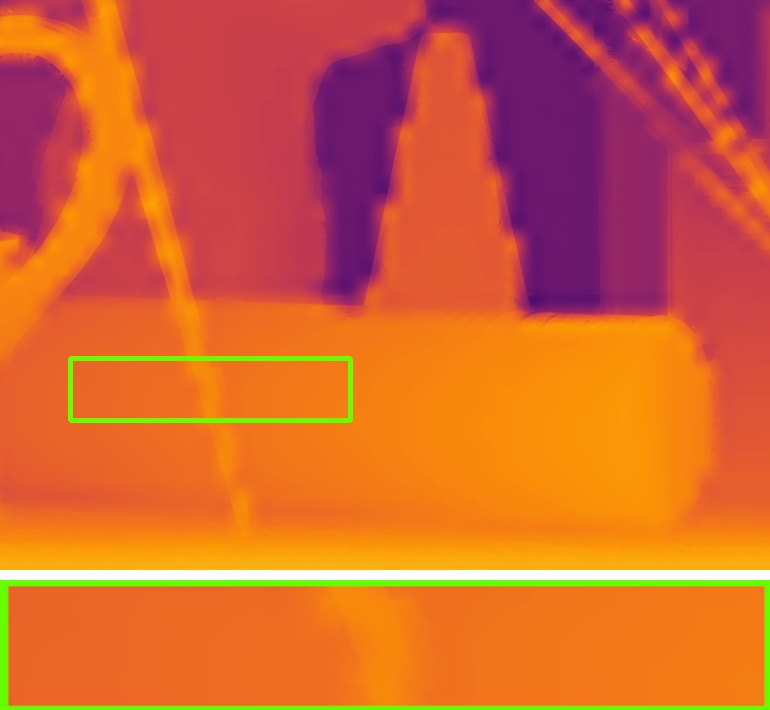}
    		\end{minipage}
		%\label{fig:grid_4figs_1cap_4subcap_2}
    	}\subfigure[JBU]{
		\begin{minipage}[b]{0.19\linewidth}
			\includegraphics[width=1\linewidth]{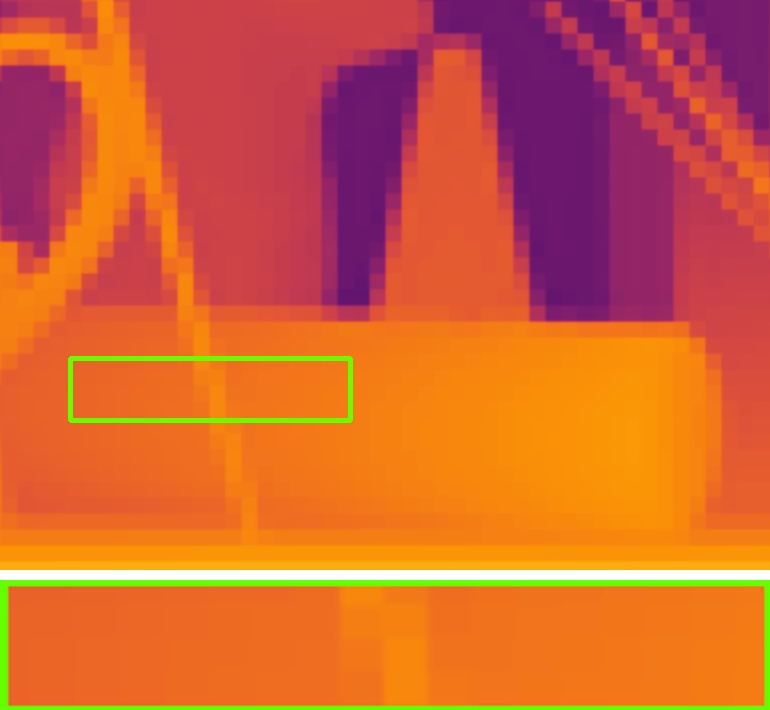} 
		\end{minipage}
		%\label{fig:grid_4figs_1cap_4subcap_3}
	}\subfigure[MuGF]{
    		\begin{minipage}[b]{0.19\linewidth}
  		 	\includegraphics[width=1\linewidth]{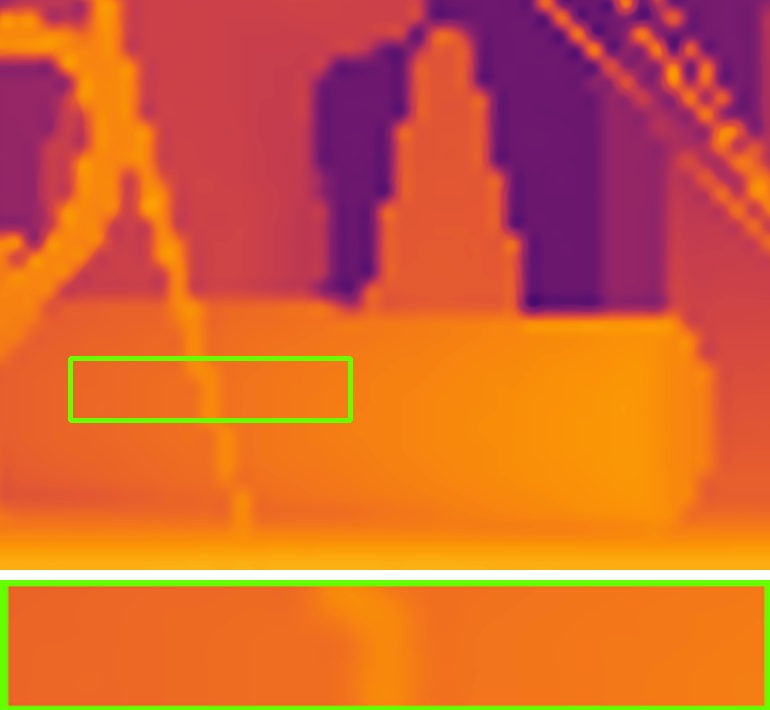}
    		\end{minipage}
		%\label{fig:grid_4figs_1cap_4subcap_2}
    	}\subfigure[CUNet]{
		\begin{minipage}[b]{0.19\linewidth}
			\includegraphics[width=1\linewidth]{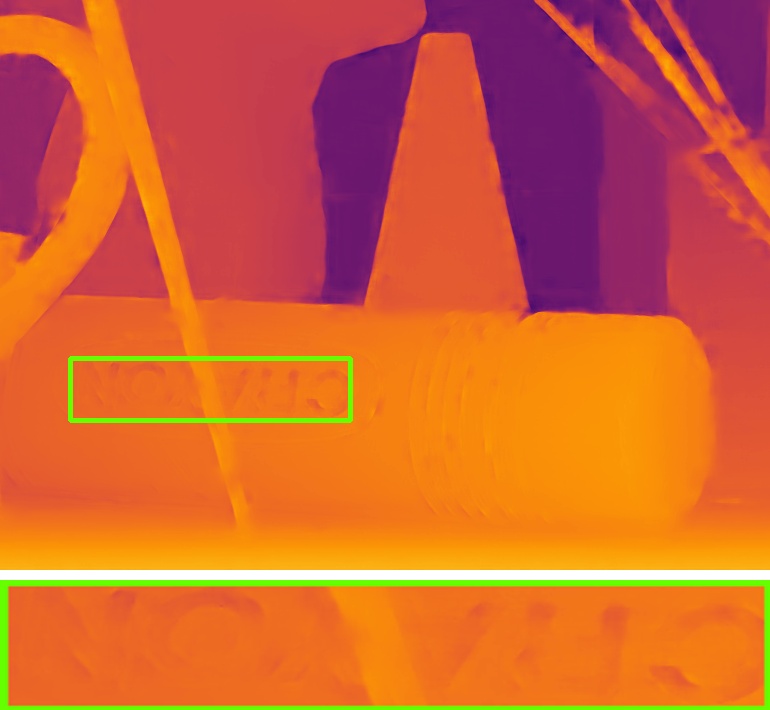} 
		\end{minipage}
		%\label{fig:grid_4figs_1cap_4subcap_3}
	}
	\subfigure[DJFR]{
		\begin{minipage}[b]{0.19\linewidth}
			\includegraphics[width=1\linewidth]{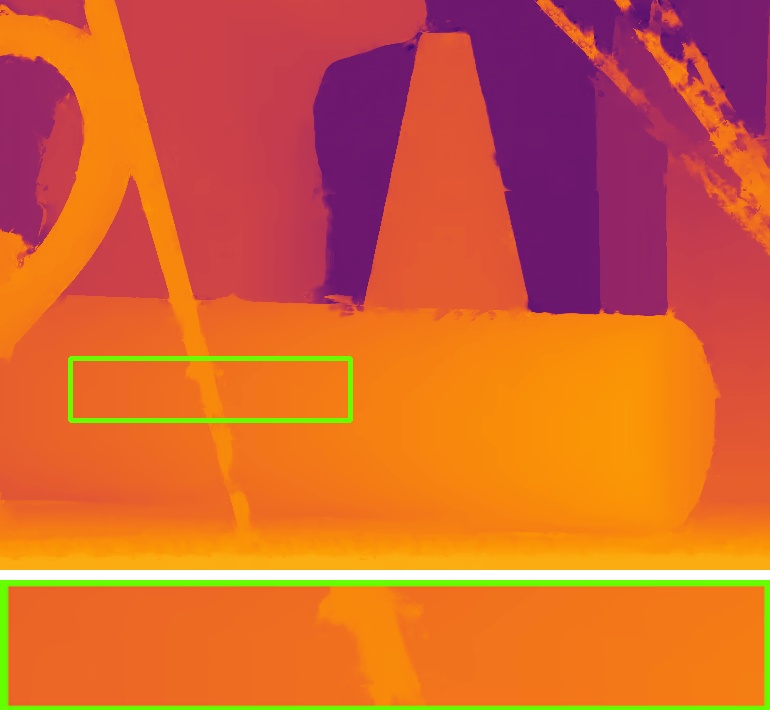} 
		\end{minipage}
		%\label{fig:grid_4figs_1cap_4subcap_3}
	}\subfigure[PAC]{
		\begin{minipage}[b]{0.19\linewidth}
			\includegraphics[width=1\linewidth]{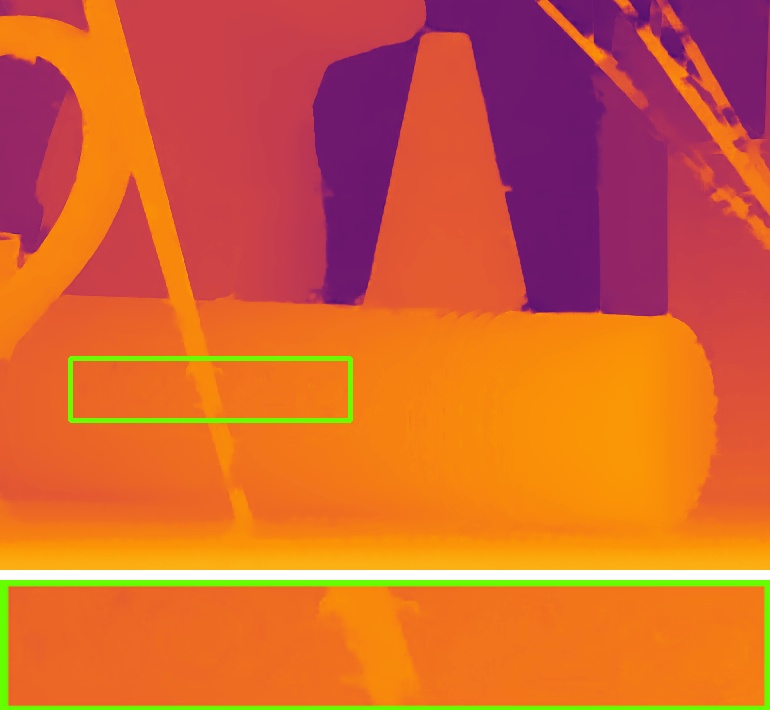} 
		\end{minipage}
		%\label{fig:grid_4figs_1cap_4subcap_3}
	}\subfigure[DKN]{
    		\begin{minipage}[b]{0.19\linewidth}
		 	\includegraphics[width=1\linewidth]{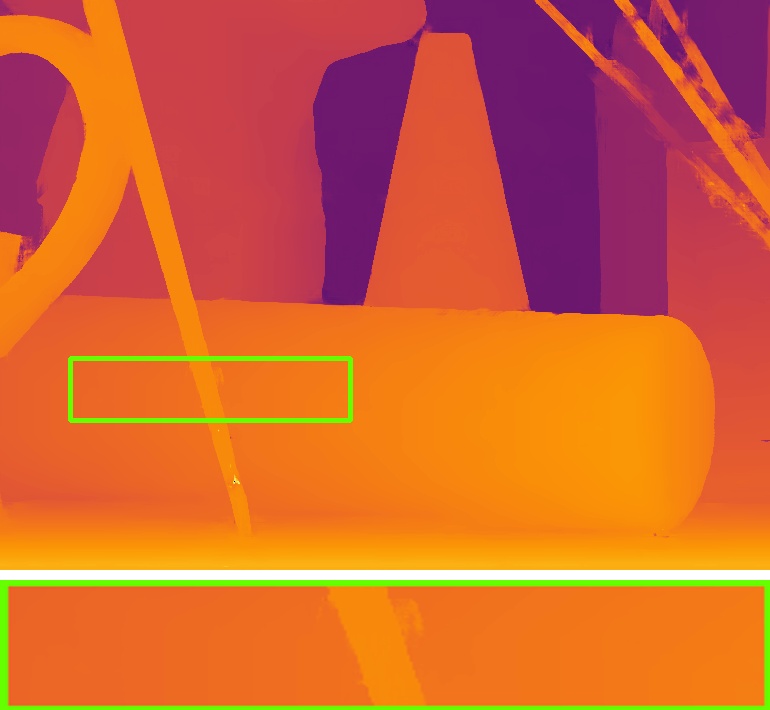}
    		\end{minipage}
		%\label{fig:grid_4figs_1cap_4subcap_4}
    	}\subfigure[Ours]{
    		\begin{minipage}[b]{0.19\linewidth}
  		 	\includegraphics[width=1\linewidth]{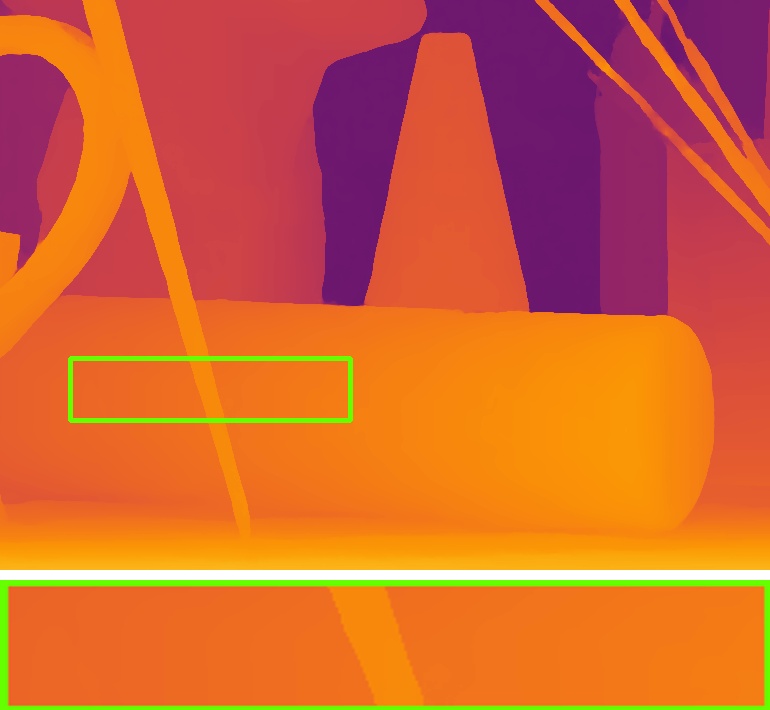}
    		\end{minipage}
		%\label{fig:grid_4figs_1cap_4subcap_2}
    	}\subfigure[Ground Truth]{
    		\begin{minipage}[b]{0.19\linewidth}
  		 	\includegraphics[width=1\linewidth]{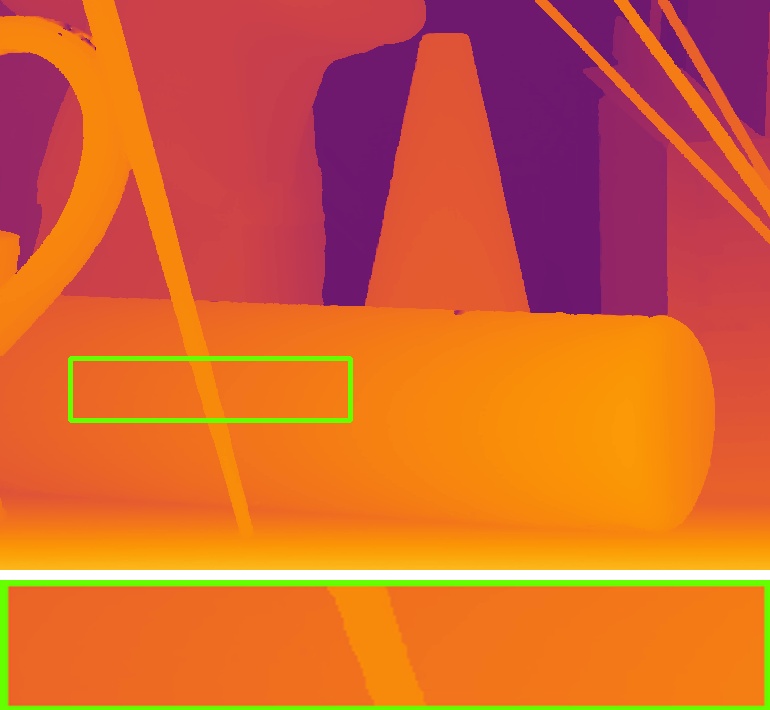}
    		\end{minipage}
		%\label{fig:grid_4figs_1cap_4subcap_2}
    	}
	\end{center}
% 	\vspace{-0.1in}
	\caption{Guided image filtering on a RGB/D image pair for 16 $\times$ guided super-resolution: (a) Guidance image, (b) GF~\cite{GF}, (c) JBU~\cite{JBU}, (d) MuGF~\cite{mugf}, (e) CUNet~\cite{CUNet}, (f) DJFR~\cite{DJFR}, (g) PAC~\cite{PanNet}, (h) DKN~\cite{DKN}, (i) Ours, (j) Ground truth.  The results of (b)-(c) suffer from edge blurring artifact and the results of (d)-(h) suffer from texture-copying artifacts. Our result produces much sharper edges. Please enlarge the PDF for more details.}
	\label{fig:fm}
\end{figure*}

\section{Introduction}

\IEEEPARstart{G}{uided} filter (GF), also named joint filter, is tailored to transfer structural information from a guidance image to a target one. The popularity of GF can be attributed to its ability in handling visual signals in various domains and modalities, where one modal signal serves as the guidance to improve the quality of the other one. It has been a useful tool for many image processing and computer vision tasks, such as depth map super-resolution~\cite{JBU, DJFR}, scale-space filtering~\cite{sdf, RGF}, cross-modality image restoration~\cite{GF, mugf, guide-filter-SR2019}, structure-texture separation~\cite{texture2012, texture2013, RTV}, image semantic segmentation~\cite{FBS, DKN} and so on.

In the literature, GF has been extensively studied, ranging from the classical bilateral filter to the emerging deep learning-based ones. The pioneer bilateral filter \cite{bilateral} constructs spatially-varying kernels, where local image structures of the guidance image are explicitly involved into filtering process through the photometric similarity. The guided image filtering scheme proposed by He et al.~\cite{GF} takes a more rigorous manner to exploit the structure information of the guidance, which computes a locally linear model over the guidance image for filtering. These filters consider only the information contained in the guidance image in filtering. However, since there typically exist significantly different edges in the two images, simply transferring all patterns of the guidance to the target would introduce various artifacts. Some works~\cite{sdf, mugf} propose to utilize the optimization-based manner to find mutual structures for propagation while suppressing inconsistent ones. However, it is challenging to select reference structures and propagate them properly by hand-crafted objective functions. In addition, the computational complexity of these methods is usually high.

In recent years, learning-based approaches for GF design are becoming increasingly popular, which derive GF in a purely data-driven manner. They allow the networks to learn how to adaptively select structures to transfer, and thus have the ability to handle more complicated scenarios. For instance, in~\cite{DFN}, a dynamic filter network (DFN) is proposed where pixel-wise filters are generated dynamically using a separate sub-network conditioned on the guidance. Unlike DFN,  Su et al.~\cite{PanNet} adapts a standard spatially invariant kernel at each pixel by multiplying it with a spatially varying filter. Although with increased flexibility thanks to their adaptive nature, ~\cite{DFN} and \cite{PanNet} still suffer from the same drawback as~\cite{GF, bilateral} that only the guidance information is considered in filters design. Some recent methods attempt to exploit the target and guidance information jointly. For instance, Li et al.~\cite{DJFR} propose to leverage two sub-networks to extract informative features from both the target and guidance images, which are then concatenated as inputs for the fusion network to selectively transfer salient structures from the guidance to the target. Instead of regressing the filtering results directly from the network, Kim et al.~\cite{DKN} proposes to use spatially variant weighted averages, where the set of neighbors and the corresponding kernel weights are learned in an end-to-end manner. However, in the designed networks of these methods,  the simple concatenation or element-wise multiplication is exploited to combine multi-modal information, which is not that effective. There is no mechanism to distinguish the contributions of the guidance and the target to the final filtering result, and thus would also lead to erroneous structure propagation. In addition, the guidance and target images are treated as independent information since existing methods typically utilize two separate networks for feature extraction, thus the complementary information contained in the two images cannot be fully exploited.

By reviewing existing GF methods, it can be found that most of them concentrate their efforts on how to transfer structural information from the guidance to the target. However, for some scenarios, such as cross-modality image restoration~\cite{CUNet} and guided super-resolution~\cite{GSPRT}, multi-modal data has significantly different characteristics due to the difference of sensing principle, making the guidance not always trustworthy. In view of this, we argue that the purpose of GF should be two-folds: 1) apply the guidance as a prior for reconstruction of regions in the target where there are structure-consistent contents; and 2) derive a plausible prediction for regions in the target with inconsistent contents of the guidance. The latter represents the case that the guidance is no longer reliable, so we have to rely on the target itself for reconstruction. Most existing GF methods only concern structure transferring from the guidance, but neglect structure prediction from the target, leading to erroneous or extraneous artifacts in the output. It implies that instead of performing regression on guidance only, as done in ~\cite{GF, bilateral}, we should perform \textit{dual regression} on both the guidance and the target, and combine them adaptively in a smarter manner instead of simple concatenation or element-wise multiplication, as done in ~\cite{DJFR, DKN}. ``Dual regression" and ``smart combination" bring the main motivations of our proposed method. 

Accordingly, in this paper, we propose an effective deep attentional guided image filtering scheme, which constructs filter kernels by fully considering information from both guidance and target images. Specifically, an attentional kernel learning module is proposed to generate dual sets of filter kernels from the guidance and the target, respectively.  Moreover, pixel-wise contributions of the guidance and the target to the final filtering result are automatically learned. In this way, we can adaptively apply the guidance as a prior for reconstruction of target regions where there are structure-consistent contents with the guidance; and derive a prediction for target regions with inconsistent contents by regression on the target itself. We show an illustrated example in Fig.~\ref{fig:fm}, which presents the visual filtering results comparison of our scheme with the state-of-the-art guided depth super-resolution methods.
It can be found that our proposed method is capable of producing high-resolution depth image with clear boundaries as well as avoiding texture-copying artifacts. %Fig.~\ref{fig:fm} (a) and Fig.~\ref{fig:fm} (j) are guidance image and ground truth depth map respectively. Fig.~\ref{fig:fm} (b) - Fig.~\ref{fig:fm} (h) are the depth maps generated by other methods. Fig.~\ref{fig:fm} demonstrates that our proposed method is capable of producing depth images with clear boundaries as well as avoiding texture-copying artifacts.%Experimental results on various guided image filtering applications demonstrate the effectiveness of the proposed method.

The main contributions of the proposed method are summarized as follows:

\begin{itemize}
    \item We propose an attentional kernel learning (AKL) module for guided image filtering, which generates dual sets of filter kernels from both guidance and target, and then adaptively combines these kernels by modeling the pixel-wise dependencies between the two images in a learning manner. Compared with existing kernel generation approaches, the proposed method is more robust when there are inconsistent structures
  between the guidance and the target.
    \item We propose a multi-scale guided filtering module, which generates the filtering result in a coarse-to-fine manner. Correspondingly, we propose a multi-scale fusion strategy with deep supervision to fully explore the intermediate results in the coarse-to-fine process. To the best of our knowledge, this is the first guided filter framework that learns the multi-scale kernels to filter the target image at different scales in the embedding space.
    \item We evaluate the performance of the proposed method on various computational photography and computer vision tasks, such as guided image super-resolution, cross-modality image restoration, texture removal, and semantic segmentation. The quantitative and qualitative results demonstrate the effectiveness and universality of the proposed method. 
    \item Considering that there is no standard protocol to train and evaluate the performance of guided image filtering algorithms,  we reimplement eight recently proposed state-of-the-art deep learning-based guided filtering models and unify their settings to facilitate fair comparison. All of the codes and trained models are publicly available\footnote{https://github.com/zhwzhong/DAGF} to encourage reproducible research.
\end{itemize}

The remainder of this paper is organized as follows. Sect.~\ref{gfr} gives a brief introduction to the relevant works of guided filter. Sect.~\ref{sec_3} introduces the proposed method for guided image filtering.  Sect.~\ref{exp} provides experimental comparisons with existing state-of-the-art methods for a varied range of guided filtering tasks. Ablation experiments are presented in Sect.~\ref{abl} to analyze the network hyper-parameters and verify the advantage of each components proposed in our model. We conclude the paper in Sect.~\ref{con}.

\section{Guided Filters Revisiting}
\label{gfr}

In this section, we start with a revisiting of formal definitions of popular variants of guided filters in the literature, and then explain our generalization of them to derive the proposed deep attentional guided image filter.

\subsection{Classical Guided Filters}

Define the guidance image as $\bm{g}$ and the target image as $\bm{t}$, the  output $\bm{f}$ of guided filtering can be represented as:
\begin{equation}
    \bm{f}_i = \sum_j \bm{W}_{i,j}(\bm{g},\bm{t})\bm{t}_j,
\end{equation}
where $i$ and $j$ are pixel coordinates; $\bm{W}_{i,j}$ is the filter kernel weight, whose parameters $(\bm{g},\bm{t})$ mean that it can be derived from either $\bm{g}$ or $\bm{t}$, or both.

In the classical bilateral filter and guided image filter, $\bm{W}_{i,j}$ is only dependent on the guidance $\bm{g}$. Specifically, the filter weight in bilateral filter is defined as:
\begin{equation}
    \bm{W}^{BF}_{i,j} = \frac{1}{C_i}\exp\left(-\frac{\|i-j\|}{\sigma_s}\right)\exp\left(-\frac{\|\bm{g}_i-\bm{g}_j\|}{\sigma_r}\right),
\end{equation}
where $C_i$ is the normalization parameter; $\sigma_s$ and $\sigma_r$ are parameters for geometric and photometric similarity, respectively. In guided image filter (He et al.,~\cite{GF}), the filter kernel weight is defined as:
\begin{equation}
    \bm{W}^{GIF}_{i,j} = \frac{1}{|\bm{N}_k|^2}\sum_{k:(i,j)\in \bm{N}_k}\left(1+\frac{(\bm{g}_i-\mu_k)(\bm{g}_j-\mu_k)}{\sigma^2_k+\epsilon}\right),
\end{equation}
where $|\bm{N}_k|$ is the number of pixels in a window $\bm{N}_k$; $\mu_k$ and $\sigma^2_k$ are the mean and variance of $\bm{g}$ in $\bm{N}_k$.

\subsection{Learning-based Guided Filters}

Among deep learning based approaches for guided filter design, dynamic filter network~\cite{DFN} first defines a filter-generating network (FGN) that takes the guidance $\bm{g}$ as input to obtain location-specific dynamic filters $\bm{F}_\theta = \text{FGN}(\boldsymbol{g,\theta})$, which are then applied to the target image $\bm{t}$ to yield the output ${\bm{f}} = \bm{F}_{\bm{\theta}}(\bm{t})$. Pixel-adaptive convolution~\cite{PanNet} defines the filter kernel by multiplying a spatially varying filter on standard spatially invariant kernel:
\begin{equation}
    \bm{f}_i = \sum_{j\in \bm{N}_i} \bm{K}(\bm{g}_i,\bm{g}_j)\bm{W}[p_i-p_j]\bm{t}_j+b,
\end{equation}
where $\bm{W}$ is the spatially invariant kernel; $\bm{K}(\cdot,\cdot)$ is a varying filter kernel function that has a fixed form such as Gaussian, $[p_i-p_j]$ denotes the index offset of kernel weights. 
From the above formulation, it can be found that, similar to bilateral filter and guided image filter, dynamic filter network and pixel-adaptive convolution also only depend on the guidance $\bm{g}$ in defining the filter kernels. When there are inconsistent structures in the guidance and the target, this approach would generate annoying artifacts in the output. 

The recent deep joint filtering (DJF) method~\cite{DJFR} alleviates this drawback by jointly leveraging features of both the guidance and the target. It designs two-branch sub-networks to extract features from the guidance and the target respectively, which are passed through a fusion sub-network to output the filtering result. The joint filter $\bm{\mathrm{\Phi}}$ is learned in an end-to-end manner by the following optimization:
\begin{equation}
    \bm{\mathrm{\Phi}}^* = \arg\min_{\bm{\mathrm{\Phi}}}\|\bm{f}^{gt} - \bm{\mathrm{\Phi}}(\bm{g},\bm{t})\|^2,
\end{equation}
where $\bm{f}^{gt}$ is the ground truth of the output.
In contrast to the implicit filter learning approach of DJF,  deformable kernel networks (DKN)~\cite{DKN} explicitly learns the kernel weights $\bm{K}$ and offsets $\bm{s}$ using two-branch sub-networks from the two images. Concretely, the filtering is performed by
\begin{equation}
    \bm{f}_i = \sum_{j\in\bm{N}_i} \bm{W}_{i,\bm{s}(j)}(\bm{g},\bm{t})\bm{t}_{\bm{s}(j)},
\end{equation}
with
\begin{equation}
    \bm{W}(\bm{g},\bm{t}) = \bm{K}(\bm{g}) \odot  \bm{K}(\bm{t}),
\end{equation}
where $\bm{K}(\bm{g})$ and $ \bm{K}(\bm{t})$ are kernel weights learned from the guidance and the target, respectively, $\odot $ denotes element-wise multiplication. Although DJF and DKN achieve better performance than previous methods, they treat the guidance and target images as independent information and utilize separate networks for kernel learning, thus the complementary information contained in the two images cannot be fully exploited. In addition, the fusion approach of multi-modal weights through element-wise multiplication is not effective, in which the guidance and the target contribute equally to the final filtering results. 

\begin{figure*}[!t]
    \begin{center}
        \includegraphics[width=\linewidth]{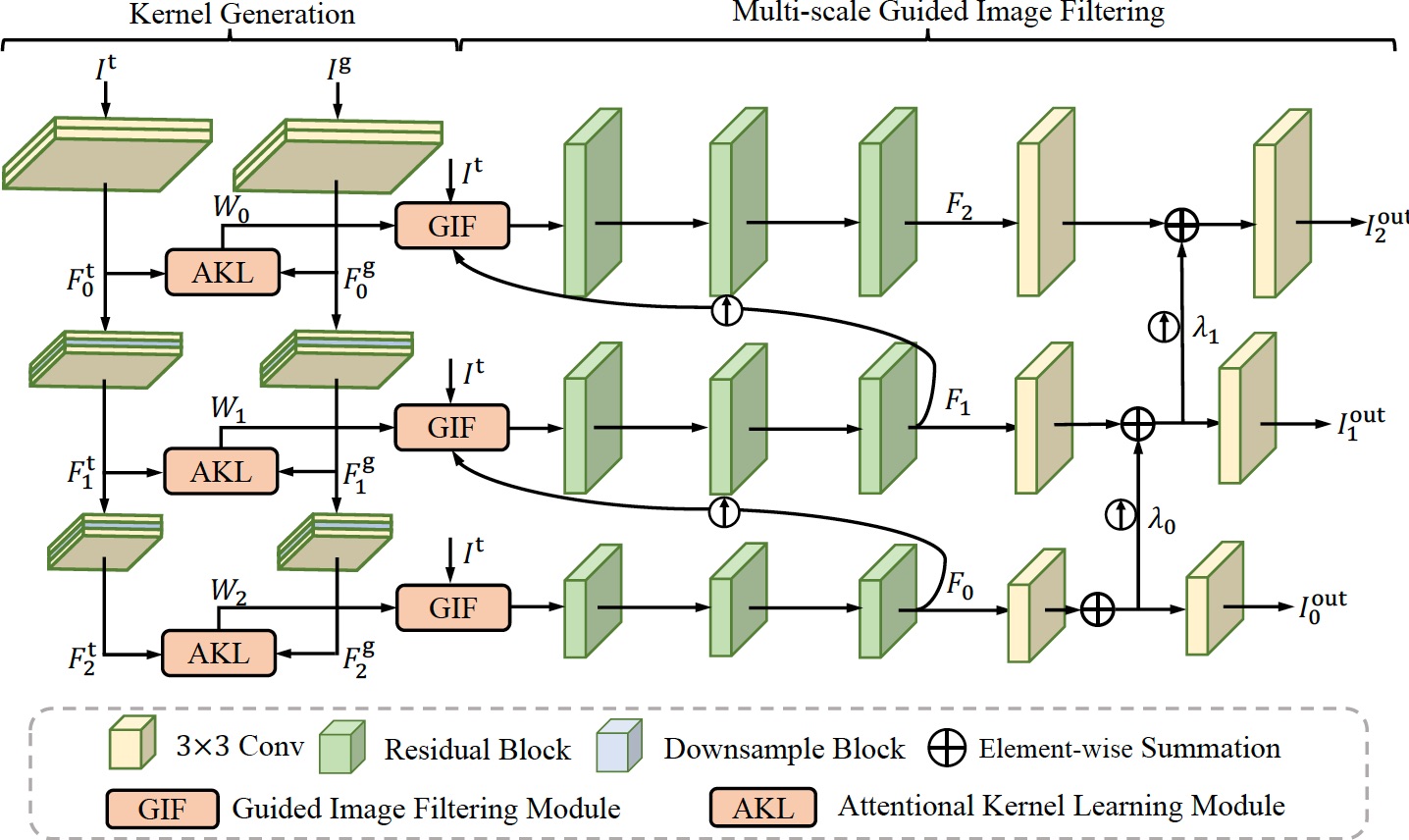}
    \end{center}
    \caption{The network architecture of the proposed deep attentional guided image filtering (DAGF) with the number of pyramid level $m=3$. DAGF consists of a kernel generation network for constructing filter kernels and a multi-scale guided image filtering network with the purpose of filtering target image by using the generated kernels.}
    \label{fig:dag}
    % \vspace{-0.1in}
\end{figure*}

\subsection{Our Strategy}
Considering the drawbacks of existing methods, we propose a deep attentional guided image filtering scheme to more effectively leverage multi-modal information. Our method performs dual regression on both guidance and target, and combines them adaptively using an attention mechanism. Mathematically, our filtering process can be generally formulated as
\begin{equation}
    \bm{f}_i = \sum_{j\in\bm{N}_i} \bm{A}_{i,j}\bm{W}^\text{g}_{i,j}\bm{t}_{j} + \sum_{j\in\bm{N}_i}(1-\bm{A}_{i,j})\bm{W}^\text{t}_{i,j}\bm{t}_{j},
\end{equation}
where $\bm{W}^\text{g}_{i,j}$ and $\bm{W}^\text{t}_{i,j}$ are filter kernels computed from the guidance and the target respectively; $\bm{A}_{i,j}$ denotes the pixel-wise reliability weight of the guidance image, which is determined automatically by considering both guidance and target information. The above formulation means that, when the guidance information is not trustworthy, we should turn to use the target information itself for regression, so as to prevent the unreliable structure propagation. %In the next section, we will introduce in detail the proposed network design. 
\section{Proposed Method}

% The goal of guided image filtering is to transfer structures of the guidance to the target for enhancing spatial resolution or suppressing noise. However, since there commonly exist inconsistent structures in the two images, directly passing all patterns of the guidance to the target could introduce significant errors.
An effective guided image filtering scheme should be able to identify the consistent structures contained in the guidance as well as avoid transferring extraneous or erroneous contents to the target. In this section, we introduce in detail the proposed deep attentioanl guided image filtering (DAGF) framework for this purpose, where the complementary information contained in the two images can be fully explored in both kernel generation and image filtering process. %In the following, we will introduce the proposed method in details.
\label{sec_3}
\subsection{Network Architecture}

The DAGF takes a target image $\bm{I}^\text{t} \in \mathbb{R}^{\text{H}\times \text{W} \times \text{C}^\text{t}}$ (\textit{e.g.}, low-resolution depth) and a guidance image $\bm{I}^\text{g} \in \mathbb{R}^{\text{H}\times \text{W} \times \text{C}^\text{g}}$ (\textit{e.g.}, high-resolution color image) as inputs, and generates a reconstructed image $\bm{I}^{\text{out}} \in \mathbb{R}^{\text{H}\times \text{W} \times \text{C}^\text{t}}$ as output, where $H, W$ and $C$ denote the height, width and the number of channels respectively. 
%In case of guided image super-resolution, to achieve resolution consistency, the input low-resolution depth is firstly upsampled by Bicubic to get its HR initilization. 

Fig.~\ref{fig:dag} illustrates the overall architecture of the proposed network, which is composed of \textit{kernel generation sub-network} and \textit{multi-scale guided filtering sub-network}.  Instead of directly predicting kernels in image domain and enlarging its receptive field by using the deformable sampling strategy as~\cite{DKN}, we employ a pyramid architecture to achieve a large receptive field, and conduct filter learning in the feature domain since deep features are more robust with respect to appearance difference of the target and the guidance. 
\begin{itemize}
    \item In the filter kernel generation sub-network, the multi-scale features of $\bm{I}^\text{t}$ and $\bm{I}^\text{g}$ are fed into the attentional kernel learning (AKL) module to generate filter kernels $\{W_i\}$. The network architecture of AKL is illustrated in Fig.~\ref{fig:akg}, where an attentional contribution module based on U-Net architecture is designed to adaptively fuse the filter kernels generated by the guidance and the target. 
    \item In the guided filtering sub-network, with the derived pixel-wise filter kernels, features of the target image are processed in a coarse-to-fine manner to get the upsampled features.   
\end{itemize}
The above process is repeated until arriving the final scale. In the following, we will elaborate these two sub-networks and the loss function design for network training.

\subsection{Filter Kernel Generation}
\label{akl}

\begin{figure}[!t]
    \centering
    \includegraphics[width=\linewidth]{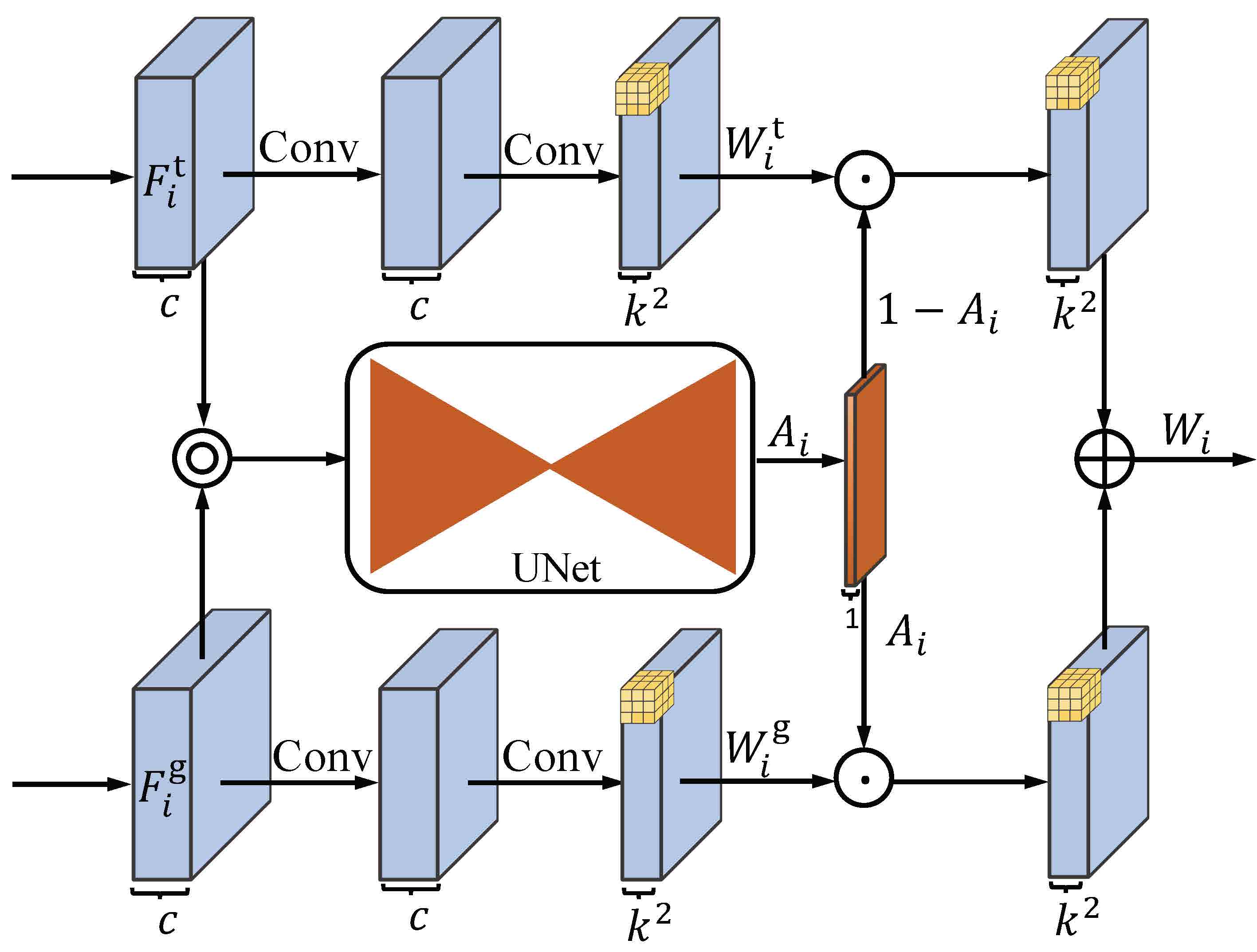}
    \caption{The network architecture of the proposed attentional kernel learning (AKL) module, where $\odot$ denotes the element-wise multiplication and $\circledcirc$ is concatenation operation.}
    \label{fig:akg}
\end{figure}

The filter kernel generation sub-network is tailored to generate spatial-variant kernels by considering the mutual dependency between the target and the guidance. As illustrated in the left part of Fig.~\ref{fig:dag}, given $\bm{I}^\text{t}$ and $\bm{I}^\text{g}$ as inputs, we first employ two-branch pyramid network to extract multi-scale features \{$\bm{F}_i^\text{t}, 0 \leq i < m$\} and $\{\bm{F}^\text{g}_i, 0\leq i < m\}$ from the target and the guidance, respectively. We take the target branch as an example, which is done by:
\begin{align}
    \bm{F}^\text{t}_0 &= \mathrm{Conv}(\mathrm{Conv}(\bm{I}^\text{t})), \\
    \bm{F}^\text{t}_i &= \mathrm{{Down}}( \bm{F}^\text{t}_{i-1}), 0<i<m,
\end{align}
where $m$ denotes the level of pyramid network and $\mathrm{Conv}(\cdot)$ is the convolution operator; $\mathrm{Down}(\cdot)$ represents the down-sample block with scale factor 2, which is implemented by two convolution layers and a inverse pixel-shuffle~\cite{invp} operation.
%\{$\bm{F}_i^\text{t}, 0 \leq i < m$\} are the extracted multi-scale target features. the guidance features  are obtained with the same operations as the target branch. 

For guided image filtering, the prior information for reconstruction is either from the guidance image if there are consistent structures between the guidance and target images, or from the target image itself if there is no reliable guidance information. This inspires us to design dual regression over the guidance and the target, respectively, instead of only relying on  the guidance as done in most existing methods. To this end, as shown in Fig.~\ref{fig:akg},  we propose an attentional kernel learning (AKL) module. It takes the extracted guidance and target features as inputs and consists of two steps: dual kernels generation and adaptive kernels combination.

The first step is the dual kernels generation, which is formulated as:
\begin{align}
    \bm{W}_i^\text{t} &= \mathrm{Conv}(\mathrm{Conv}(\bm{F}_i^\text{t})), 0 \leq i < m, \\
    \bm{W}_i^\text{g} &= \mathrm{Conv}(\mathrm{Conv}(\bm{F}_i^\text{g})), 0 \leq i < m, 
\end{align}
where $\bm{W}^\text{t}_i$ and $\bm{W}^\text{g}_i$ are the $i$-th constructed filter kernels from the target and guidance features respectively. The spatial-resolution of $i$-th kernels is the same as the one of its corresponding input features while the number of channels are $k^\text{2}$ where $k$ is the desired filter kernel size. However, these kernels generated by the target or guidance information alone cannot explore the dependencies among them, making the filtering outputs suffer from blurring or texture copying artifacts. To alleviate this problem, we introduce adaptive kernels combination module based on a light-weight UNet architecture, which takes both guidance and target features as inputs and models the pixel-wise dependencies among them in a learning manner. This process is formulated as:
\begin{align}
    \bm{A}_i = \mathrm{UNet} ([\bm{F}_i^\text{t}, \bm{F}_i^\text{g}]), 0 \leq i < m,
\end{align}
where $\mathrm{UNet}$ is a five-layer U-like~\cite{ronneberger2015u} network, $[\cdot, \cdot]$ denotes concatenation operation; $ \bm{A}_i$ is the output of this module, which can be considered as an attention map to adaptively combine kernels constructed from guidance and target features. The final guided filter kernels can be derived as:
\begin{align}
    \bm{W}_i = \bm{A}_i \odot \bm{W}^\text{g}_i + (\bm{1} - \bm{A}_i) \odot \bm{W}^\text{t}_i, 0 \leq i < m,
    \label{eq1}
\end{align}
where $\bm{W}_i$ is the generated $i$-th filter kernel; $\bm{1}$ denotes the all-1 matrix; $\odot$ means element-wise multiplication.

\subsection{Multi-scale Guided Filtering}
\label{msf}
After generating the guided filter kernels, the following step is to perform filtering on the target image, which is done by the guided filtering sub-network. As shown in the right part of Fig.~\ref{fig:dag}, it takes the target image $\bm{I}^\text{t}$ as the input, and progressively filters the input target image  by using the learned filter kernels $\{\bm{W}_0, \cdots, \bm{W}_{m-1}\}$ in a coarse-to-fine manner. 

Specifically, given $\bm{I}^\text{t}$ as input, we first utilize Bicubic to resize it to the same resolution as its corresponding filter kernels:
\begin{align}
    \bm{\hat{I}^\text{t}} = \text{Bicubic} (\bm{I}^\text{t}).
\end{align}
Then the filtering process can be formulated as:
\begin{align}
    \bm{F}_0 &= \mathrm{ResNet}( \mathrm{GIF} ( \mathrm{Conv} ( \bm{\hat{I}^\text{t}}) ) , \bm{W}_0,),\\
    \bm{F}_i &= \mathrm{ResNet}( \mathrm{GIF} ( [\bm{F}_{i-1}^{\uparrow}, \mathrm{Conv} ( \bm{\hat{I}^\text{t}})] ,\bm{W}_i)  ), 0<i < m,
\end{align}
where $[\cdot, \cdot]$ means concatenation operation and $\mathrm{ResNet}(\cdot)$ is the function including three residual blocks (He et al.,~\cite{ResNet}); $\bm{F}_i$ is $i$-th filtered target feature; $\uparrow$ is upsampling operation. $\mathrm{GIF}(\cdot)$ is a filtering operation that conducts filtering operation on the corresponding target features. Specifically, we first reshape the third dimension of the filter from $k^\text{2}$ to $k\times k$, then the filtering process for a pixel \{$(u, v) | 0\leq u < H, 0 \leq v < W$\} can be defined as following:
\begin{align}
    \bm{F}(u, v) = \sum_{x=-\sigma}^{\sigma}\sum_{y=-\sigma}^{\sigma} \bm{W}_{u, v}(x,y) \cdot \tilde{\bm{F}}(u-x, v-y),
\end{align}
where $\sigma=\lfloor k/2 \rfloor$; $\bm{\hat{F}}$ is the output of the GIF module.

Based on $\{\bm{F}_i\}_{i=0}^{m-2}$, we can obtain the filter results of DAGF by using the proposed the multi-scale fusion strategy:
\begin{align}
    & \hat{\bm{F}}_0 = \mathrm{Conv}(\bm{F}_0), \\
   & \hat{\bm{F}}_i = \mathrm{Conv}(\bm{F}_i) + \lambda_{i-1} \cdot \hat{\bm{F}}_{i-1}^{\uparrow} , 0 < i < m,\\
   & \bm{I}_i^{\mathrm{out}} = \mathrm{Conv}(\hat{\bm{F}}_i) + \bm{I}^\text{t}, 0 \leq i < m-1,
\end{align}
where $\lambda_i$ is a learnable parameter that is initialized as 0. The parameter enables the output layer first to rely on features of the current layer and then gradually learn to combine high-level features from previous layers. Therefore, the output of the last layer can enjoy the merit of preserving both high-level contextual details and low-level spatial information. $\{\bm{I}_i^{\text{out}}\}_{i=0}^{m-2}$ are the intermediate multi-scale results and $\bm{I}_{m-1}^{\text{out}}$ is the final filtering result of the proposed scheme. %The residual learning strategy is employed to accelerate the training process.

\subsection{Loss function}
\label{lf}
We adopt the residual learning strategy to train the proposed method. Let $\bm{I}^\text{g}$ and $\bm{I}^\text{t}$ be the input guidance and target image, $\bm{I}^\text{h}$ be the corresponding ground-truth image. The proposed DAGF network aims to learn the residual between $\bm{I}^\text{h}$ and $\bm{I}^\text{t}$. The overall all loss function is composed of three terms: a $\bm{\text{L}}_1$ loss $\mathcal{L}_1$, a multi-stage loss $\mathcal{L}_{ms}$ and a boundary-aware loss $\mathcal{L}_{b}$:
\begin{itemize}
    \item \textbf{$\bm{\text{L}_1}$ loss}. $\mathcal{L}_1$ measures the pixel-wise errors between the output image $\bm{I}^\text{out}_{m-1}$ and its corresponding residual image $\bm{I}^\text{r}$:
\begin{align}
    \mathcal{L}_1 = ||\bm{I}^\text{h} - \bm{I}^\text{out}_{m-1} ||_1.
\end{align}
\item \textbf{Multi-stage loss}. To stabilize the network training process and promote the multi-stage guided filtering module to learn more effective parameters, we propose a multi-stage loss to enforce all intermediate results to be close to the ground truth residual image:
%by using multi-scale ground-truth images. All of the intermediate results are encouraged to be close to the ground-truth at the corresponding resolution:
\begin{align}
    \mathcal{L}_{ms} = \frac{1}{m-1}\sum_{i=0}^{m-2}||\bm{I}^\text{h} - \text{Bicubic}(\bm{I}^\text{out}_i)||_1,
\end{align}
where $m$ is the number of pyramid levels. We use Bicubic interpolation to resize the output image $\bm{I}^\text{out}_i$ to the same resolution as the ground truth target image $\bm{I}^\text{h}$.
\item \textbf{Boundary-aware loss}. Optimizing the pixel-wise loss (e.g., $\mathcal{L}_1$ and $\mathcal{L}_2$) typically cannot preserve high-frequency structure information well, and tends to produce blurry images as all  pixels are treated equally. To mitigate this problem and encourage the network to give more emphasis on the high-frequency parts, we propose a boundary-aware loss to promote our model to generate sharper boundaries. Specifically, we first employ Sobel operator $\nabla$ to detect the boundary information of the ground truth and 
the network output, and obtain the boundary mask $\bm{M}$:
\begin{align}
    \bm{M} = (\nabla_x \bm{I}^\text{h} - \nabla_x \bm{I}^\text{out}_{m-1}) \odot (\nabla_y \bm{I}^\text{h} - \nabla_y \bm{I}^\text{out}_{m-1}),
\end{align}
then the boundary-aware loss is defined as:
\begin{align}
    \mathcal{L}_{ba} = ||\bm{M} \odot \bm{I}^\text{h} - \bm{M} \odot \bm{I}^\text{out}_{m-1} ||_1,
\end{align}
where $\odot$ denotes element-wise multiplication.
%%% Depth Map Suoer-Resolution

%%%% End
\end{itemize}

With these three losses, the total loss is then formulated as:
\begin{equation}
    \mathcal{L} = \omega_1 \cdot \mathcal{L}_1 + \omega_2 \cdot \mathcal{L}_{ba} + \omega_3 \cdot \mathcal{L}_{ms},
\end{equation}
where $\omega_1, \omega_2$ and $\omega_3$ are hyper-parameters to balance these loss functions. We set $\omega_3=1$ to stabilize the training procedure at early stage and then progressively decay to zero with the training progress to boost the performance of final output. We set $\omega_1= 1, \omega_2=10$, respectively.
\subsection{Implementation Details}
In our model, we set the number of pyramid levels as $m = 3$ and the size of generated kernel in AKL modules as $3\times 3$. The ablation study presented blow will verify the effectiveness of our configuration. The hyper-parameters of our model are $\omega_1=1, \omega_2 = 0.001$ and $\omega_3 = 1$ . All the convolution layers within the proposed methods are sized of $3\times 3$ and the channels of intermediate features are 32. We use PReLU~\cite{PReLU} as the default activation function. We utilize PixelShuffle~\cite{PixelShuffle} and InvPixelShuffle~\cite{invp} as the up-sampling and down-sampling operators to resize the features in our model.

In the training phase, the batch size is set as 32 and we random crop $256\times 256$ image patches from the target and guidance images as inputs. We augment the training data with random flipping and rotation. Adam~\cite{adam:} with $\beta_1=0.9$ and $\beta_2=0.999$ is employed as optimizer. The initial learning rate is set as $1\times 10^{-4}$ and we halve it every 80 epochs, stop the training after 100 epochs. Our model is implemented by Pytorch~\cite{PyTorch} and trained on one RTX 1080ti GPU. Training the proposed method roughly takes 2 day for NYU v2~\cite{NYU} datasets.

Our network takes three channels guidance and one channel target images as inputs. For the multi-channels target images, we apply the trained model separately for each channel and the outputs are combined to obtain the final result. For the single-channel guidance image, we copy the single-channel three times to generate three-channels guidance image.

\begin{table*}[!t]\setlength{\tabcolsep}{10pt}\renewcommand{\arraystretch}{1.2}
	\begin{center}
	\caption{\label{tab:nyu_result} Quantitative comparison for depth image super-resolution on four standard RGB/D datasets in terms of average RMSE values. Following the experimental setting of~\cite{DJF, DKN}, we calculate the average RMSE values in cenrimeter for NYU v2~\cite{NYU} dataset. For other datasets, we compute the RMSE values by scaling the depth value to the range [0, 255]. The best performance for each case are highlighted in \textbf{boldface} while the second best ones are \underline{underscored}. For RMSE metric, the lower values mean the better performance.}
		\begin{tabular}{lcccccccccccc}
		\toprule
		{Datasets} & \multicolumn{3}{c}{Middlebury} & \multicolumn{3}{c}{Lu} & \multicolumn{3}{c}{NYU v2} &  \multicolumn{3}{c}{Sintel} \\
		\cmidrule(lr){2-4} \cmidrule(lr){5-7} \cmidrule(lr){8-10} \cmidrule(lr){11-13} 
		{Method}  &$4\times$ & $8 \times$  & $16\times$  & $4\times$ & $8 \times$  & $16\times$  &$4\times$ & $8 \times$  & $16\times$ &$4 \times$ & $8 \times$  & $16\times$\\
		\midrule
		Bicubic & 4.44 & 7.58 & 11.87 & 5.07 & 9.22 & 14.27 & 8.16 & 14.22 & 22.32  & 10.11 & 14.51 & 19.95 \\
		MRF~\cite{MRF} & 4.26 & 7.43 & 11.80 & 4.90 & 9.03 & 14.19 & 7.84 & 13.98 & 22.20 &9.87& 13.45& 18.19  \\
		GF~\cite{GF}) & 4.01 & 7.22 & 11.70 & 4.87 & 8.85 & 14.09 & 7.32 & 13.62 & 22.03  & 8.83 & 12.60 & 18.78 \\		
		TGV~\cite{TGV}) & 3.39 & 5.41 & 12.03 & 4.48 & 7.58 & 17.46 & 6.98 & 11.23 & 28.13 & 8.30 & 13.05 & 19.96  \\
% 		Park~\citep{Park} & 2.82 & 4.08 & 7.26 & 4.09 & 6.19 & 10.14 & 5.21 & 9.56 & 18.10 \\
		SDF~\cite{sdf}) & 3.14 & 5.03 & 8.83 & 4.65 & 7.53 & 11.52 & 5.27 & 12.31 & 19.24 & 9.20& 13.63& 19.36 \\
		FBS~\cite{FBS}) & 2.58&  4.19 & 7.30&  3.03 & 5.77&  8.48 & 4.29 & 8.94 & 14.59 & 8.29& 10.31& 16.18\\
		JBU~\cite{JBU}) & 2.44 & 3.81 & 6.13 & 2.99 & 5.06 & 7.51 & 4.07 & 8.29 & 13.35 &8.25& 11.74& 16.02  \\
		\multicolumn{10}{l}{Experiment results for depth map super-resolution (Nearest-neighbour down-sampling).}  \\
		DGF~\cite{DGF}) & 3.92& 6.04& 10.02& 2.73& 5.98& 11.73& 4.50& 8.98& 16.77& 7.53& 11.53& 17.50  \\
		DJF~\cite{DJF}) & 2.14 & 3.77 & 6.12 & 2.54 & 4.71 & 7.66 & 3.54 & 6.20 & 10.21 & 7.09& 9.12& 12.36 \\
		DMSG~\cite{DMSG}) &  \underline{1.79} & 3.39 & 5.87 & 2.48 & 4.74 & 7.51 & 3.48 & 6.07 & 10.27 & 6.80& 9.09& 11.81 \\
		DJFR~\cite{DJFR}) & 1.98 & 3.61 & 6.07 &  \underline{2.22} &  4.54 & 7.48 & 3.38 & 5.86 & 10.11 &7.05& 9.12& 12.61 \\
		DSRN~\cite{DepthSR}) & 2.08 &  3.26 &  5.78 & 2.57 &  4.46 &  6.45 & 3.49 &  5.70 &  9.76 & 7.29 &  9.43 &  11.62  \\
		PAC~\cite{PanNet}) & 1.91 & 3.20 & 5.60 & 2.48 & 4.37 & 6.60 & 2.82 & 5.01 & 8.64 &6.79& \underline{8.36} & \underline{11.02} \\
		DKN~\cite{DKN}) & 1.93 &  \underline{3.17} &  \underline{5.49} & 2.35 & \underline{4.16} & \underline{6.33} &  \underline{2.46} &  \underline{4.76} &  \underline{8.50}  & \underline{6.84} & 8.61& 11.21 \\
		DAGF(Ours) & \textbf{1.78}& \textbf{2.73}&\textbf{4.75}& \textbf{1.96}& \textbf{3.81}& \textbf{6.16}& \textbf{2.35}& \textbf{4.62}& \textbf{7.81}& \textbf{6.72}& \textbf{8.35}& \textbf{10.64}  \\
		\multicolumn{10}{l}{Experiment results for depth map super-resolution (Bicubic down-sampling).}  \\
		DGF~\cite{DGF})  & 1.94 & 3.36 & 5.81 & 2.45 & 4.42 & 7.26 & 3.21 & 5.92 & 10.45 & 5.91 & 8.02 & 11.17  \\
		DJF~\cite{DJF}) & 1.68 & 3.24 & 5.62 & 1.65 & 3.96 & 6.75 & 2.80 & 5.33 & 9.46 & 5.30 & 7.53 & 10.41 \\
		DMSG~\cite{DMSG}) & 1.88 & 3.45 & 6.28 & 2.30 & 4.17 & 7.22 & 3.02 & 5.38 & 9.17 & 4.73& 6.26 & \underline{8.36} \\
		DJFR~\cite{DJFR}) & 1.32 & 3.19 & 5.57 & 1.15 & 3.57 & 6.77 & 2.38 & 4.94 & 9.18 & 4.90 & 7.39 & 10.33\\
		DSRN~\cite{DepthSR}) & 1.77 &  3.05 &  4.96 & 1.77 &  3.10 &  \underline{5.11} & 3.00 &  5.16 &  8.41 & 4.49 &  6.53 &  9.28 \\
		PAC~\cite{PanNet}) &  1.32 & 2.62 & 4.58 & 1.20 & 2.33 & 5.19 & 1.89 & 3.33 & 6.78 & 4.42 & 6.13 & 8.42\\
		DKN~\cite{DKN}) &  \underline{1.23} & \underline{2.12} &\underline{4.24} & \underline{0.96} & \underline{2.16} & \underline{5.11} & \underline{1.62} & \underline{3.26} & \underline{6.51} & \underline{4.38} & \underline{5.89}  & 8.40  \\
		DAGF(Ours) & \textbf{1.15} & \textbf{1.80}& \textbf{3.70} & \textbf{0.83} & \textbf{1.93} & \textbf{4.80} & \textbf{1.36} & \textbf{2.87} & \textbf{6.06} & \textbf{3.84} &\textbf{5.59} & \textbf{7.44}  \\
% 		\multicolumn{10}{l}{Experiment results for depth map super-resolution (bicubic down-sampling and gaussian noisy).}  \\
% 		DGF~(Wu et al., \cite{DGF}) & 2.70 & 4.13 & 6.38 & 4.06 & 5.85 & 8.39 & 6.52 & 9.23 & 13.00 & 6.94 & 9.03 & 12.05 \\
% 		DJF~(Li et al., \cite{DJF}) & 1.80 &  2.99 &  5.16 & 1.85 &  3.13 &  5.39 & 3.74 &  5.95 &  9.61 & 4.88 &  6.93 &  10.05 \\
% 		DMSG~(Hui et al., \cite{DMSG}) & 1.79 &  2.69 &  4.75 & 1.88 &  \underline{2.79} &  4.84 & 3.60 &  5.31 &  \underline{8.07} & 4.74 &  6.36 &  \underline{8.72} \\
% 		DJFR~(Li et al., \cite{DJFR}) &1.86 &  3.07 &  5.27 & 1.91 &  3.21 &  5.51 & 4.01 &  6.21 &  9.90 & 5.10 &  7.12 &  10.23  \\
% 		DSRN~(Guo et al., \cite{DepthSR}) & 1.84 &  2.99 &  4.70 & 1.97 &  2.98 &  5.94 & 4.36 &  6.31 &  9.75 & 5.49 &  7.21 &  9.80 \\
% 		PAC~(Su et al., \cite{PanNet}) & 1.81 & 2.94 & 5.08 & 1.93 & 3.44 & 6.18 & 4.23 & 6.24 & 9.54 & 5.40 & 7.32 & 9.89 \\
% 		DKN~(Kim et al., \cite{DKN}) &\underline{1.76} &  \underline{2.68} &  \underline{4.55} & \underline{1.81} &  2.82 &  \underline{4.81} & \underline{3.39} &  \underline{5.24} &  8.41 & \underline{4.51} &  \underline{6.25} &  9.20 \\
% 		DAGF (Ours) & \textbf{1.72} &  \textbf{2.61} &  \textbf{4.24} & \textbf{1.74} &  \textbf{2.72} &  \textbf{4.51} & \textbf{3.25} &  \textbf{5.01} & \textbf{7.54} & \textbf{4.42} &  \textbf{6.09} & \textbf{8.25} \\
		\bottomrule
		\end{tabular}
	\end{center}
\end{table*}
\begin{figure*}[!tb]
	\begin{center}
		\subfigure[Guidance]{
		\begin{minipage}[b]{0.15\linewidth}
			\includegraphics[width=1\linewidth]{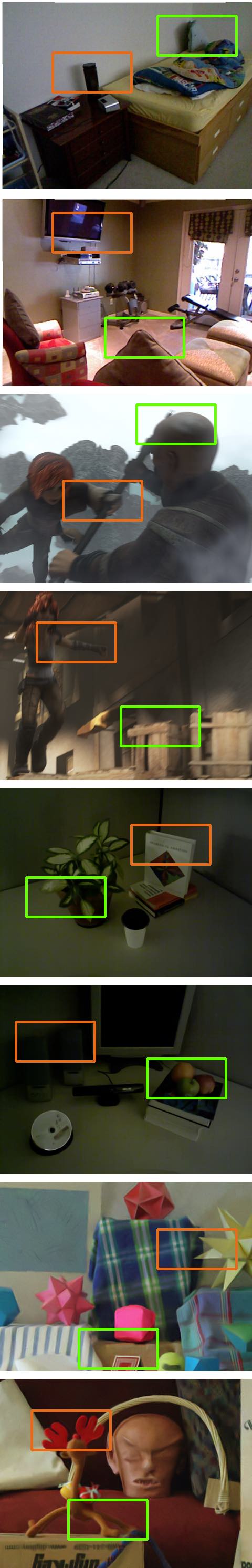} 
		\end{minipage}
		\hspace{-0.089in}
	}\subfigure[JBU]{
    		\begin{minipage}[b]{0.1\linewidth}
  		 	\includegraphics[width=1\linewidth]{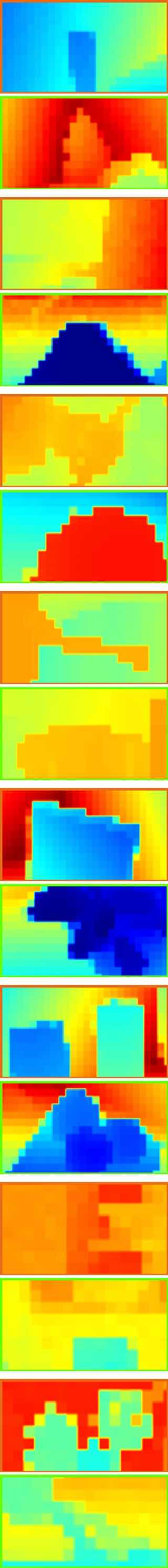}
    		\end{minipage}
    		\hspace{-0.089in}
    	}\subfigure[GF]{
		\begin{minipage}[b]{0.1\linewidth}
			\includegraphics[width=1\linewidth]{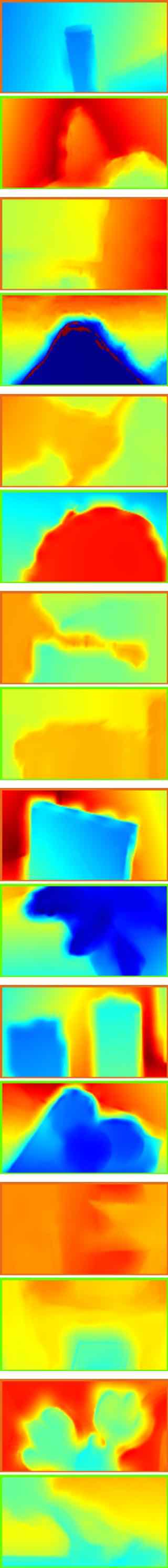} 
		\end{minipage}
		\hspace{-0.089in}
	}\subfigure[DMSG]{
		\begin{minipage}[b]{0.1\linewidth}
			\includegraphics[width=1\linewidth]{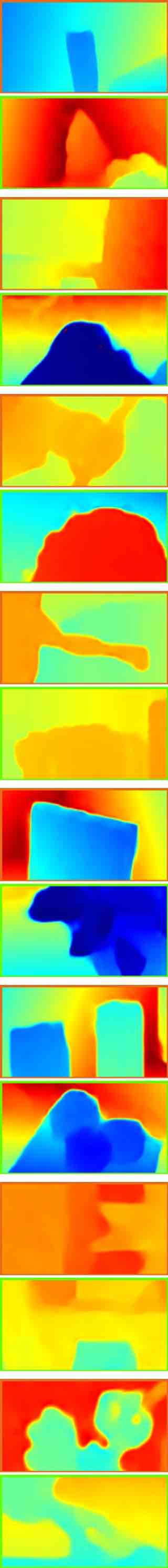} 
		\end{minipage}
		\hspace{-0.089in}
	}\subfigure[DJFR]{
		\begin{minipage}[b]{0.1\linewidth}
			\includegraphics[width=1\linewidth]{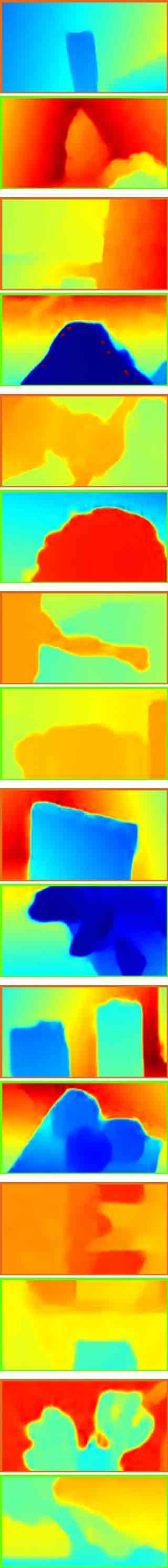} 
		\end{minipage}
		\hspace{-0.089in}
	}\subfigure[PAC]{
		\begin{minipage}[b]{0.1\linewidth}
			\includegraphics[width=1\linewidth]{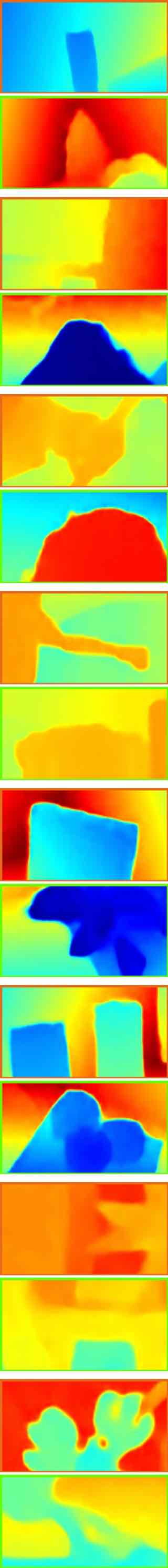} 
		\end{minipage}
		\hspace{-0.089in}
	}\subfigure[DKN]{
		\begin{minipage}[b]{0.1\linewidth}
			\includegraphics[width=1\linewidth]{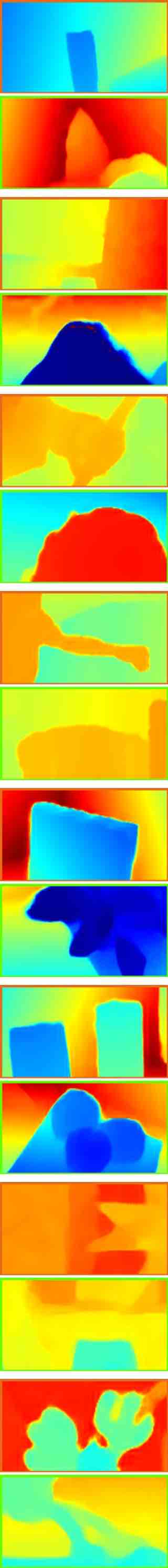} 
		\end{minipage}
		\hspace{-0.089in}
	}\subfigure[DAGF]{
		\begin{minipage}[b]{0.1\linewidth}
			\includegraphics[width=1\linewidth]{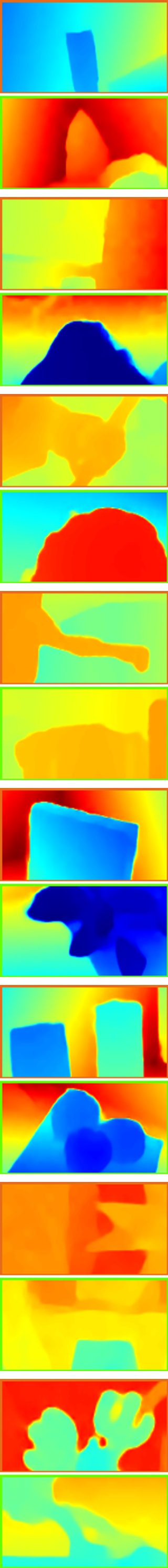} 
		\end{minipage}
		\hspace{-0.089in}
	}\subfigure[GT]{
		\begin{minipage}[b]{0.1\linewidth}
			\includegraphics[width=1\linewidth]{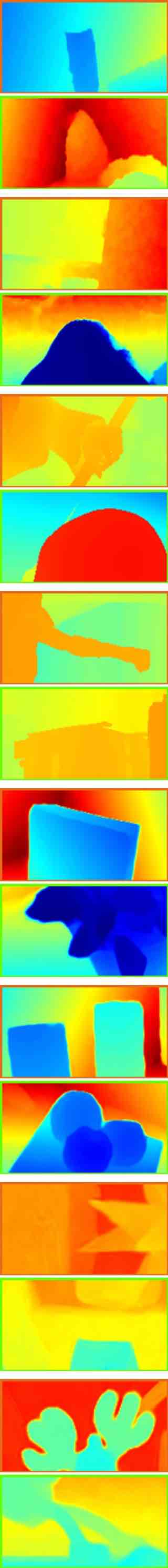} 
		\end{minipage}
	}
	\end{center}
% 	\vspace{-0.1in}
	\caption{Qualitative comparison for recovered depth maps (8×). (a) Guidace image, (b): JBU~\cite{JBU}, (c): GF~\cite{GF}, (d): DMSG~\cite{DMSG}, (e): DJFR~ \cite{DJFR}, (f): PAC~\cite{PanNet}, (g): DKN~\cite{DKN}, (h): DAGF, and (i) Ground truth depth map. Top to bottom: Each two rows present recovered depth maps on the NYU v2~\cite{NYU}, Sintel~\cite{Sintel}, Lu~\cite{Lu} and Middlebury~\cite{middleblur_data_1} datasets respectively. Please enlarge the PDF for more details.}
	\label{fig:fig_all}
\end{figure*}

\section{Experiments}
\label{exp}
In this section, we conduct extensive experiments to evaluate the performance of the proposed method on a wide range of guided image filtering tasks, including  guided image super-resolution (\eg\ depth image super-resolution and saliency map super-resolution, Sect.~\ref{gsr}), cross-modality image restoration (\eg\ joint depth image super-resolution and denoising, and flash/non-flash image denoising, Sect.~\ref{cm}), texture removal (Sect.~\ref{tr}) and image semantic segmentation (Sect.~\ref{ss}).

For fair comparison, the results for the compared methods are generated by using the source codes released by their authors with the default parameter settings, and all learning based methods are trained and tested on the same datasets.

\begin{figure*}[!htb]
	\begin{center}
		\subfigure[Guidance]{
		\begin{minipage}[b]{0.114\linewidth}
			\includegraphics[width=1\linewidth]{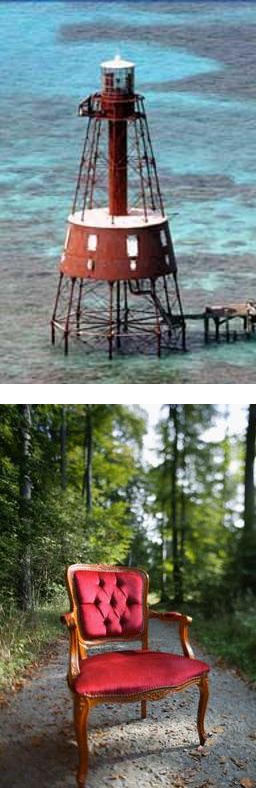} 
		\end{minipage}
	}\subfigure[Bicubic]{
    		\begin{minipage}[b]{0.114\linewidth}
  		 	\includegraphics[width=1\linewidth]{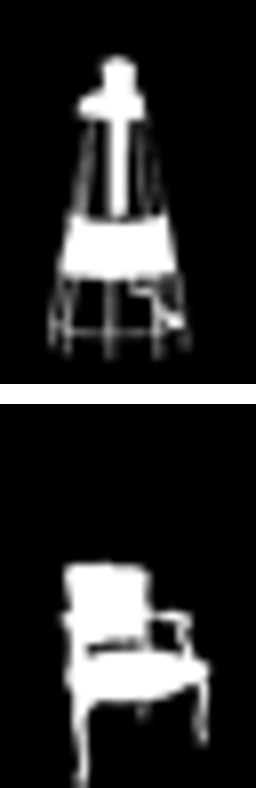}
    		\end{minipage}
    	}\subfigure[DMSG]{
		\begin{minipage}[b]{0.114\linewidth}
			\includegraphics[width=1\linewidth]{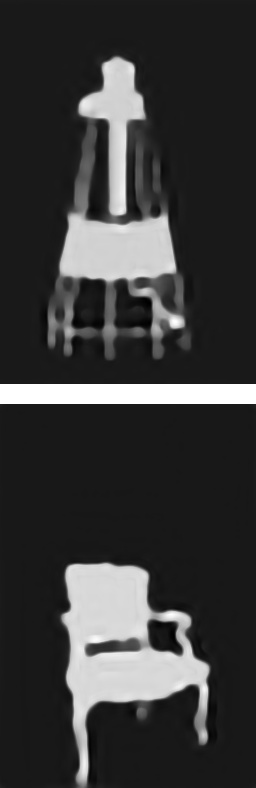} 
		\end{minipage}
% 		\hspace{-0.1in}
	}\subfigure[DJFR]{
		\begin{minipage}[b]{0.114\linewidth}
			\includegraphics[width=1\linewidth]{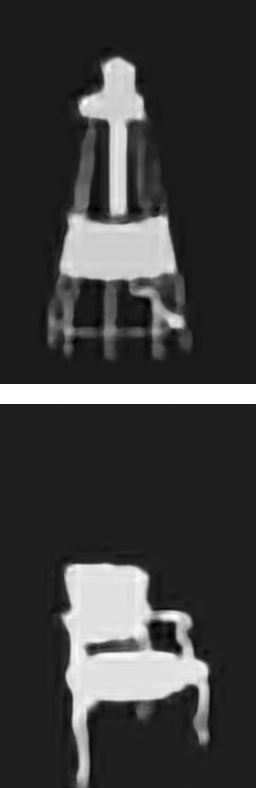} 
		\end{minipage}
	}\subfigure[PAC]{
		\begin{minipage}[b]{0.114\linewidth}
			\includegraphics[width=1\linewidth]{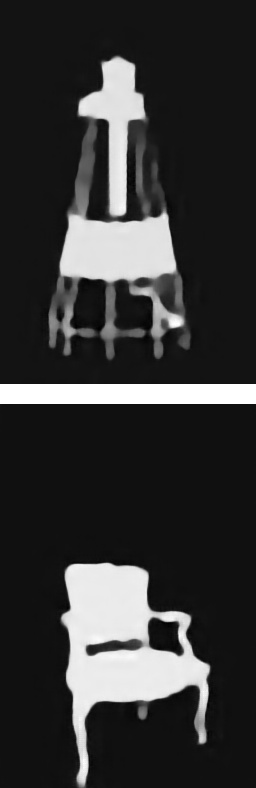} 
		\end{minipage}
	}\subfigure[DKN]{
		\begin{minipage}[b]{0.114\linewidth}
			\includegraphics[width=1\linewidth]{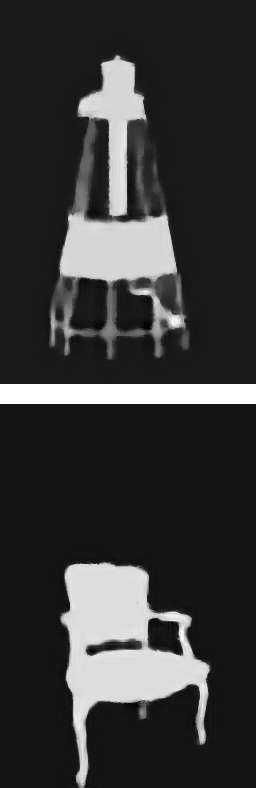} 
		\end{minipage}
	}\subfigure[DAGF]{
		\begin{minipage}[b]{0.114\linewidth}
			\includegraphics[width=1\linewidth]{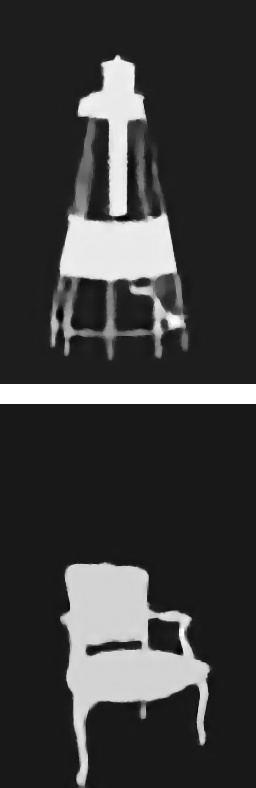} 
		\end{minipage}
	}\subfigure[GT]{
		\begin{minipage}[b]{0.114\linewidth}
			\includegraphics[width=1\linewidth]{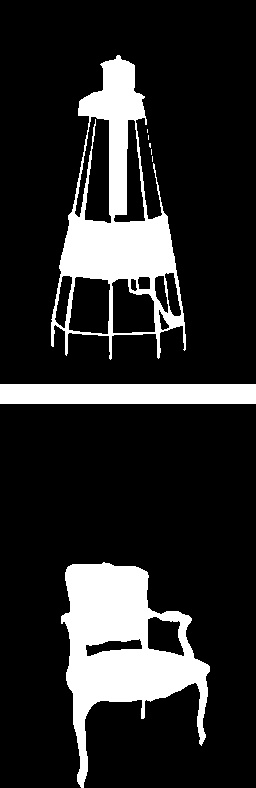} 
		\end{minipage}
	}
	\end{center}
% 	\vspace{-0.1in}
	\caption{Visual comparison of $8\times$ saliency map super-resolution on the DUT-OMRON dataset~\cite{DUT}: (a): Guidance, (b): low-resolution image, (c): DMSG~\cite{DMSG}, (d): DJFR~\cite{DJFR}, (e): PAC~\cite{PanNet}, (f): DKN~\cite{DKN}, (e): DAGF, (h): Ground truth. Please enlarge the PDF for more details.}
	\label{fig:fig_sal}
\end{figure*}

\subsection{Guided Image Super-resolution}
\label{gsr}
Guided image super-resolution (GSR) is a classic computer vision task which aims to reconstruct a high-resolution (HR) image from a low-resolution (LR) one with the help of a HR image from another modality. For example, we can obtain a HR depth by GSR using a LR depth and a HR RGB image as inputs, where the HR RGB image serves as the guidance.

Following the experimental settings of~\cite{DJFR, DKN}, we train our model on the task of depth image super-resolution, and then evaluate the performance of the model on tasks of depth image super-resolution and saliency map super-resolution, the latter one is used to verify the generalization ability of our model. 

\textbf{RGB-guided Depth Super-resolution.} For this task, we use the first 1000 RGB-D image pairs from NYU v2 dataset~\cite{NYU} as the training set. In order to make fair comparison with existing methods, we exploit the nearest-neighbour downsampling as the standard downsampling operator to generate LR target image from the ground-truth. Three scales are considered, including $4\times, 8\times, 16\times$. To show the effectiveness of the proposed method, we further conduct experiments on Bicubic downsampling as done in~\cite{DKN}. The performance of the proposed method is evaluated on the following four standard benchmark datasets:
\begin{itemize}
    \item [\textbullet] Sintel dataset~\cite{Sintel}: this dataset consists of 1064 image pairs which are obtained by an animated 3D movie.
    \item [\textbullet] NYU v2 dataset~\cite{NYU}: this dataset contains 1449 image pairs acquired by Microsoft Kinect. We use the last of 449 image pairs to evaluate the performance of our method.
    \item [\textbullet] Lu dataset~\cite{Lu}: it contains 6 image pairs captured by ASUS Xtion Pro camera.
    \item [\textbullet] Middlebury dataset~\cite{middleblur_data_1, middleblur_data_2}: this dataset is captured by structure light, and we utilize the 30 image pairs from 2001-2006 datasets with the missing depth values generated by~Lu et al.~\cite{Lu}.
\end{itemize}

We compare our method with 13 state-of-the-art methods, including two local filtering-based methods: GF~\cite{GF} and JBU~\cite{JBU}; four global optimization-based methods: MRF~ \cite{MRF}, TGV~\cite{TGV}, SDF~\cite{sdf} and FBS~\cite{FBS}; seven deep learning-based methods: DGF~\cite{DGF}, DJF~\cite{DJF}, DMSG~\cite{DMSG}, DJFR~\cite{DJFR}, DSRN~\cite{DepthSR}, PAC~\cite{PanNet} and DKN~\cite{DKN}. We adopt Root Mean Square Error (RMSE) as the evaluation metric. Lower RMSE values mean higher recovery quality.

Table~\ref{tab:nyu_result} summarizes the quantitative comparison results between ours and other state-of-the-art methods. The best performance is highlighted in bold. As can be seen from this table, our method achieves the best results among all the compared methods on both synthetic and real datasets (\eg\ the Sintel and NYU v2 dataset) and on three scales. The superior performance benefits from the more precise filter kernels learned and the multi-scale filtering process. Compared with the second best results (underlined), our results obtain the gains of 0.12($4\times$), 0.24(8$\times$) and 0.39(16$\times$) withe respect to average RMSE values. 

\begin{table*}[!htp]\setlength{\tabcolsep}{12pt}\renewcommand{\arraystretch}{1.2}
    \begin{center}
        \caption{\label{tab:tab_sal}Quantitative comparison of $8 \times$ saliency map super-resolution on the DUT-OMRON dataset~\cite{DUT}. Following DJFR~ \cite{DJFR}, we use F-measure to calculate the difference between the predicted saliency map and the corresponding ground-truth. The best performance for each case is highlighted in \textbf{boldface} while the second one is \underline{underscored} For F-measure, the higher values mean the better performance.}
        \begin{tabular}{ccccccccc}
        \toprule
        Methods & Bicubic & GF~\cite{GF} &  DMSG~\cite{DMSG} & DJFR~\cite{DJFR} & PAC~\cite{PanNet} & FDKN~\cite{DKN} & DKN~\cite{DKN} & DAGF (Ours) \\
        \midrule
        Fscore &  0.853 & 0.821 & 0.910 & 0.901 & 0.922 & 0.921& \underline{0.926} & \textbf{0.932} \\
    \bottomrule
    \end{tabular}
    \end{center}
\end{table*}

To further analyze the performance of the proposed method, we present the visual results for $8\times$ depth image super-resolution in Fig.~\ref{fig:fig_all}. It can be observed that the results of  JBU~\cite{JBU} suffer from  jaggy artifacts. The results of GF~\cite{GF} are over-smoothed, which indicates that the local filter is not effective at large scale factors (\eg\ $8 \times$). Compared to GF~\cite{GF} and JBU~\cite{JBU}, the learning-based methods are capable of generating results with clearer boundaries. However, for finer details, \eg, the arm in the second image and the rope in the last image, the compared learning-based methods exhibit obvious artifacts such as blurring on the arm and wrong estimation on the rope, which implies that the downsampling degradation brings significantly damage on the small objects and therefore makes those regions harder to recover. On the contrary, the results obtained by the proposed method are clearer, sharper, and more faithful to the ground truth image.  

\textbf{RGB-guided Saliency Map Super-resolution}. To further demonstrate the generalization ability of the proposed method, we apply the model that is trained on NYU v2 dataset directly to the task of saliency map super-resolution without any fine-tuning step. Similar to DKN~\cite{DKN}, we use 5168 image pairs from DUT-OMRON dataset~\cite{DUT} to evaluate the SR performance. We use bicubic interpolation (8$\times$) to generate the low-resolution saliency maps and then super-resolve them with the corresponding high-resolution color image as the guidance. The quantitative results in terms of F-measure are listed in Table~\ref{tab:tab_sal}. As can be seen from this table, our DAGF achieves the best result among all the compared methods and outperforms the second best method by a large margin, which demonstrates the generalization ability of the proposed method. In addition, we random select two images and visualize the recovered high-resolution saliency map obtained by different methods in Fig.~\ref{fig:fig_sal}. It can be observed that the results of Bicubic are over-smoothed, in which the structure details are severely damaged. DMSG~\cite{DMSG} and DJFR~\cite{DJFR} struggle to generate clear boundaries. The results of DKN~\cite{DKN} have certain artifacts around the edge area. In contrast, our method is able to generate high-quality saliency maps as well as keep the sharpest boundaries, which indicates that the proposed method can fully take advantage of the guidance image and effectively transfer meaningful structure information. 
\begin{figure*}[!tb]
	\begin{center}
		\subfigure[Guidance]{
		\begin{minipage}[b]{0.1958\linewidth}
			\includegraphics[width=1\linewidth]{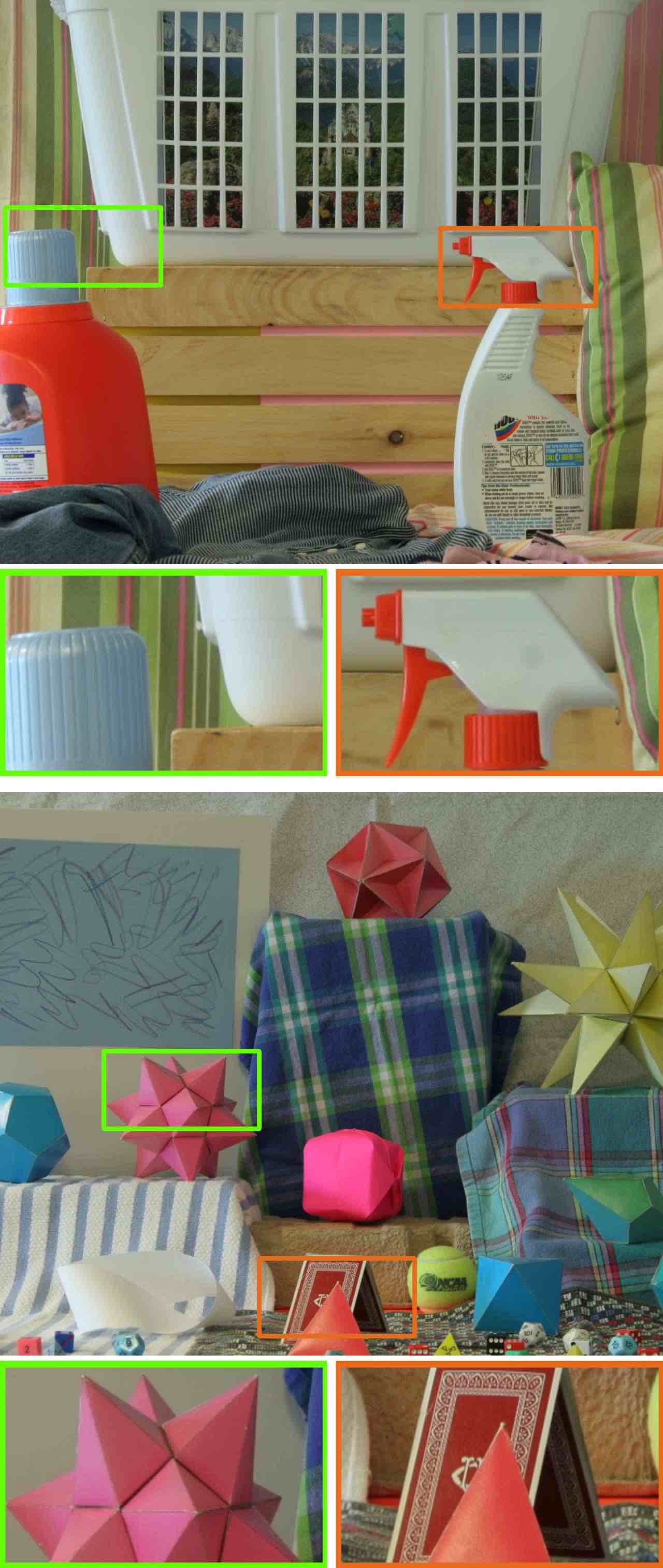} 
		\end{minipage}
		\hspace{-0.1in}
		\vspace{-0.2in}
	}\subfigure[Target]{
    		\begin{minipage}[b]{0.1958\linewidth}
  		 	\includegraphics[width=1\linewidth]{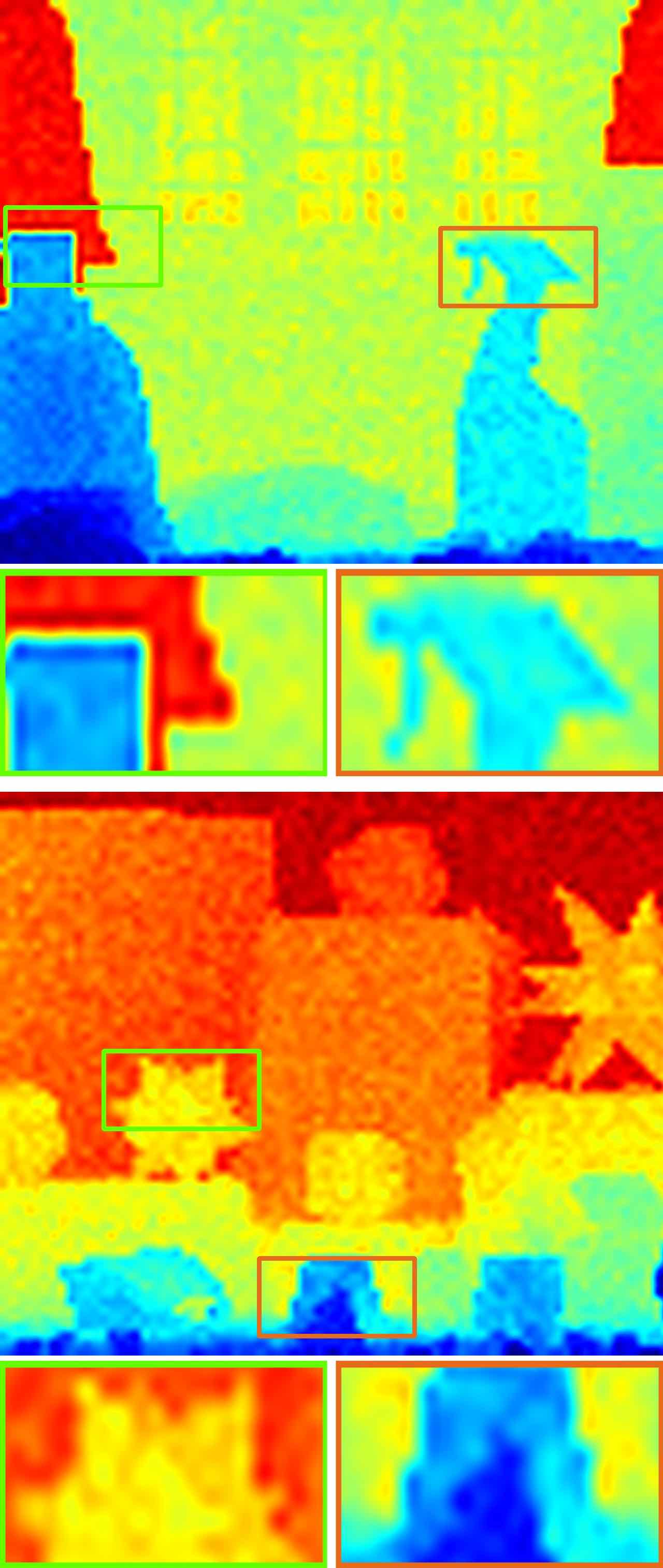}
    		\end{minipage}
    		\hspace{-0.1in}
    	}\subfigure[GF]{
		\begin{minipage}[b]{0.1958\linewidth}
			\includegraphics[width=1\linewidth]{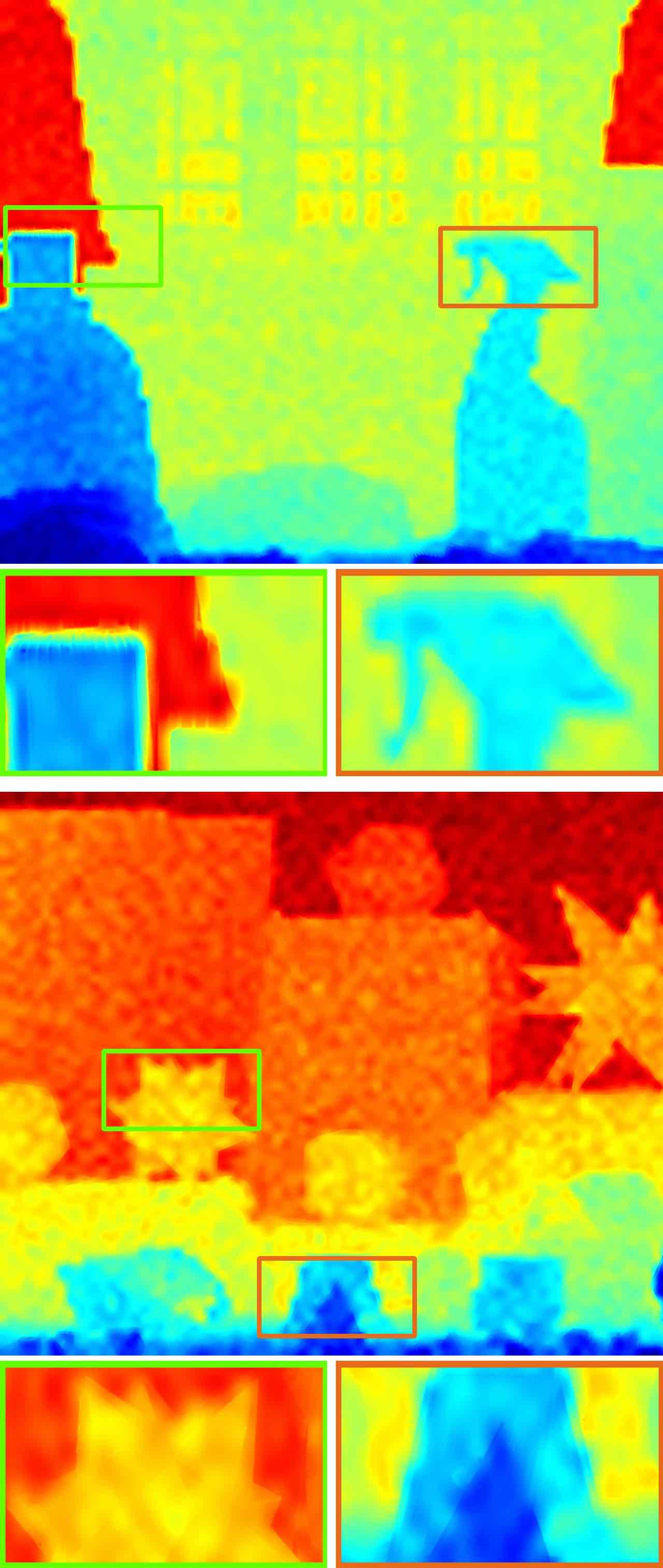} 
		\end{minipage}
		\hspace{-0.1in}
	}\subfigure[MUF]{
		\begin{minipage}[b]{0.1958\linewidth}
			\includegraphics[width=1\linewidth]{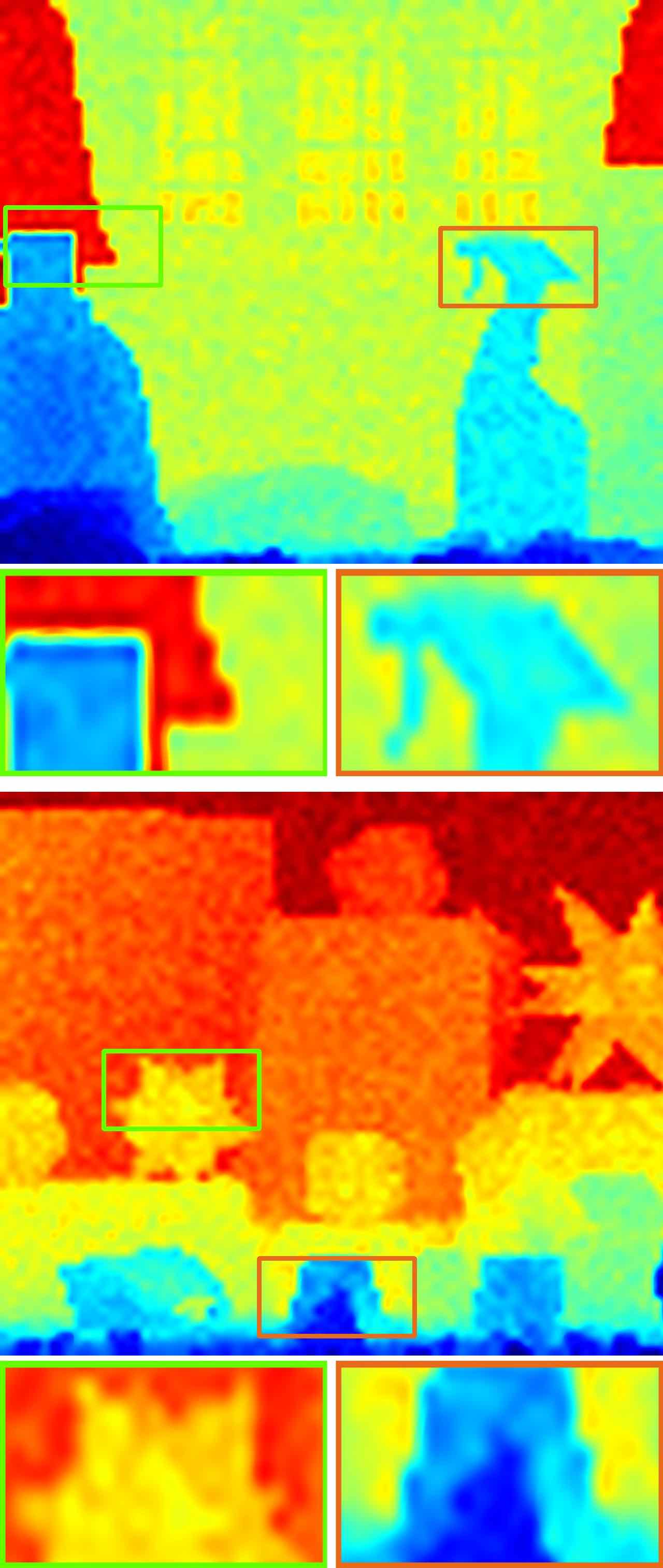} 
		\end{minipage}
		\hspace{-0.1in}
	}\subfigure[SDF]{
		\begin{minipage}[b]{0.1958\linewidth}
			\includegraphics[width=1\linewidth]{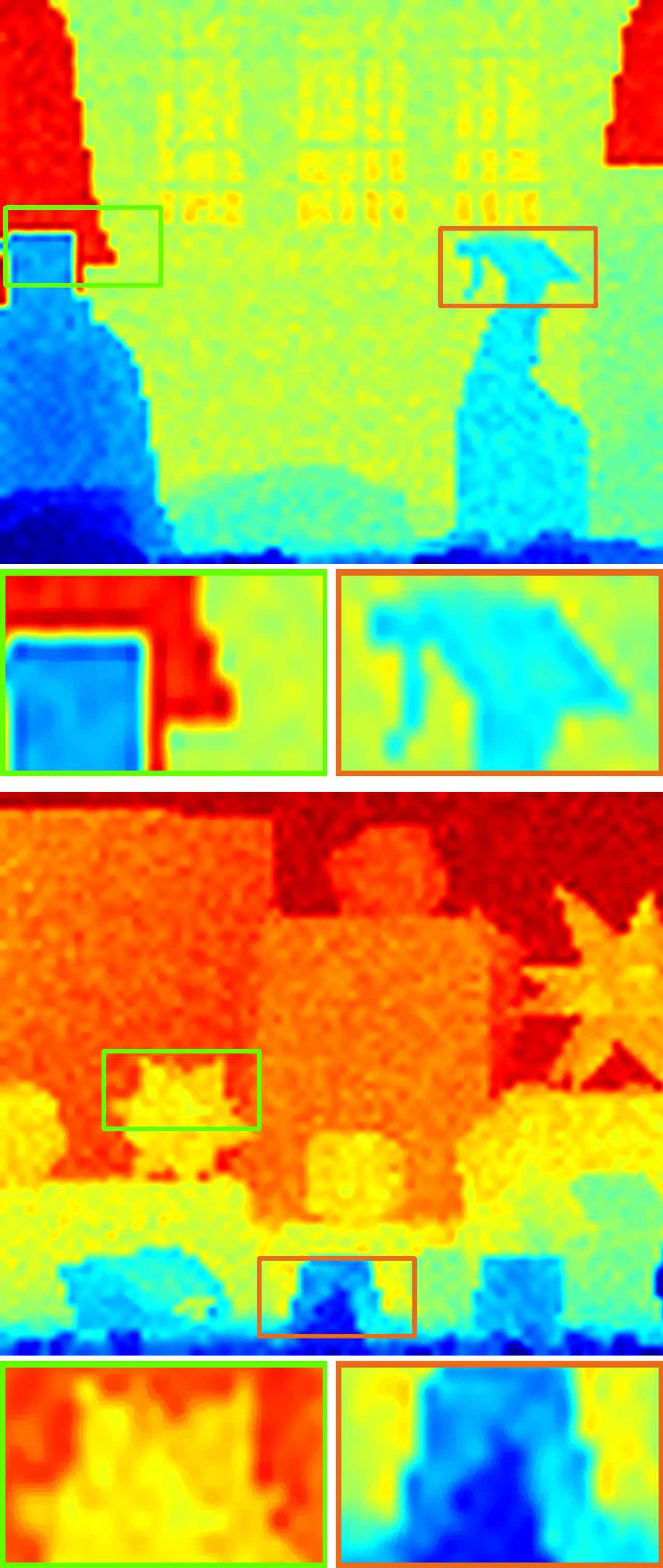} 
		\end{minipage}
		\hspace{-0.1in}
	}
	\subfigure[PAC]{
		\begin{minipage}[b]{0.1958\linewidth}
			\includegraphics[width=1\linewidth]{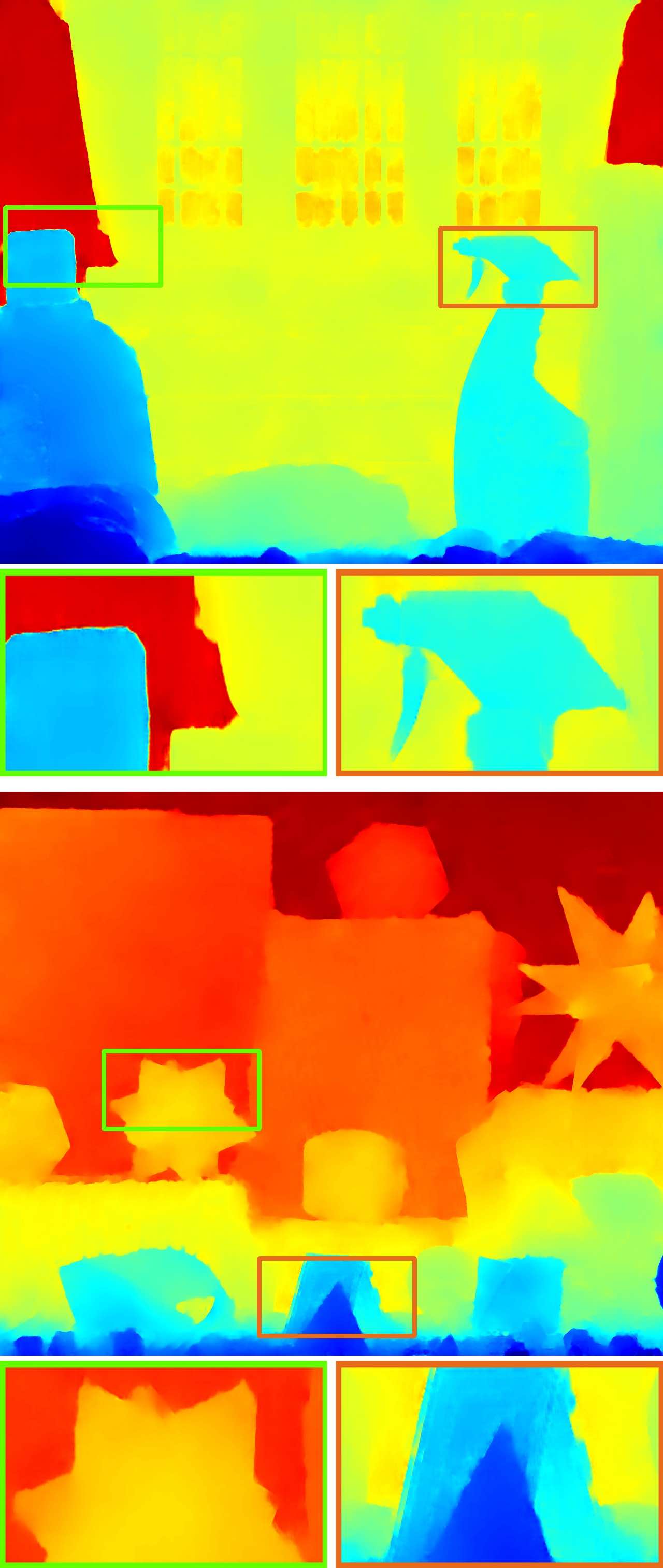} 
		\end{minipage}
		\hspace{-0.1in}
	}\subfigure[DJFR]{
		\begin{minipage}[b]{0.1958\linewidth}
			\includegraphics[width=1\linewidth]{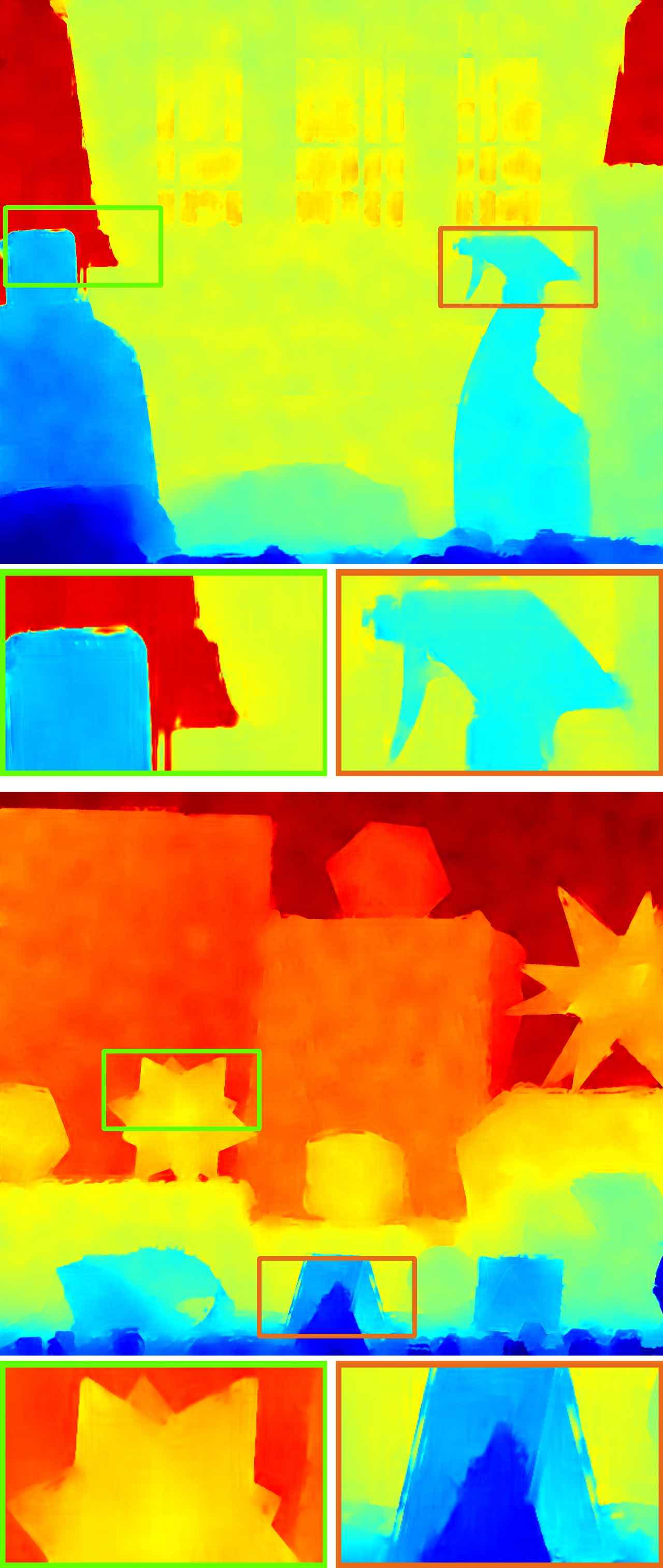} 
		\end{minipage}
		\hspace{-0.1in}
	}\subfigure[DKN]{
		\begin{minipage}[b]{0.1958\linewidth}
			\includegraphics[width=1\linewidth]{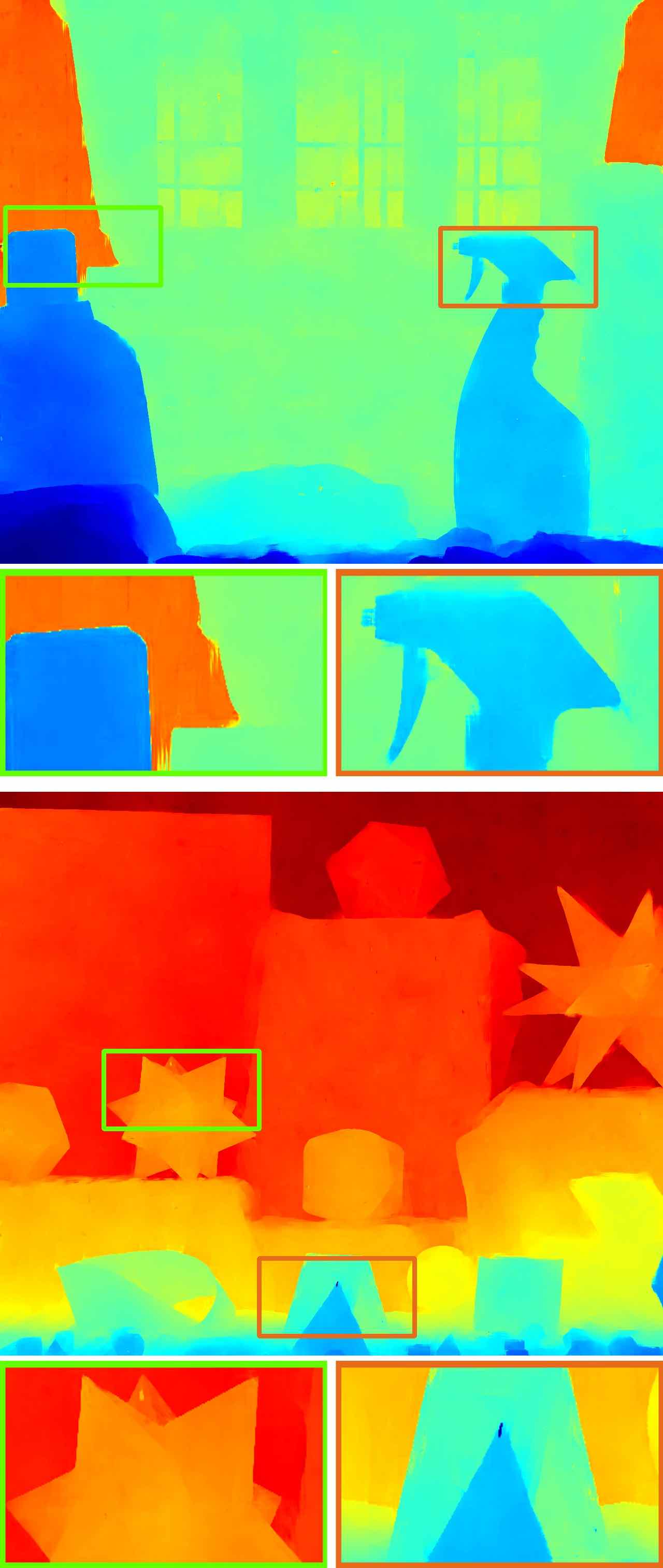} 
		\end{minipage}
		\hspace{-0.1in}
	}\subfigure[DAGF]{
		\begin{minipage}[b]{0.1958\linewidth}
			\includegraphics[width=1\linewidth]{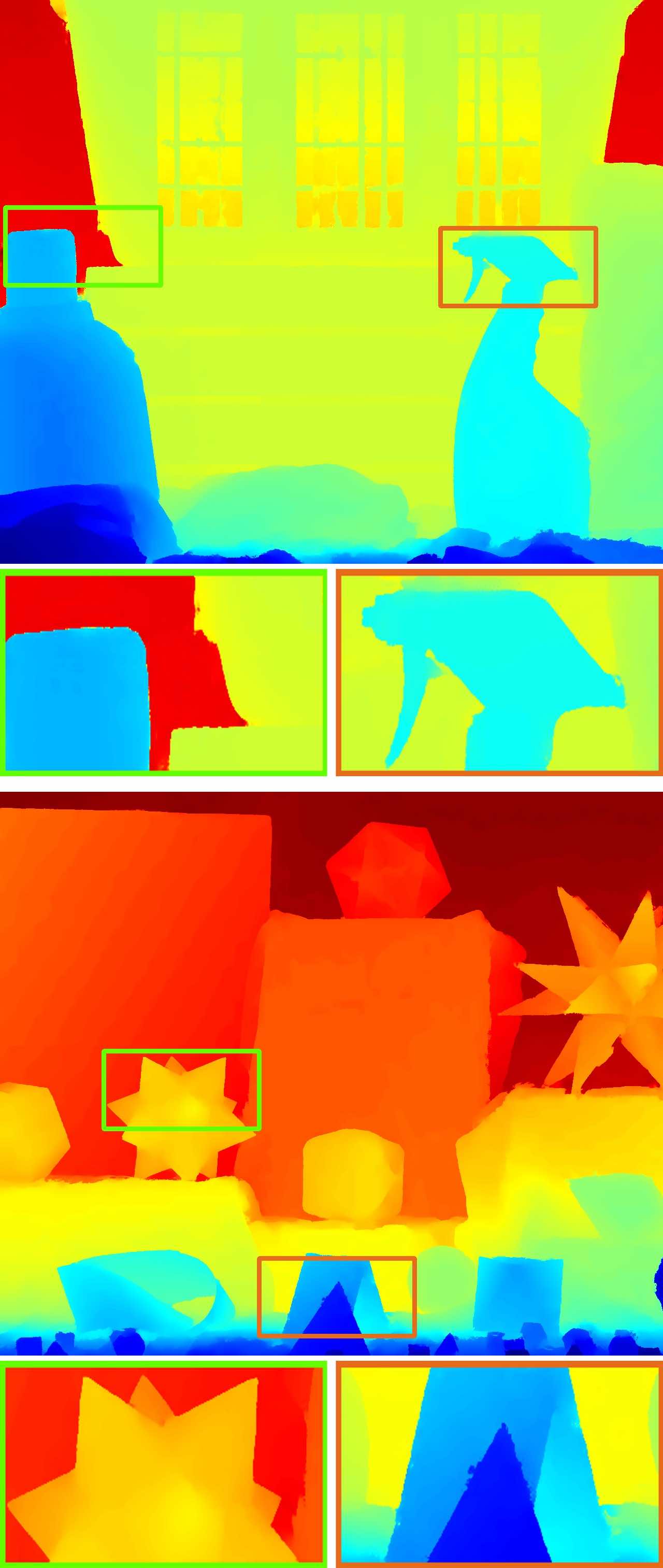} 
		\end{minipage}
		\hspace{-0.1in}
	}\subfigure[GT]{
		\begin{minipage}[b]{0.1958\linewidth}
			\includegraphics[width=1\linewidth]{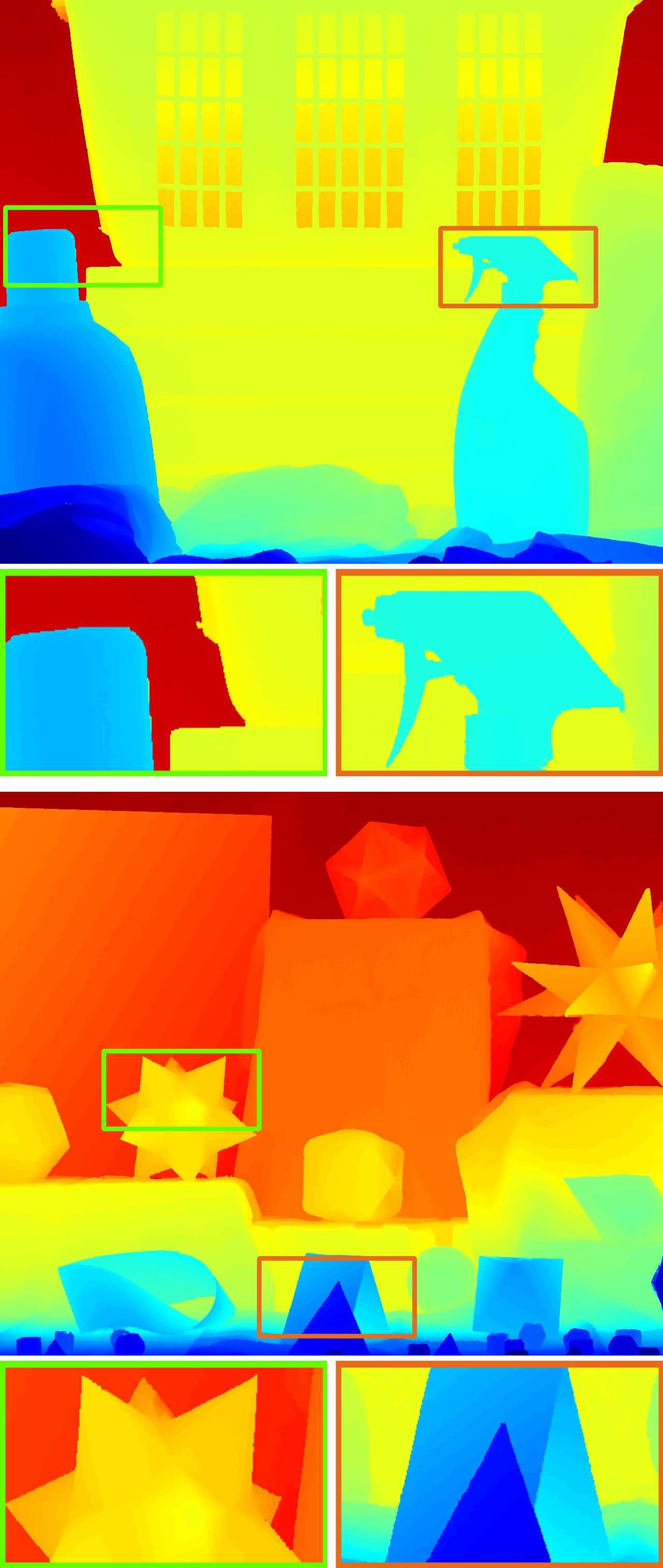} 
		\end{minipage}
	}
	\end{center}
% 		\vspace{-0.1in}
	\caption{Qualitative comparison of joint depth map super-resolution and denoising. Please enlarge the PDF for more details. (a): Guidance Image, (b): Target image, (c): GF~\cite{GF}, (d): MUF~\cite{mugf}, (e): SDF~\cite{sdf}, (f): PAC~\cite{PanNet}, (g): DJFR~\cite{DJFR}, (h): DKN~\cite{DKN}, (i): DAGF and (j): Ground-truth image. ease enlarge the PDF for more details.}
	\label{fig:fig_noisy}
\end{figure*}
\begin{figure*}[!tb]
	\begin{center}
		\subfigure[Guidance]{
		\begin{minipage}[b]{0.1385\linewidth}
			\includegraphics[width=1\linewidth]{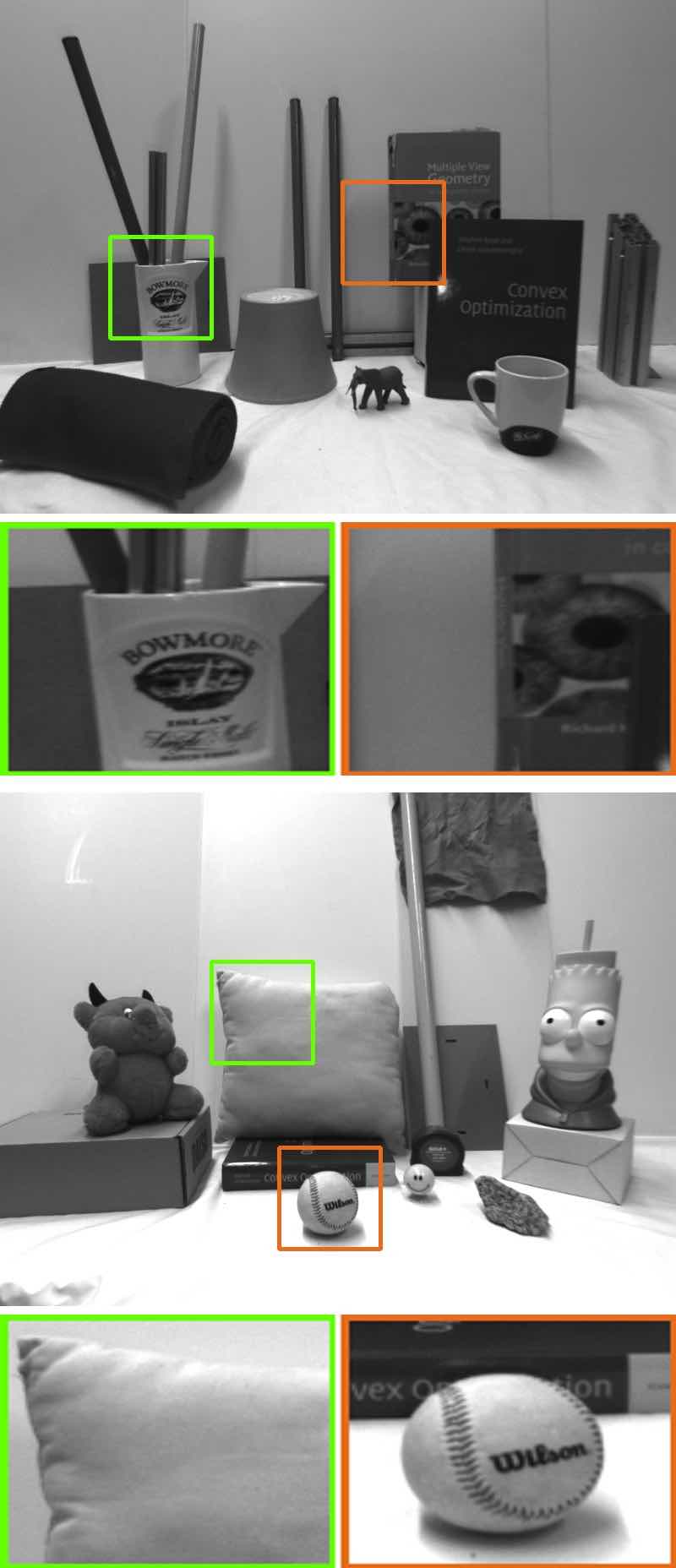} 
		\end{minipage}
		\hspace{-0.09in}
	}\subfigure[Target]{
    		\begin{minipage}[b]{0.1385\linewidth}
  		 	\includegraphics[width=1\linewidth]{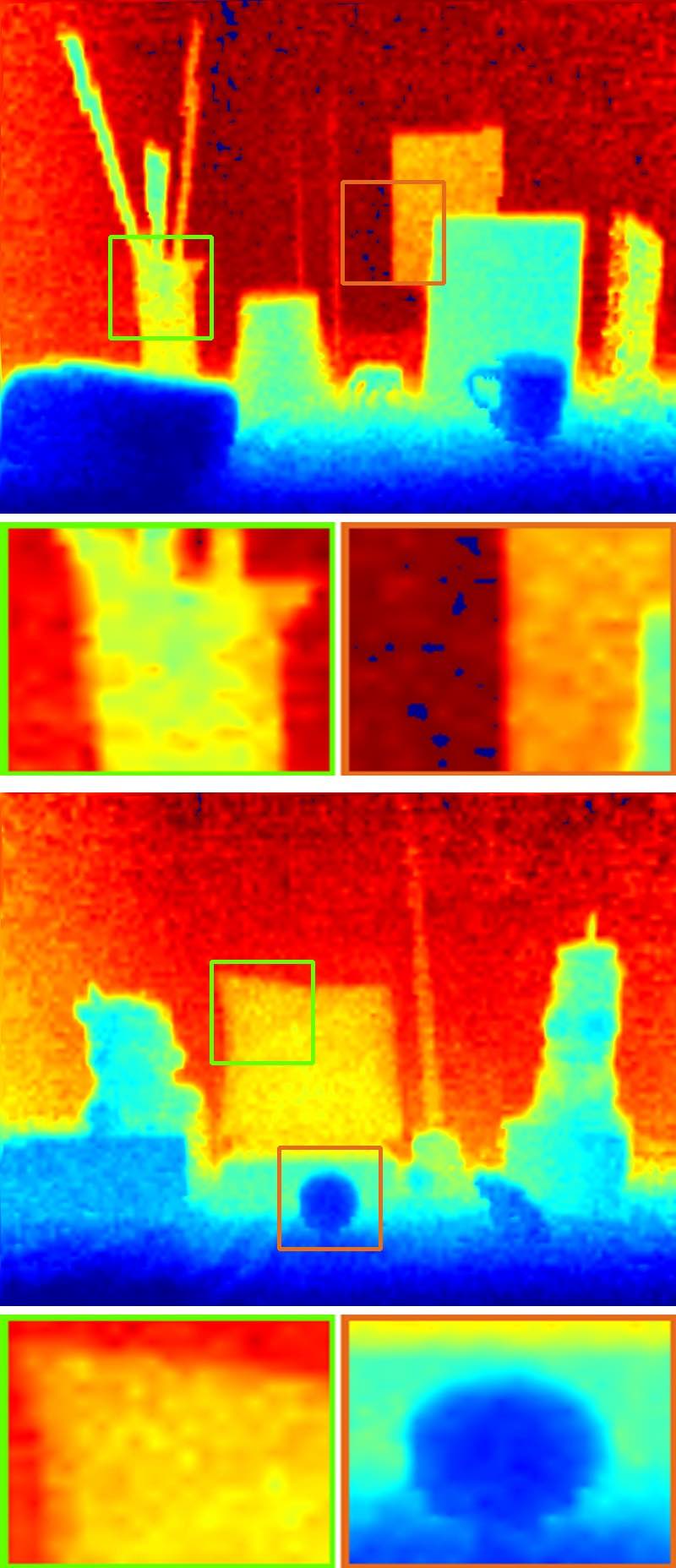}
    		\end{minipage}
    		\hspace{-0.09in}
    	}\subfigure[SDF]{
		\begin{minipage}[b]{0.1385\linewidth}
			\includegraphics[width=1\linewidth]{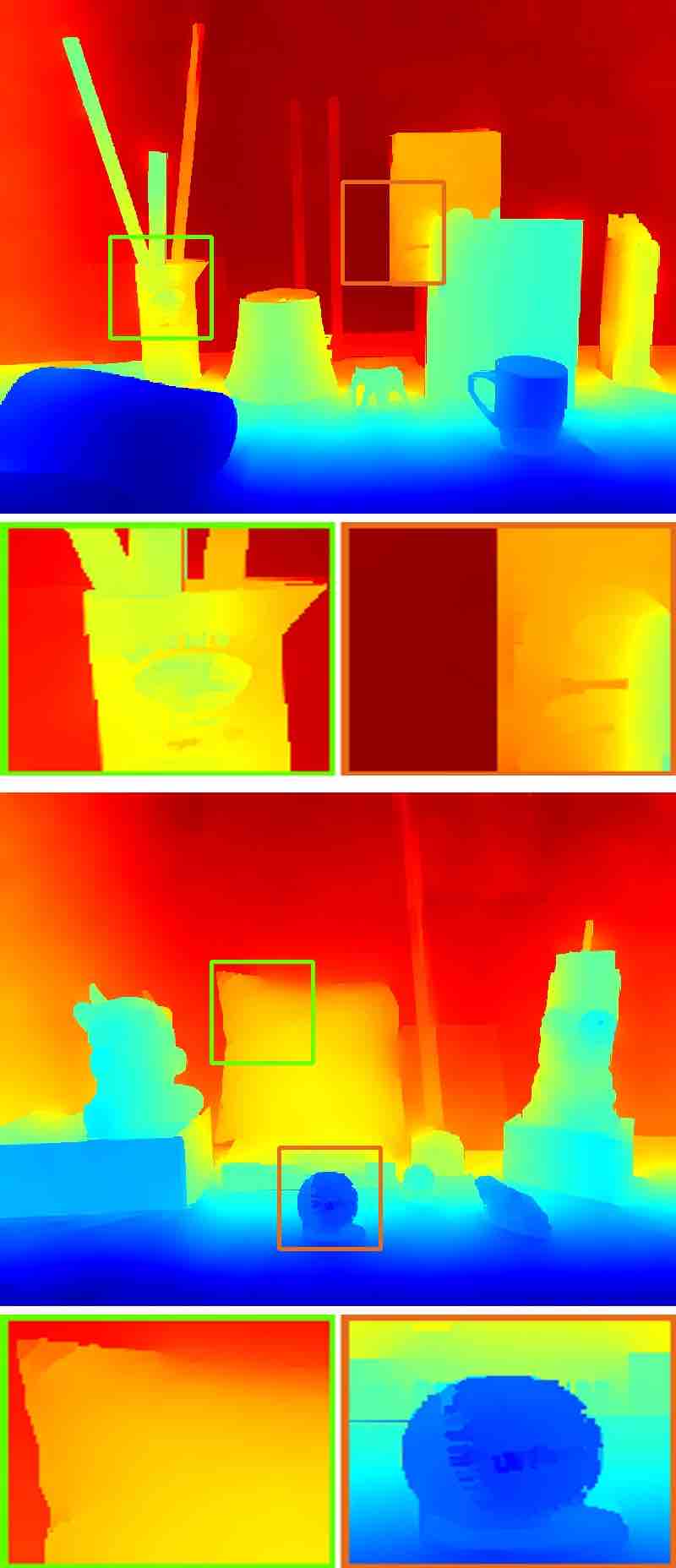} 
		\end{minipage}
		\hspace{-0.09in}
	}\subfigure[DGDIE]{
		\begin{minipage}[b]{0.1385\linewidth}
			\includegraphics[width=1\linewidth]{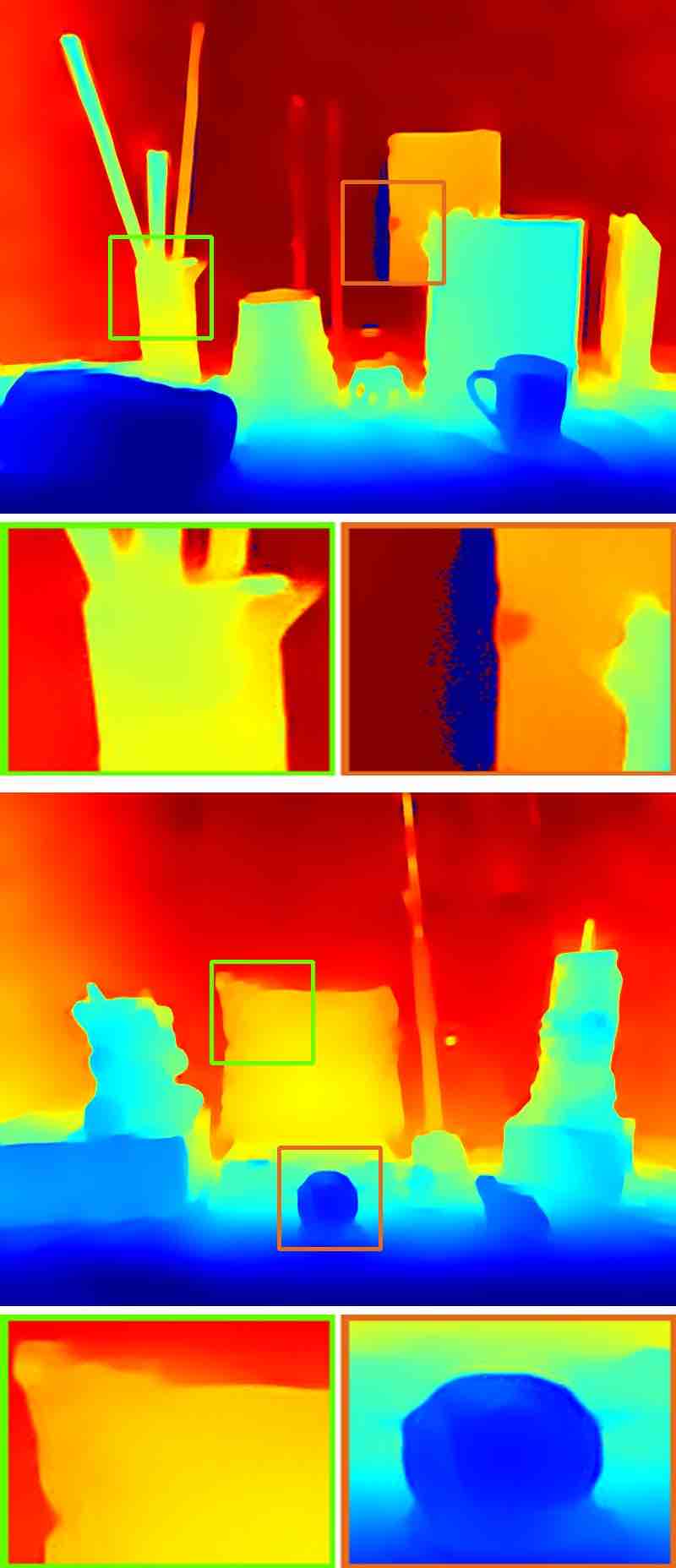} 
		\end{minipage}
		\hspace{-0.09in}
	}\subfigure[DKN]{
		\begin{minipage}[b]{0.1385\linewidth}
			\includegraphics[width=1\linewidth]{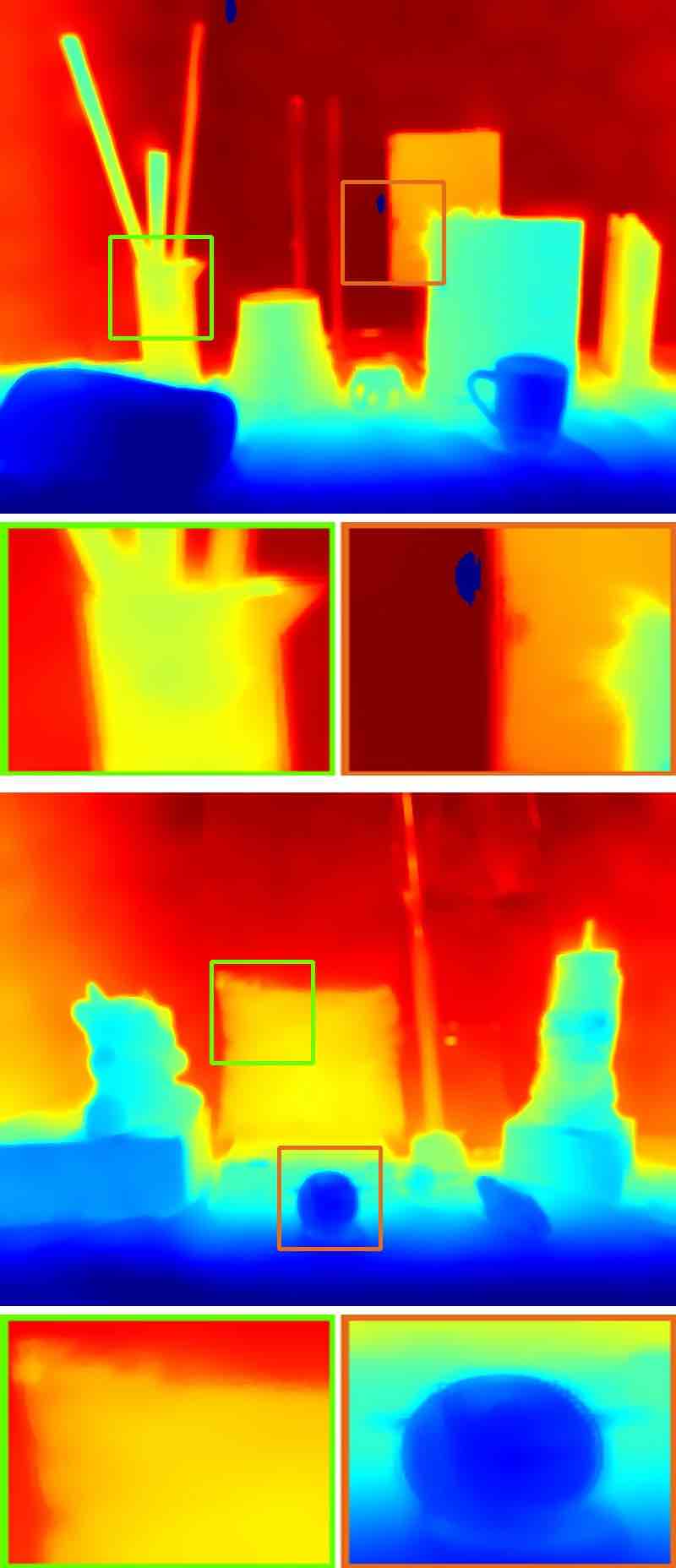} 
		\end{minipage}
		\hspace{-0.09in}
	}\subfigure[DAGF]{
		\begin{minipage}[b]{0.1385\linewidth}
			\includegraphics[width=1\linewidth]{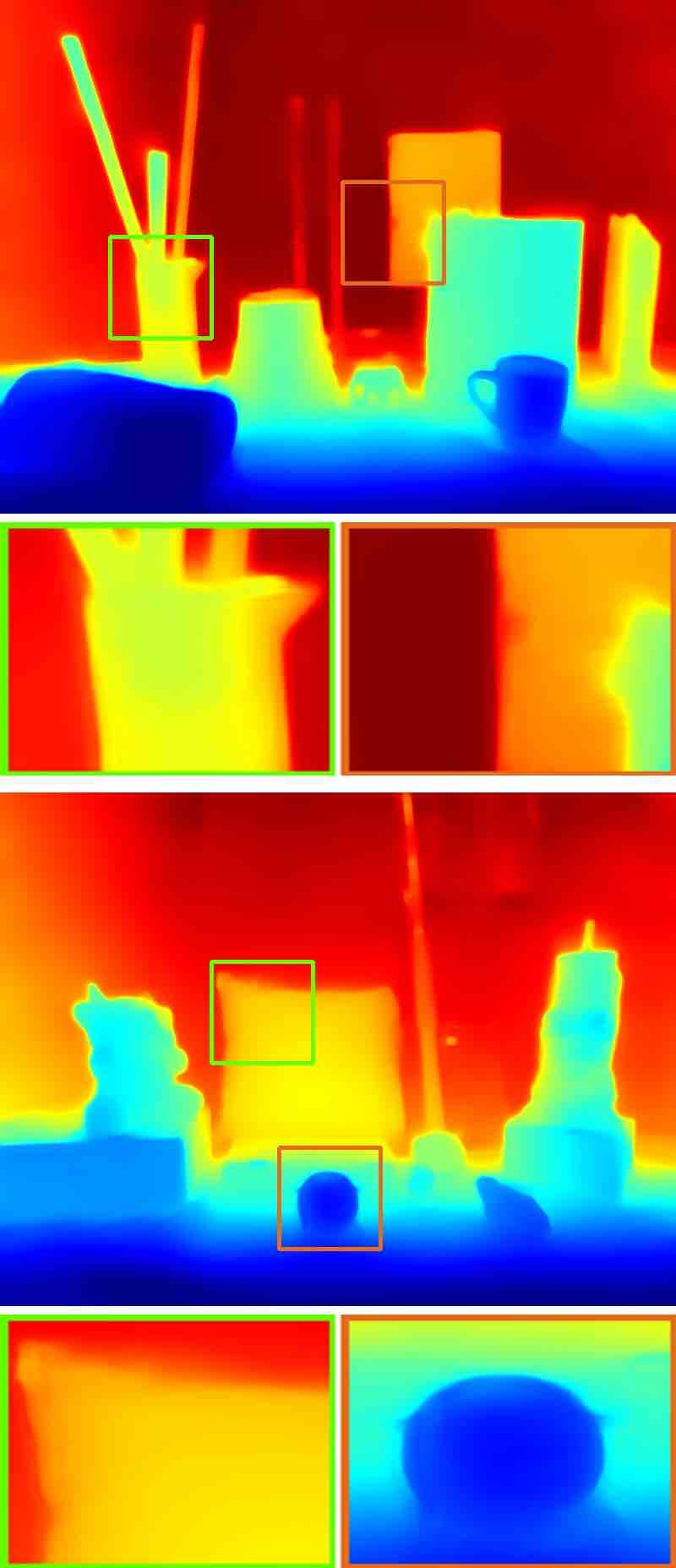} 
		\end{minipage}
		\hspace{-0.09in}
	}\subfigure[GT]{
		\begin{minipage}[b]{0.1385\linewidth}
			\includegraphics[width=1\linewidth]{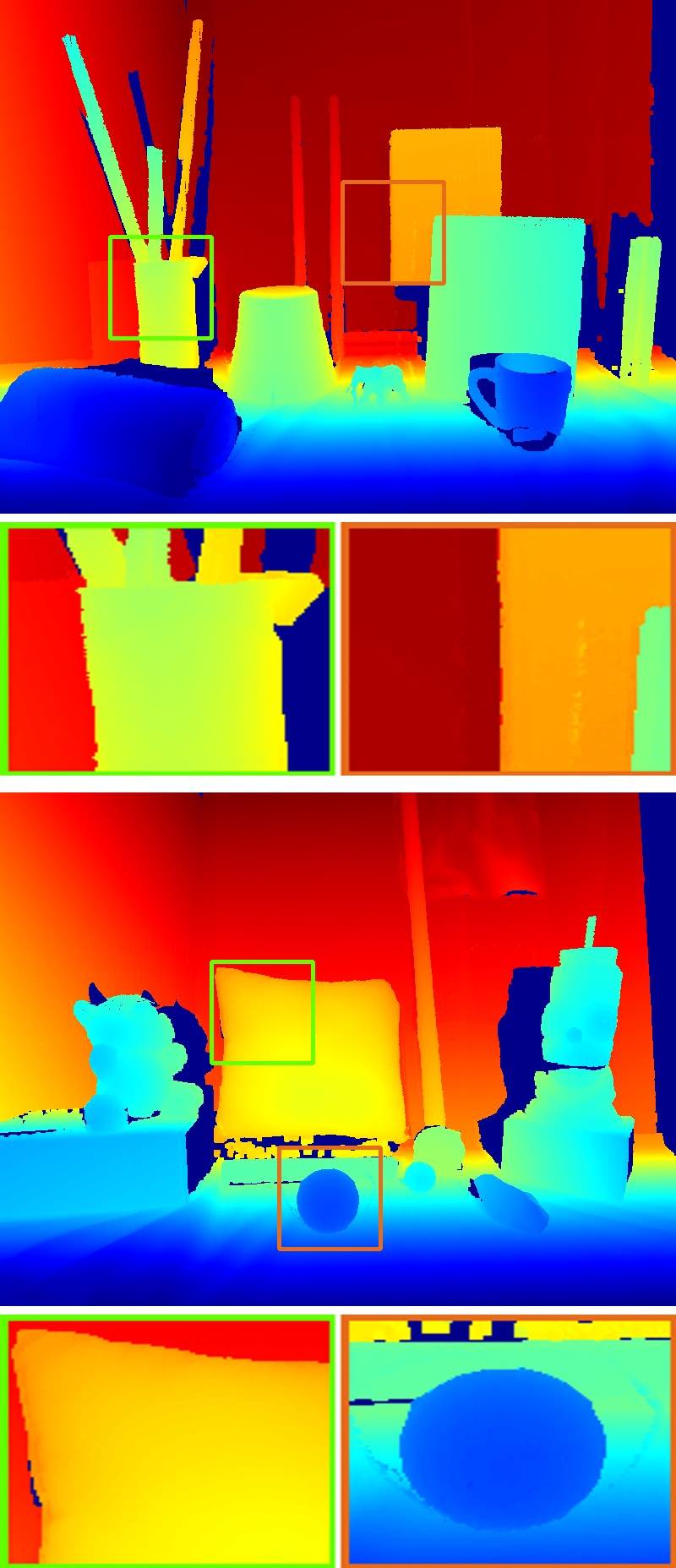} 
		\end{minipage}
		\hspace{-0.09in}
	}
	\end{center}
% 	\vspace{-0.1in}
	\caption{Visual comparison of realistic depth map super-resolution on two examples (\textit{books} and \textit{devil}) from ToFMark~\cite{TGV} dataset: (a): Guidance image, (b): Target image, (c): SDF~\cite{sdf}, (d): DGDIE~\cite{gu2017learning}, (e): DKN~\cite{DKN}, (f): DAGF, (e): Ground truth. Please enlarge the PDF for more details.}
	\label{fig:fig_real}
\end{figure*}
\begin{table*}[!tb]\setlength{\tabcolsep}{10pt}\renewcommand{\arraystretch}{1.2}
	\begin{center}
	\caption{\label{tab:jsd_result} Quantitative comparison for joint depth image super-resolution and denoising on four standard RGB/D datasets in terms of average RMSE values. Following the experimental setting of \cite{DJF, DKN}), we calculate the average RMSE values in cenrimeter for NYU v2~\cite{NYU} dataset. For other datasets, we compute the RMSE values by scaling the depth value to the range [0, 255]. The best performance for each case are highlighted in \textbf{boldface} while the second best ones are \underline{underscored}. For RMSE metric, the lower values mean the better performance.}
		\begin{tabular}{lccccccccccccccc}
		\toprule
		{Datasets} & \multicolumn{3}{c}{Middlebury} & \multicolumn{3}{c}{Lu} & \multicolumn{3}{c}{NYU v2} &  \multicolumn{3}{c}{Sintel} \\
		\cmidrule(lr){2-4} \cmidrule(lr){5-7} \cmidrule(lr){8-10} \cmidrule(lr){11-13} 
		{Method}  &$4\times$ & $8 \times$  & $16\times$  & $4\times$ & $8 \times$  & $16\times$  &$4\times$ & $8 \times$  & $16\times$ &$4 \times$ & $8 \times$  & $16\times$\\
		\midrule
		DGF~\cite{DGF} & 2.70 & 4.13 & 6.38 & 4.06 & 5.85 & 8.39 & 6.52 & 9.23 & 13.00 & 6.94 & 9.03 & 12.05 \\
		DJF~\cite{DJF} & 1.80 &  2.99 &  5.16 & 1.85 &  3.13 &  5.39 & 3.74 &  5.95 &  9.61 & 4.88 &  6.93 &  10.05 \\
		DMSG~\cite{DMSG} & 1.79 &  2.69 &  4.75 & 1.88 &  \underline{2.79} &  4.84 & 3.60 &  5.31 &  \underline{8.07} & 4.74 &  6.36 &  \underline{8.72} \\
		DJFR~\cite{DJFR} &1.86 &  3.07 &  5.27 & 1.91 &  3.21 &  5.51 & 4.01 &  6.21 &  9.90 & 5.10 &  7.12 &  10.23  \\
		DSRN~\cite{DepthSR} & 1.84 &  2.99 &  4.70 & 1.97 &  2.98 &  5.94 & 4.36 &  6.31 &  9.75 & 5.49 &  7.21 &  9.80 \\
		PAC~\cite{PanNet} & 1.81 & 2.94 & 5.08 & 1.93 & 3.44 & 6.18 & 4.23 & 6.24 & 9.54 & 5.40 & 7.32 & 9.89 \\
		DKN~\cite{DKN} &\underline{1.76} &  \underline{2.68} &  \underline{4.55} & \underline{1.81} &  2.82 &  \underline{4.81} & \underline{3.39} &  \underline{5.24} &  8.41 & \underline{4.51} &  \underline{6.25} &  9.20 \\
		DAGF (Ours) & \textbf{1.72} &  \textbf{2.61} &  \textbf{4.24} & \textbf{1.74} &  \textbf{2.72} &  \textbf{4.51} & \textbf{3.25} &  \textbf{5.01} & \textbf{7.54} & \textbf{4.42} &  \textbf{6.09} & \textbf{8.25} \\
		\bottomrule
		\end{tabular}
	\end{center}
\end{table*}
\begin{table}[!htb]\setlength{\tabcolsep}{12pt}\renewcommand{\arraystretch}{1.2}
    \begin{center}
    \caption{\label{tab:tab_real}Quantitative comparison for realistic depth image super-resolution in terms of RMSE values on the ToFMark~\cite{TGV} dataset. The best performance for each case are highlighted in \textbf{boldface} while the second ones are \underline{underscored}.}
         \begin{tabular}{lccc}
        \toprule
        Methods & Books & Devil & Shark \\
            \midrule
        Bilinear & 17.10 & 20.17 & 18.66 \\
        JBU~\cite{JBU} & 16.03 & 18.79 & 27.57 \\
        GF~\cite{GF} & 15.74 & 18.21 & 27.04 \\
        TGV~\cite{TGV} & 12.36 & 15.29 & 14.68 \\
        SDF~\cite{sdf}  & 12.66 & 14.33 & 10.68 \\
        Yang~\cite{2015Color} & 12.25 & 14.71 & 13.83 \\
        DGDIE~\cite{gu2017learning} & 12.32 & 14.06 & 9.66 \\
        DKN~\cite{DKN} & \underline{11.81} & \underline{13.54} & \underline{9.11} \\
        DAGF (Ours) & \textbf{11.80} & \textbf{13.47} & \textbf{9.07} \\
    \bottomrule
    \end{tabular}
    \end{center}
\end{table}

\subsection{Cross-modality Image Restoration}
\label{cm}
For the task of cross-modality image restoration, we first conduct experiments on joint depth image super-resolution and denoising to show the superiority of the proposed method. Moreover, to verify the ability of the proposed method on dealing with various visual domains, we apply the trained models on two noise reduction tasks using flash/non-flash and RGB/NIR image pairs. Finally, we conduct experiments on ToF Mark dataset~\cite{Ferstl2013Image}. It contains three real world depth images acquired by Time of Flight (ToF) camera, which have complicated multi-modality degradation

\textbf{Joint Depth Image Super-resolution and Denoising}. Depth images acquired by ranging sensors are typically noisy. In order to simulate the data acquisition process of the depth sensor, we add Gaussian noise with variance as 25 to the low-resolution target depth images. We use the same experimental settings as the task of GSR in Sect~\ref{gsr} to train our model. 
We compare our method with ten state-of-the-art methods, including GF~\cite{GF}, MUF~\cite{mugf} and SDF~\cite{sdf}, which are traditional model-based methods; and DGF~\cite{DGF}, DJF~\cite{DJF}, DMSG~\cite{DMSG}, DJFR~\cite{DJFR}, DSRN~\cite{DepthSR}, PAC~\cite{PanNet}, DKN~\cite{DKN}, which are deep learning-based methods. Since most of the existing methods do not provide experimental results for this task, we retrain all the deep learning-based methods with the same training and test dataset as ours.

The quantitative results in terms of RMSE values for four benchmark datasets are reported in Table~\ref{tab:jsd_result}, from which we can see that the proposed method can obtain consistently better results than the existing state-of-the-art methods, especially for the $8\times$ and $16\times$ cases which are more difficult to recover. This is mainly because that: 1) we employ a pyramid architecture to extract multi-modality features for guided kernel generation, thus the multi-scale complementary information can be obtained; 2) for guided image filtering, we leverage the coarse-to-fine strategy to filter the low-resolution target image and thus the structure details can be progressively recovered; 3) compared to single loss at the end of network, the proposed multi-scale loss can bring stronger supervision to our model.

Fig.~\ref{fig:fig_noisy} further demonstrates the visual superiority of the proposed method for joint depth image super-resolution and denoising ($16\times$ Bicubic downsampling and Gaussian noise). The results of GF~\cite{GF}, MUF~\cite{mugf} and SDF~\cite{sdf}  still contain much noise, and the visual quality of the whole image is poor. This is due to that these methods are based on the locally linear assumption and they employ the mean filter to calculate the coefficients for pixel-wise linear representations. The methods of PAC~\cite{PanNet} and DJFR~\cite{DJFR} can remove noise well, while they cannot preserve the sharp edge and introduce ringing artifacts. The results of DKN are clearer and sharper than previous methods. However, they suffer from color distortion, which attributes to the batch normalization used in DKN~\cite{DKN}. In contrast, our method is able to remove the noise effectively and produces the clearest and sharpest boundaries.

\begin{figure*}[!tb]
	\begin{center}
		\subfigure[Guidance]{
		\begin{minipage}[b]{0.1385\linewidth}
			\includegraphics[width=1\linewidth]{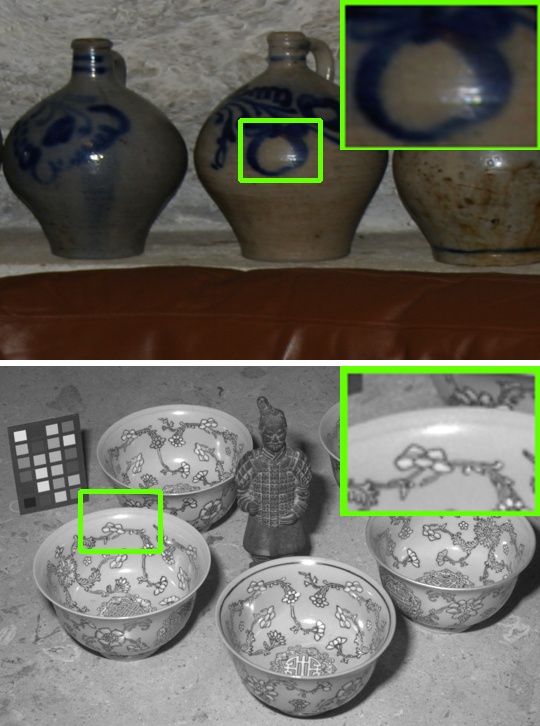} 
		\end{minipage}
		\hspace{-0.09in}
	}\subfigure[Target]{
    		\begin{minipage}[b]{0.1385\linewidth}
  		 	\includegraphics[width=1\linewidth]{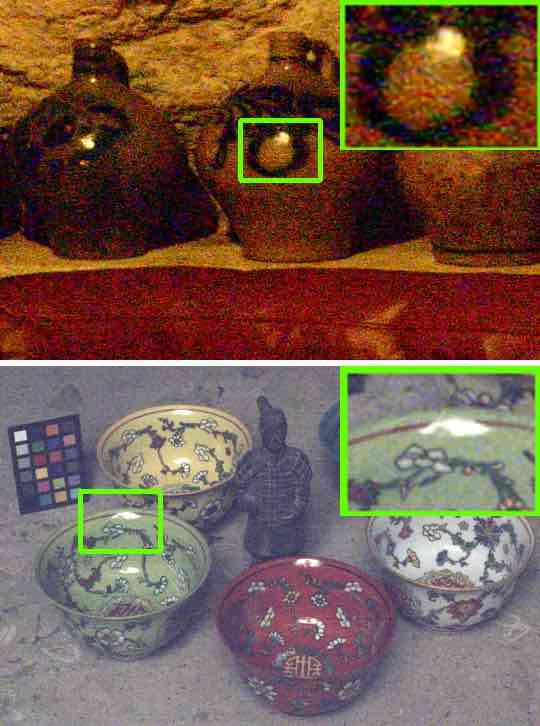}
    		\end{minipage}
    		\hspace{-0.09in}
    	}\subfigure[SDF]{
		\begin{minipage}[b]{0.1385\linewidth}
			\includegraphics[width=1\linewidth]{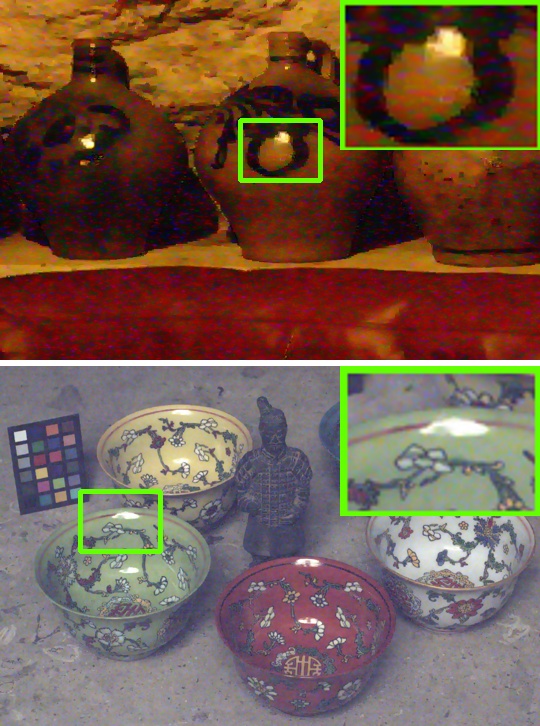} 
		\end{minipage}
		\hspace{-0.09in}
	}\subfigure[RTV]{
		\begin{minipage}[b]{0.1385\linewidth}
			\includegraphics[width=1\linewidth]{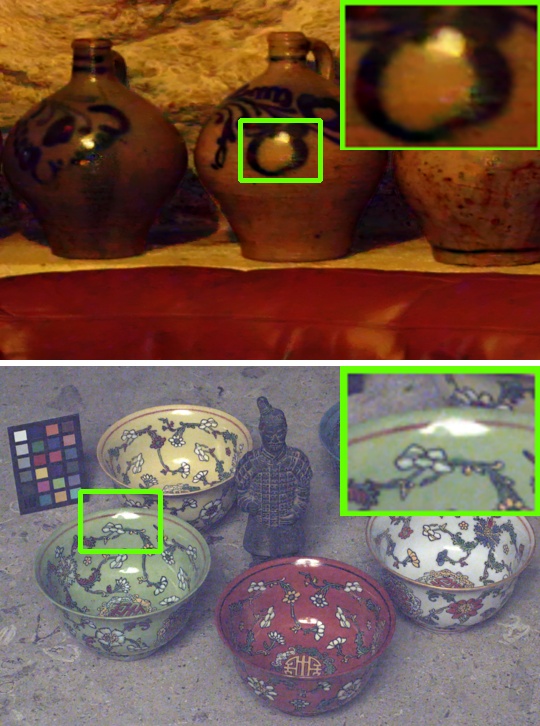} 
		\end{minipage}
		\hspace{-0.09in}
	}\subfigure[DJFR]{
		\begin{minipage}[b]{0.1385\linewidth}
			\includegraphics[width=1\linewidth]{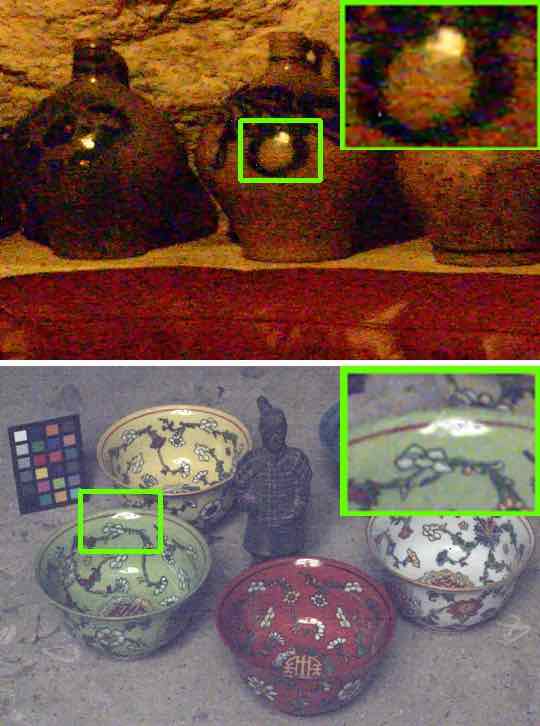} 
		\end{minipage}
		\hspace{-0.09in}
	}\subfigure[DKN]{
		\begin{minipage}[b]{0.1385\linewidth}
			\includegraphics[width=1\linewidth]{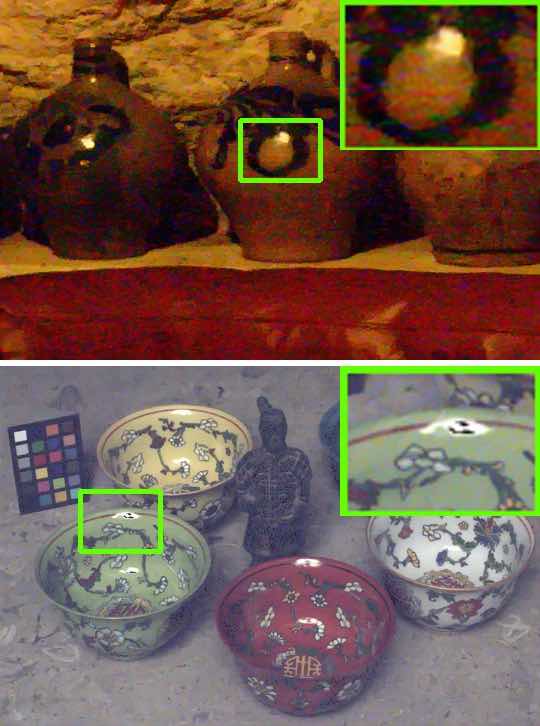} 
		\end{minipage}
		\hspace{-0.09in}
	}\subfigure[DAGF]{
		\begin{minipage}[b]{0.1385\linewidth}
			\includegraphics[width=1\linewidth]{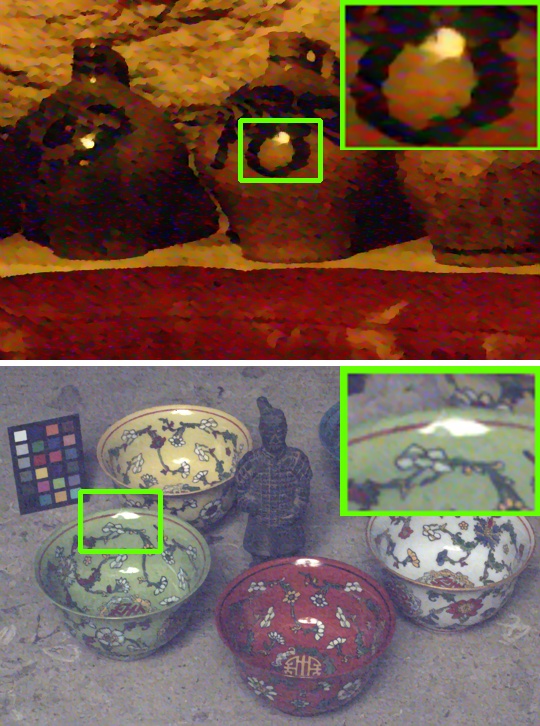} 
		\end{minipage}
		\hspace{-0.09in}
	}
	\end{center}
% 		\vspace{-0.1in}
	\caption{Visual comparison of cross-modality image restoration. Top: flash guided non-flash image denoising. Bottom: NIR guided color image denoising. (a): Guidance image, (b): Target image, (c): SDF~\cite{sdf}, (d): RTV~\cite{RTV}, (e): DJFR~\cite{DJFR}, (f): DKN~\cite{DKN}, (f): DAGF. Please enlarge the PDF for more details.}
	\label{fig:flash}
\end{figure*}
\begin{figure*}[!htb]
	\begin{center}
		\subfigure[Target]{
		\begin{minipage}[b]{0.1385\linewidth}
			\includegraphics[width=1\linewidth]{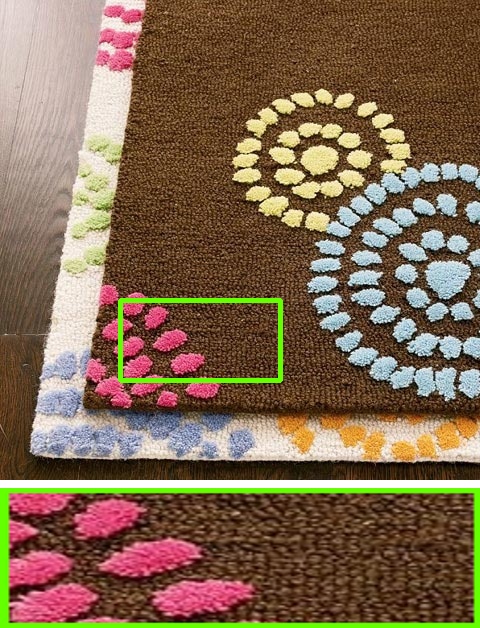} 
		\end{minipage}
		\hspace{-0.09in}
	}\subfigure[RTV]{
    		\begin{minipage}[b]{0.1385\linewidth}
  		 	\includegraphics[width=1\linewidth]{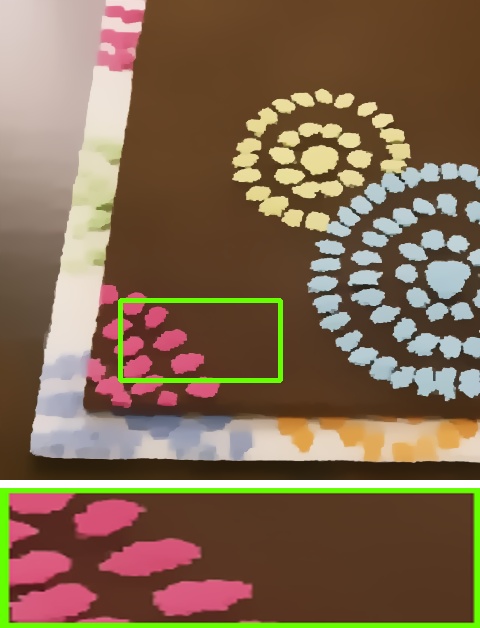}
    		\end{minipage}
    		\hspace{-0.09in}
    	}\subfigure[RGF]{
		\begin{minipage}[b]{0.1385\linewidth}
			\includegraphics[width=1\linewidth]{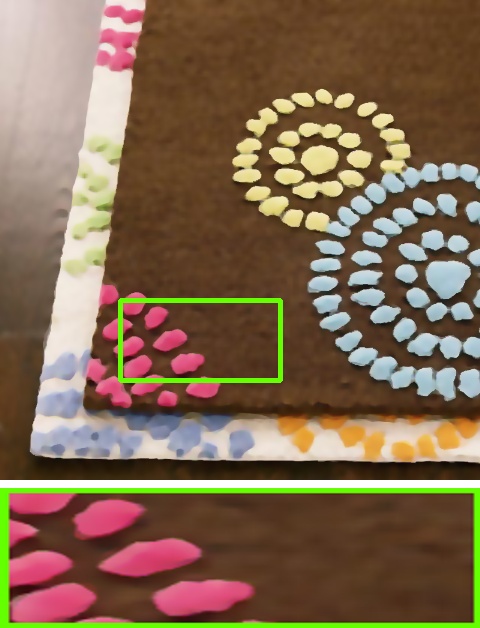} 
		\end{minipage}
		\hspace{-0.09in}
	}\subfigure[SDF]{
		\begin{minipage}[b]{0.1385\linewidth}
			\includegraphics[width=1\linewidth]{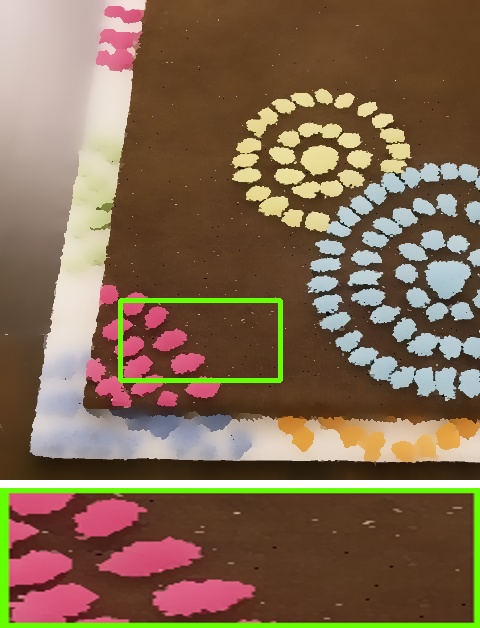} 
		\end{minipage}
		\hspace{-0.09in}
	}\subfigure[DJFR]{
		\begin{minipage}[b]{0.1385\linewidth}
			\includegraphics[width=1\linewidth]{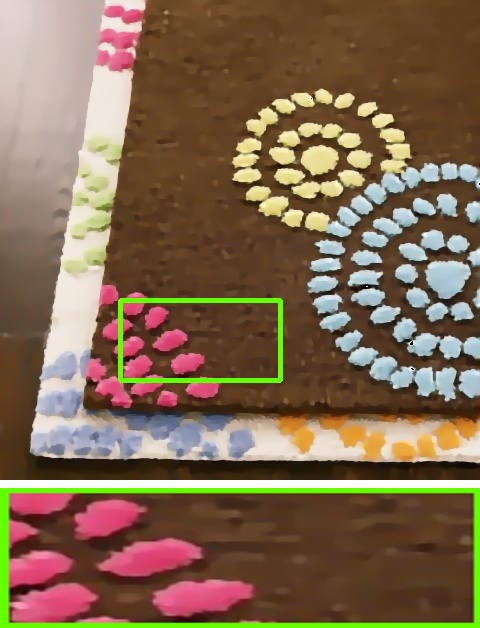} 
		\end{minipage}
		\hspace{-0.09in}
	}\subfigure[DKN]{
		\begin{minipage}[b]{0.1385\linewidth}
			\includegraphics[width=1\linewidth]{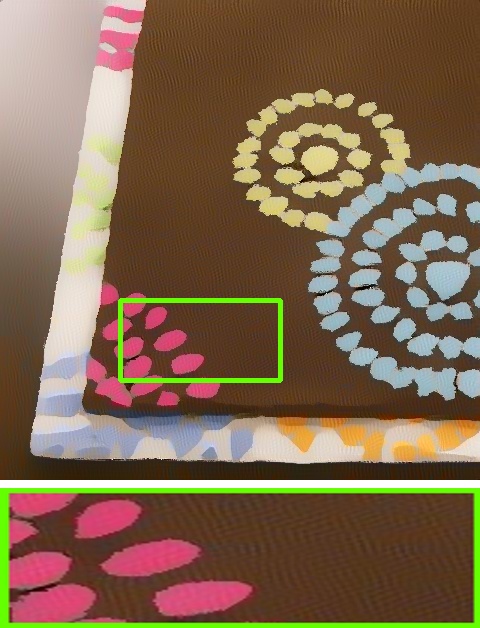} 
		\end{minipage}
		\hspace{-0.09in}
	}\subfigure[DAGF]{
		\begin{minipage}[b]{0.1385\linewidth}
			\includegraphics[width=1\linewidth]{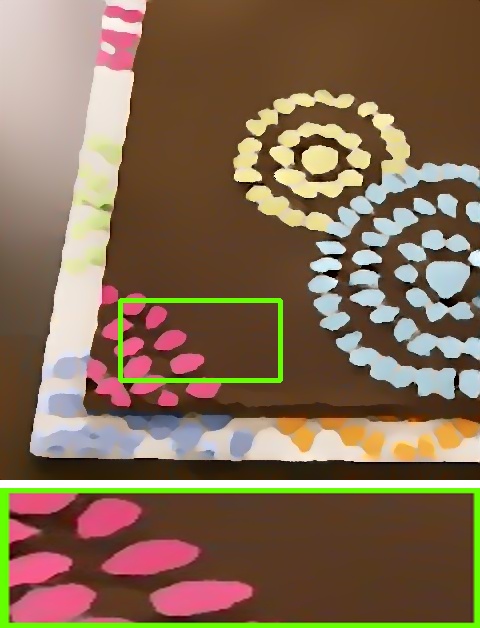} 
		\end{minipage}
	}
	\end{center}
% 	\vspace{-0.1in}
	\caption{Visual comparisons of texture remove results. (a): Target image, (b): RTV~\cite{RTV}, (c): RGF~\cite{RGF}, (d): SDF~\cite{sdf}, (e): DJFR~\cite{DJFR}, (f): DKN~\cite{DKN}, (g): DAGF. Please enlarge the PDF for more details.}
	\label{fig:tr}
\end{figure*}

\textbf{Cross-modality Image Restoration}. We further demonstrate that our model trained for depth image denoising can be generalized to address other cross-modality image restoration tasks, such as flash guided non-flash image denoising and NIR guided color image restroation. Fig.~\ref{fig:flash} shows the visual comparison among existing state-of-the-art methods and ours. All of the deep learning-based methods (\eg\  DJFR~\cite{DJFR} and DKN~\cite{DKN}) are tested with the same setting as ours. Among the compared methods, SDF~\cite{sdf}  and RTV~\cite{RTV} are specially designed for this task. As can be seen from Fig.~\ref{fig:flash}, DJFR~\cite{DJFR} cannot remove noise, and the results of DKN~\cite{DKN} suffer from halo artifacts. On the contrary, the proposed DAGF can produce more convincing results with less artifact. The method of RTV~\cite{RTV} which is specially designed for this task, obtains the best performance.

\begin{table}[!tb]\setlength{\tabcolsep}{25pt}\renewcommand{\arraystretch}{1.2}
    \begin{center}
         \caption{\label{tab:tab_seg}Quantitative comparison for semantic segmentation in terms of average IoU on the validation set of Pascal VOC 2012. The best performance for each case are highlighted in \textbf{boldface} while the second ones are \underline{underscored}.}
         \begin{tabular}{lc}
        \toprule
        Methods & Mean IoU \\
            \midrule
            Deeplab-V2~\cite{DeepLab}& 70.69  \\
            DenseCRF~\cite{CRF} & 71.98 \\
            DGF~\cite{DGF} & 72.96 \\
            DJFR~\cite{DJFR} & 73.30  \\
            FDKN~\cite{DKN} & \underline{73.60} \\
            DAGF (Ours) & \textbf{73.76} \\
    \bottomrule
    \end{tabular}
    \end{center}
\end{table}
\textbf{Realistic Depth Image Super-resolution}. To further evaluate the robustness of the proposed method, we conduct experiments on ToFmark dataset~ \cite{Ferstl2013Image}, which include real ToF sensor data and thus have complicated multi-modality degradation. Following the experimental protocol of DGDIE~\cite{gu2017learning}, we first perform image completion on the acquired depth images and then send them to our model ($4\times$ super-resolution and denoising) trained on NYU v2 dataset~\cite{NYU} to obtain the final results. We compare our method with a recently proposed deep learning-based method (\eg\ DKN~\cite{DKN}) and some traditional methods (\eg\ TGV~\cite{TGV}, SDF~\cite{sdf}, DGDIE~\cite{gu2017learning}). As shown in Table~\ref{tab:tab_real}, our method constantly obtains the best objective results for the three test images. Fig.~\ref{fig:fig_real} presents visual comparison results for two images (\textit{books} and \textit{devil}). Form these figures, it is easy to observe that the results of SDF~\cite{sdf} suffer from texture-copying artifacts. The results of DKN~\cite{DKN} are smooth and blurred, since DKN generates filter kernels without considering the inconsistence between color and depth image. The results of DGDIE~\cite{gu2017learning} are clear but they deviate from the ground truth. By comparison, the results of the proposed method are sharper and much closer to the ground truth, especially at the boundary regions.
\begin{figure*}[!htb]
	\begin{center}
		\subfigure[RGB Image]{
		\begin{minipage}[b]{0.155\linewidth}
			\includegraphics[width=1\linewidth]{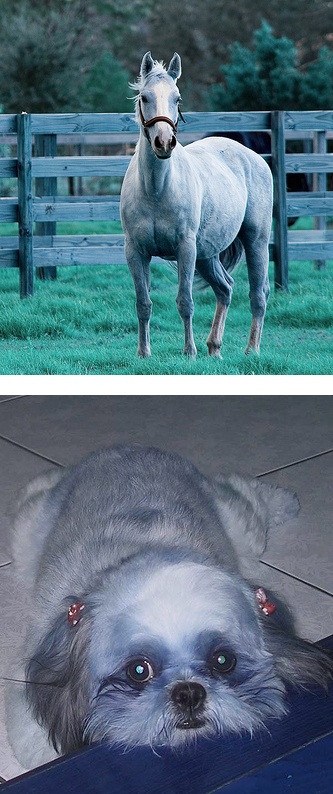} 
		\end{minipage}
	}\subfigure[Deeplab-V2]{
    		\begin{minipage}[b]{0.155\linewidth}
  		 	\includegraphics[width=1\linewidth]{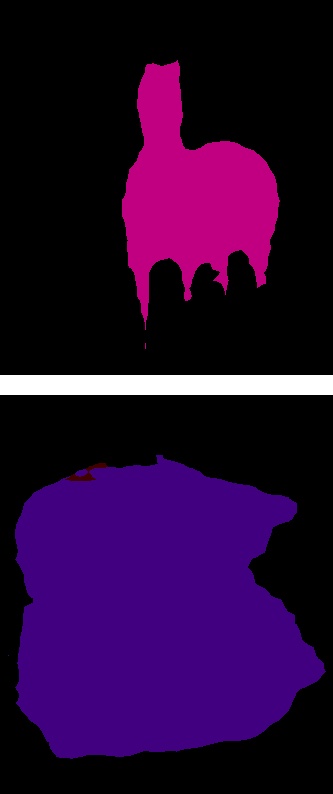}
    		\end{minipage}
    	}\subfigure[DGF]{
		\begin{minipage}[b]{0.155\linewidth}
			\includegraphics[width=1\linewidth]{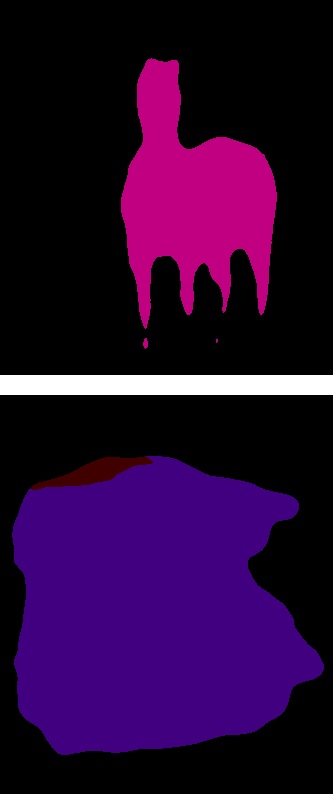} 
		\end{minipage}
	}\subfigure[FDKN]{
		\begin{minipage}[b]{0.155\linewidth}
			\includegraphics[width=1\linewidth]{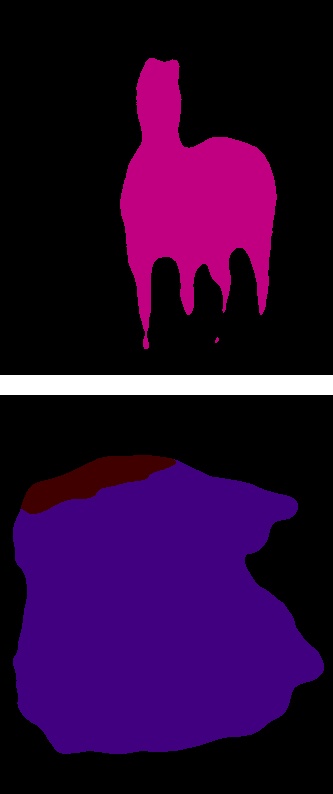} 
		\end{minipage}
	}\subfigure[DAGF]{
		\begin{minipage}[b]{0.155\linewidth}
			\includegraphics[width=1\linewidth]{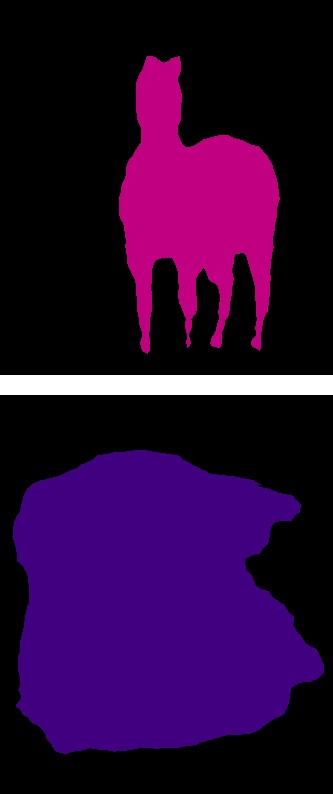} 
		\end{minipage}
	}\subfigure[GT]{
		\begin{minipage}[b]{0.155\linewidth}
			\includegraphics[width=1\linewidth]{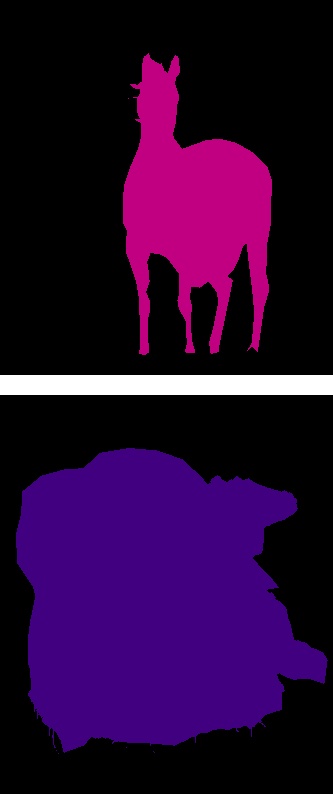} 
		\end{minipage}
	}
	\end{center}
% 	\vspace{-0.1in}
	\caption{Visual comparison of semantic segmentation on the validation set of Pascal VOC 2012 dataset~\cite{VOC}. (a): RGB image, (b): Deeplab-V2~\cite{DeepLab}, (c): DGF~\cite{DGF}, (d): FDKN~\cite{DKN}, (e): DAGF, (f): ground truth image.  Please enlarge the PDF for more details.}
	\label{fig:seg_fig}
\end{figure*}
\subsection{Texture Removal}
\label{tr}
Texture removal is the task of extracting semantically meaningful structures from textured surfaces.  For this task, we use the textured image itself as the guidance, and apply our model trained for depth image denoising iteratively to remove small-scale textures. We compare our method with RTV~\cite{RTV}, RGF~\cite{RGF}, SDF~\cite{sdf}, DJFR~\cite{DJFR} and DKN~\cite{DKN}. For deep learning-based methods, we follow DKN~\cite{DKN}, set the number of iterations as 4, and for other methods we carefully fine-tune the parameters to provide the best results. The visual comparison are presented in Fig.~\ref{fig:tr}. Obviously, our method outperforms other compared methods, and it can painlessly remove small-scale textures as well as preserve the global color variation and main edges.

\subsection{Semantic Segmentation}
\label{ss}
Semantic segmentation is a fundamental computer vision task, which aims at assigning pre-defined labels to each pixel of an image. In DGF~\cite{DGF}, the author proposed to use guided image filtering as a layer to replace the time-consuming fully connected conditional random filed (CFR)~\cite{CRF} for semantic segmentation. We demonstrate that the proposed DAGF can be applied to this problem. Following DGF~\cite{DGF}, we plug the proposed model into DeepLab-v2~\cite{DeepLab} and train the whole network in an end-to-end manner, and thus the offline post-processing of CRFs can be avoided. We utilize the Pascal VOC 2012 dataset~\cite{VOC} in our experiment, which contains 1264, 1229 and 1456 images for training, validation and testing, respectively. Similar to DGF~\cite{DGF}, we augment the training set with the annotations provided by~\cite{2011Semantic}, resulting in 10582 images. The 1449 images in the validation set are employed to evaluated the proposed method. 

We use the mean intersection-over-union (IoU) score as evaluation metric and report the quantitative results for the validation set of Pacal VOC dataset~\cite{VOC} in Table~\ref{tab:tab_seg}. The baseline denotes DeepLab-v2~\cite{DeepLab} without CRF. As can be seen from this table, our method outperforms the second best model DKN~\cite{DKN} by $0.16\%$ mIoU and other models by a large margin. We visualize the segmentation results among our method and other compared methods in Fig~\ref{fig:seg_fig}, from which we can see that our method is capable of generating results with accurate and complete object boundaries.

\section{Ablation Study}
\label{abl}
\begin{table}[!tb]\setlength{\tabcolsep}{10pt}\renewcommand{\arraystretch}{1.2}
    \begin{center}
        \caption{\label{tab:tab_kernel}\textbf{Ablation study}. Quantitative comparison of different size of kernel ($k\times k$) and the number of pyramid level ($m$).}
        \begin{tabular}{lcccc}
        \toprule
        Kernel Size & $m = 1$ & $m = 2$ & $m = 3$  & $m = 4$  \\
        \midrule
        $1 \times 1$ & 12.8 & 11.72 & 11.41 & 11.26 \\
        $3 \times 3$ & 8.97 & 8.12 & 7.81 & 7.75 \\
        $5 \times 5$ & 8.60 & 8.02 & 7.73 & 7.98\\
        $7 \times 7$ & 8.67 & 7.94 & 7.78 & 7.99\\
    \bottomrule
    \end{tabular}
    \end{center}
    We report average RMSE values of the last 449 image pairs in NYU v2 dataset~\cite{NYU}.
\end{table}
\begin{table*}[!htb]\setlength{\tabcolsep}{7.5pt}\renewcommand{\arraystretch}{1.2}
    \begin{center}
        \caption{\label{tab_abl} \textbf{Ablation Study}. Quantitative comparison of different components for $16\times$ depth image super-resolution. We chose RMSE as the evaluation metric, and the lower values indicate better performance. Model7 is our final model (DAGF).}
        \begin{tabular}{lcccccccccccccc}
        \toprule
         \multirow{2}{*}{Model} & \multicolumn{2}{l}{Kernel Generation}  & \multicolumn{3}{l}{Kernel Combination} & \multirow{2}{*}{$\mathcal{L}_{ms}$}& \multirow{2}{*}{$\mathcal{L}_{ba}$} &  \multirow{2}{*}{Middlebury}  & \multirow{2}{*}{Lu}  & \multirow{2}{*}{NYU v2} & \multirow{2}{*}{Sintel} & \multirow{2}{*}{Average} \\
        \cmidrule(lr){2-3} \cmidrule(lr){4-6}
        ~ & Target & Guidance & MUL & SUM & AKL & ~& ~ &   \\ 
        \midrule
       Model1 & $\checkmark$ &  & & ~& & & & 7.08& 7.87& 11.99& 13.67 & 10.15 \\
       Model2 &~ & $\checkmark$ & & ~& & &  & 5.68& 7.19& 9.09& 11.82 & 8.45 \\
       Model3  &  $\checkmark$ & $\checkmark$ & $\checkmark$ & & & & & 5.47 & 6.84 & 9.07 & 11.74 & 8.28\\ 
       Model4  &  $\checkmark$ & $\checkmark$ & ~ & $\checkmark$ &  & && 5.36& 6.90&  8.99& 11.65 & 8.23\\ 
       Model5 &  $\checkmark$ & $\checkmark$ & ~ & ~ &  $\checkmark$ & & &  5.06& 6.57& 8.49& 11.18 & 7.82 \\ 
       Model6 &  $\checkmark$ & $\checkmark$ & ~ & ~ &  $\checkmark$ & $\checkmark$ & & 4.88 & 6.19& 7.92& 10.89 & 7.47\\ 
       Model7 &  $\checkmark$ & $\checkmark$ & ~ & ~ &  $\checkmark$ & $\checkmark$ & $\checkmark$ & \textbf{4.75} & \textbf{6.16}& \textbf{7.81} & \textbf{10.64} & \textbf{7.34} \\ 
    \bottomrule
    \end{tabular}
    \end{center}
\end{table*}
\begin{figure*}[!htb]
	\begin{center}
		\subfigure[Guidance]{
		\begin{minipage}[b]{0.16\linewidth}
			\includegraphics[width=1\linewidth]{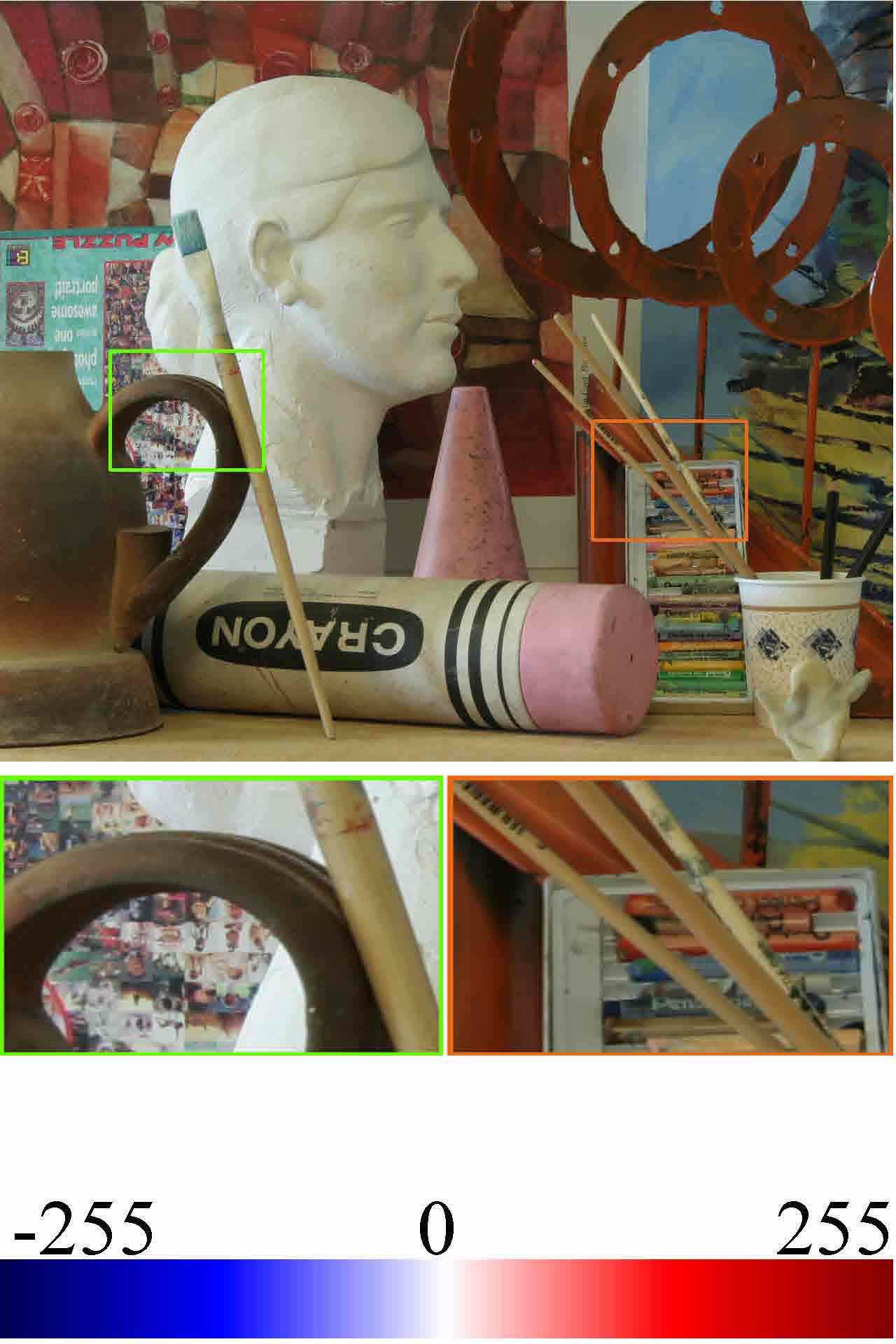} 
		\end{minipage}
		\hspace{-0.09in}
	}\subfigure[Model1]{
    		\begin{minipage}[b]{0.16\linewidth}
  		 	\includegraphics[width=1\linewidth]{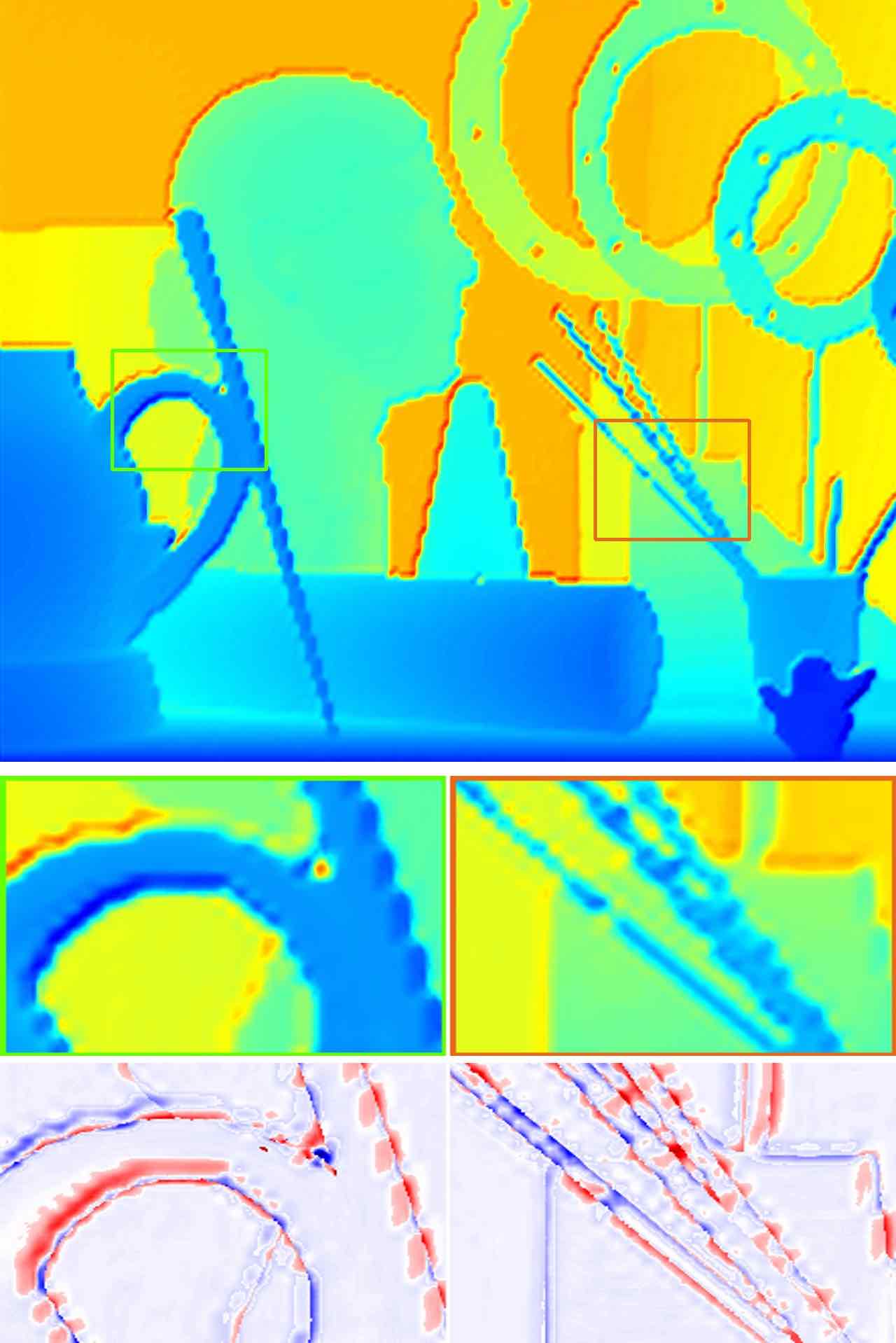}
    		\end{minipage}
    		\hspace{-0.09in}
    	}\subfigure[Model2]{
		\begin{minipage}[b]{0.16\linewidth}
			\includegraphics[width=1\linewidth]{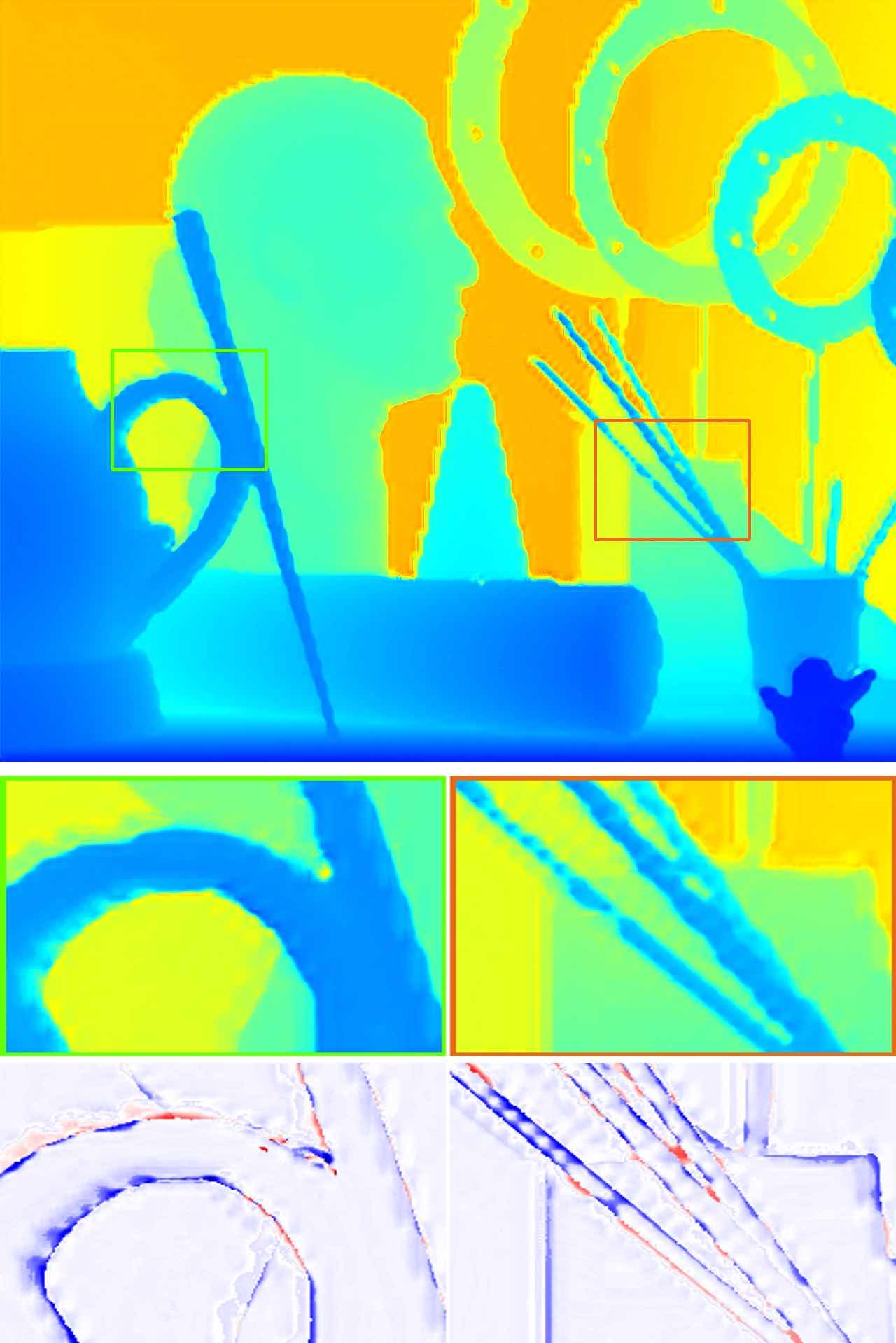} 
		\end{minipage}
		\hspace{-0.09in}
	}\subfigure[Model3]{
		\begin{minipage}[b]{0.16\linewidth}
			\includegraphics[width=1\linewidth]{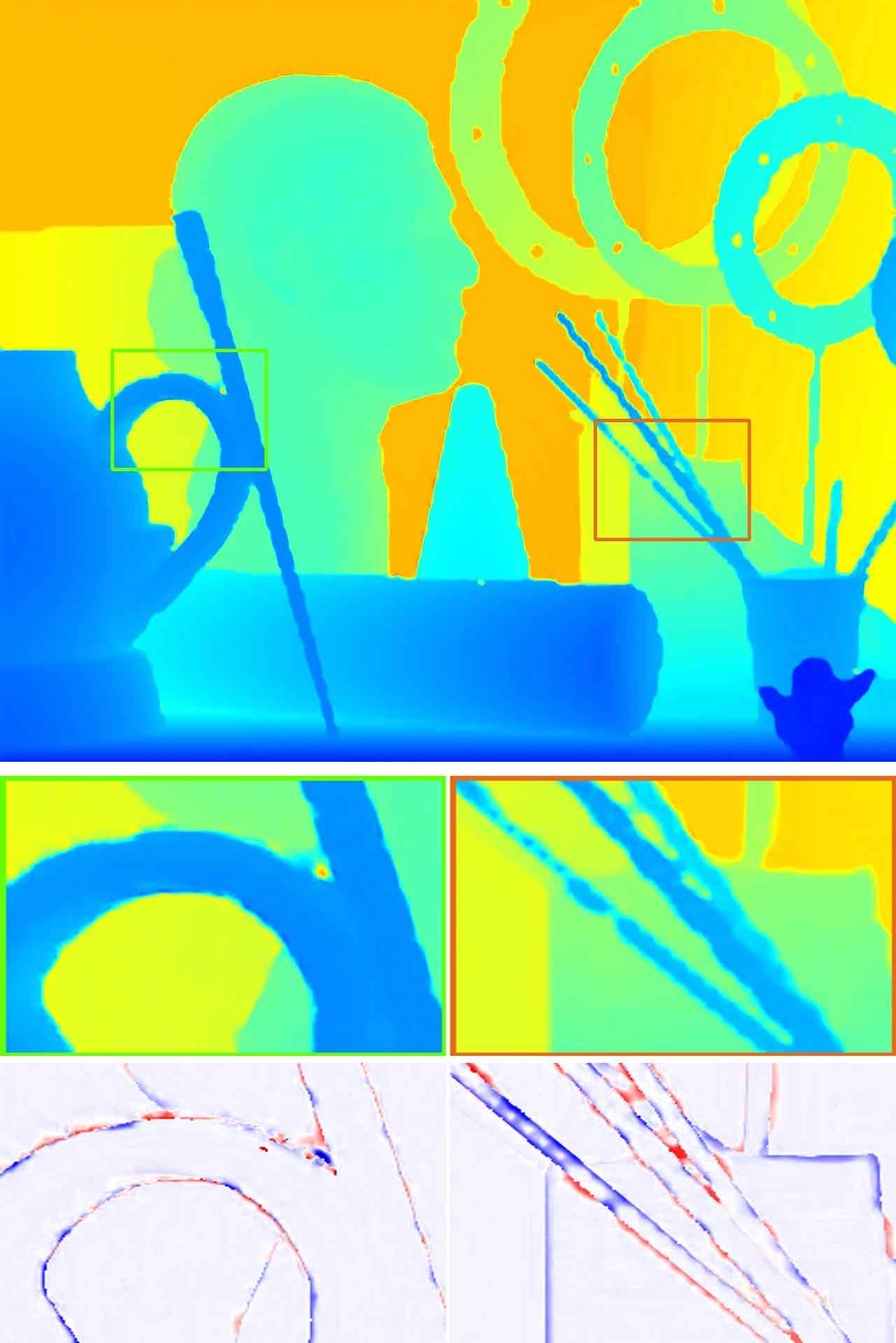} 
		\end{minipage}
		\hspace{-0.09in}
	}\subfigure[Model4]{
		\begin{minipage}[b]{0.16\linewidth}
			\includegraphics[width=1\linewidth]{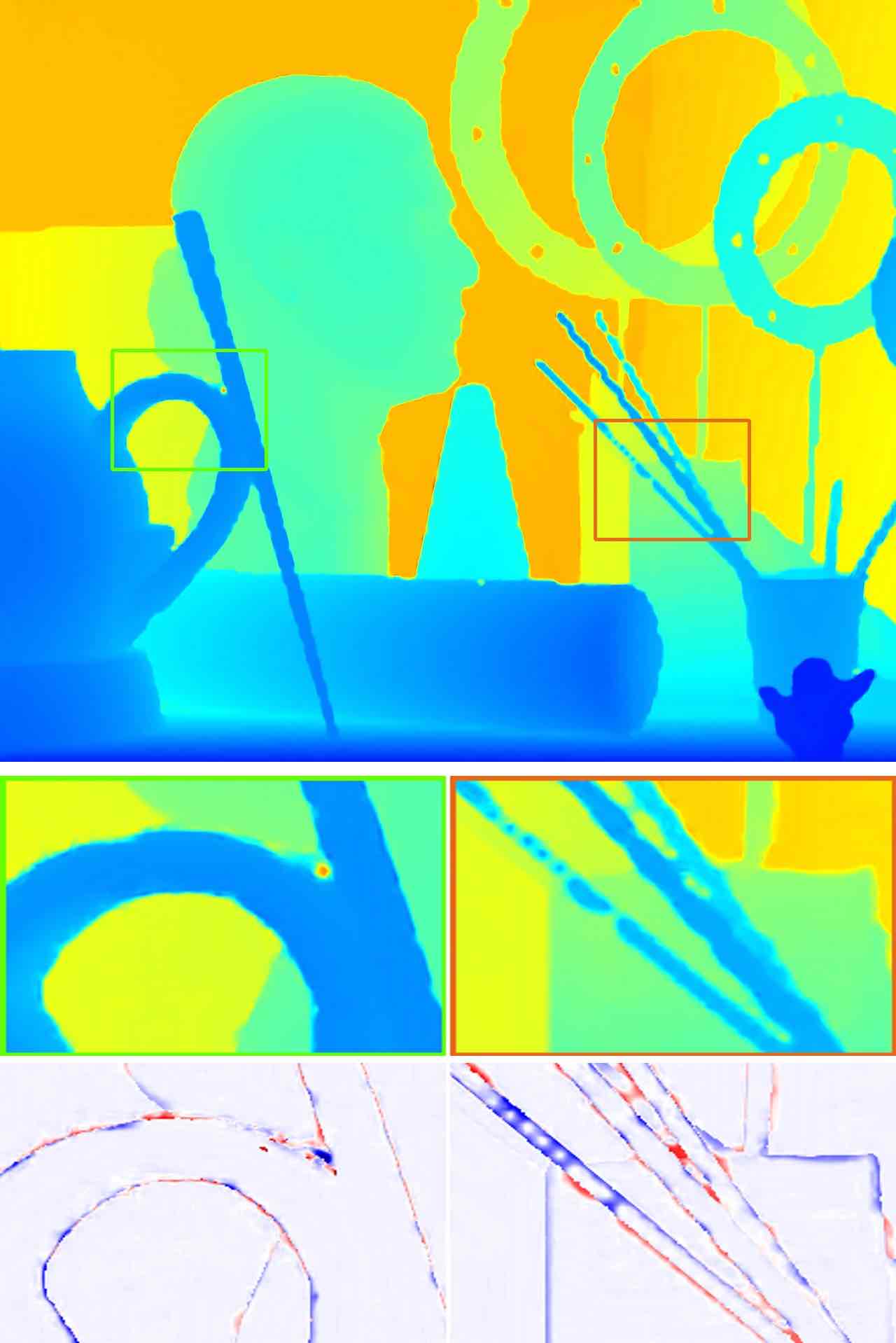} 
		\end{minipage}
		\hspace{-0.09in}
	}\subfigure[Model5]{
		\begin{minipage}[b]{0.16\linewidth}
			\includegraphics[width=1\linewidth]{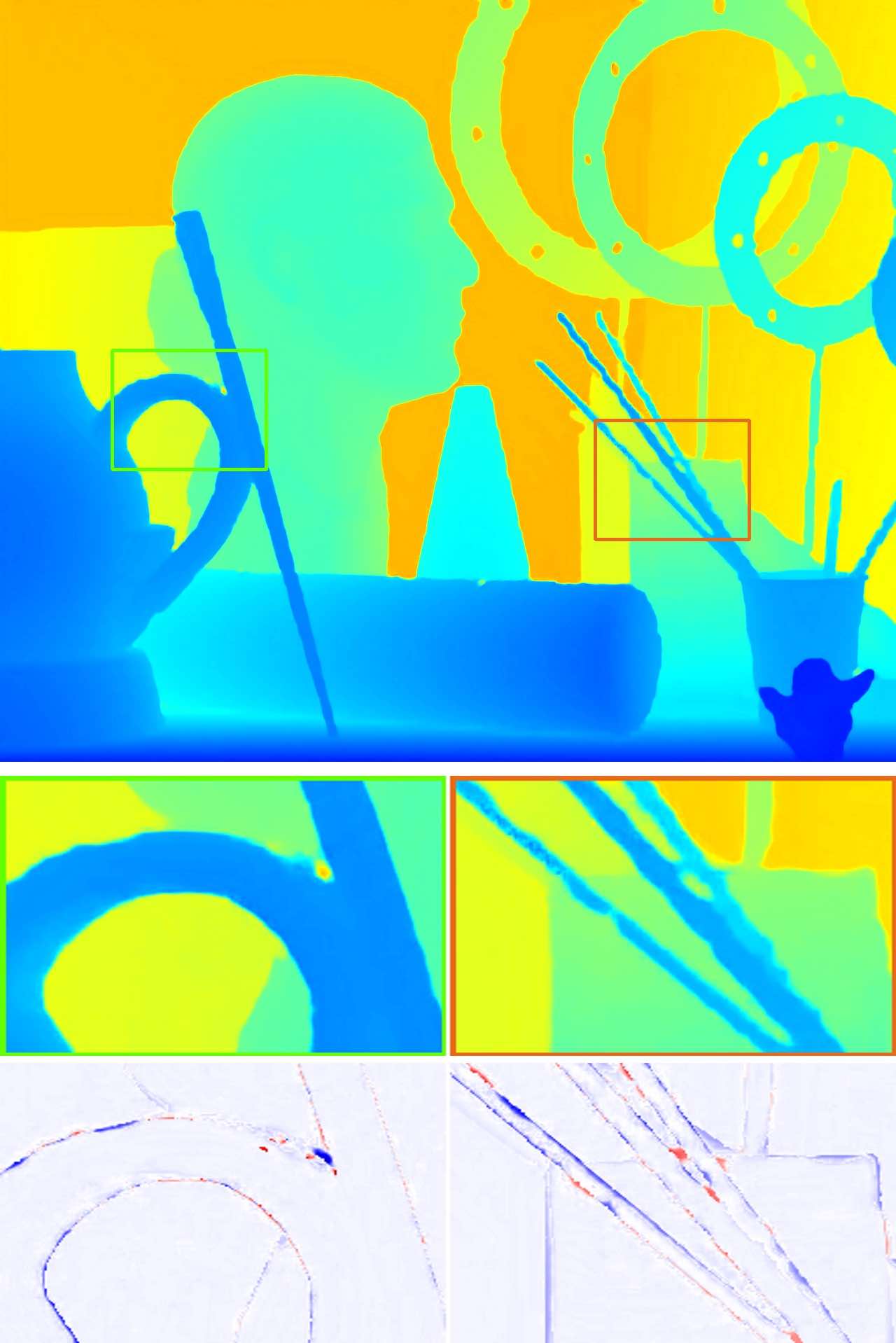} 
		\end{minipage}
	}
	\end{center}
% 		\vspace{-0.1in}
    \caption{\textbf{Ablation Study}. Visual comparison of an example without and with the proposed attentional kernel learning module (AKL) for depth image super-resolution. The first row shows the super-resolved depth images and the last row shows the error map ($\bm{I}^{\text{h}} - \bm{I}^{\text{out}}$). Please enlarge the PDF for more details.}
	\label{fig:abl_res}
\end{figure*}

In this section, we first present the hyper-parameters setting in our model, and then conduct a series of ablation experiments to investigate the effectiveness of our main contributions, \eg, attentional kernel learning module (mentioned in Sect.~\ref{akl}), multi-scale fusion (mentioned in Sect.~\ref{msf}) with deep supervision (mentioned in Sect.~\ref{lf}) and boundary-aware loss (mentioned in Sect.~\ref{lf}). In this study, we train different variants of our model on the commonly used NYU v2 dataset (Silberman et al., \cite{NYU}) with $16 \times$ nearest-neighbour downsampling and evaluate the performance of them on four benchmark datasets. The experimental settings are the same as Sect.~\ref{gsr}.

\subsection{Hyper-parameters setting}

For network hyper-parameters setting, we investigate the influence the size $k \times k$ of learned filter kernels in our kernel generation sub-network (\eg, $W_0, W_1, W_2$ in Fig.~\ref{fig:dag}) and the number of pyramid level $m$ in our model for multi-modality feature extraction to the final performance. Enlarging $k$ or $m$ can increase the receptive field of our model but at the expense of higher computational complexity. To  seek an appropriate trade-off between complexity and performance, we conduct experiments on the task of depth map super-resolution with different $k$ and $m$, and the results are summarized in Table~\ref{tab:tab_kernel}. From this table, we can see that the reconstruction performance is significantly improved when the number of pyramid levels $m$ increased from 1 to 3. However, when $m$ is too large, \eg, $m=4$, the improvements are small or even worse. We can draw the same conclusion for the size of filter kernels $k \times k$. The possible reason for this phenomenon is that the receptive filed is enough for this task when $m=3, k=3$ and larger $m$ or $k$ will burden the optimization process of network. Therefore, we set $m=3, k=3$ in our experiments.

\subsection{Ablation Experiments}
As shown in Fig. \ref{fig:dag}, our model consists of two part: kernel generation sub-network and multi-scale guided image filtering sub-network. For kernel generation sub-network, we propose to generate dual sets of kernels from the guidance and target images, and employ a tiny network to learn a weight map to adaptively combine the two sets of kernels. For guided image filtering sub-network, we progressively filter the target image with the learned multi-scale kernels. In order to fully integrate the intermediate filtered results, we propose a multi-scale feature fusion strategy and a multi-stage loss. To encourage our model to give more emphasis to the high-frequency and to generate visual pleasing results, we propose to train our model with hybird loss functions, \eg, pixel-wise loss $\mathcal{L}_1$, multi-scale loss $\mathcal{L}_{ms}$, and boundary-aware loss $\mathcal{L}_{ba}$. To analyze the contribution of each component of our model, we implement seven variants of our model:
\begin{itemize}
    \item \textbf{Model1}, which takes (target, target) as inputs for kernel generation, and is trained with $\mathcal{L}_1$ loss.
    \item \textbf{Model2}, which takes (guidance, guidance) as inputs for kernel generation, and is trained with $\mathcal{L}_1$ loss.
    \item \textbf{Model3}, which takes (target, guidance) as inputs for kernel generation, and uses element-wise multiplication to combine the generated two sets of kernels, and is trained with $\mathcal{L}_1$ loss.
    \item \textbf{Model4}, which takes (target, guidance) as inputs for kernel generation, and uses element-wise summation to combine the generated two sets of kernels,  and is trained with $\mathcal{L}_1$ loss.
    \item \textbf{Model5}, which takes (target, guidance) as inputs for kernel generation, and uses the learned weight map to adaptively combine the generated two sets of kernels, and is trained with $\mathcal{L}_1$ loss.
    \item \textbf{Model6}, which is Model5 but trained with $\mathcal{L}_1$ loss and $\mathcal{L}_{ms}$ loss.
    \item \textbf{Model7}: which is Model5 but trained with $\mathcal{L}_1$ loss, $\mathcal{L}_{ms}$ loss and $\mathcal{L}_{ba}$ loss. This is our full model.
\end{itemize}

It is noteworthy that we adjust the number of convolutional layers in multi-scale guided image filtering sub-network to guarantee that each variant could have roughly the same number of parameters with our final model. The quantitative results are shown in Table~\ref{abl}, from which we can see that the full model (Model7) achieves the best reconstruction performance across four testing datasets when compared with the ablated models, and every component proposed in our model can boost the network performance significantly. In the following, we will give a detailed analysis of each component in our method.
\begin{figure*}[!tb]
	\begin{center}
		\subfigure[Guidance]{
		\begin{minipage}[b]{0.195\linewidth}
			\includegraphics[width=1\linewidth]{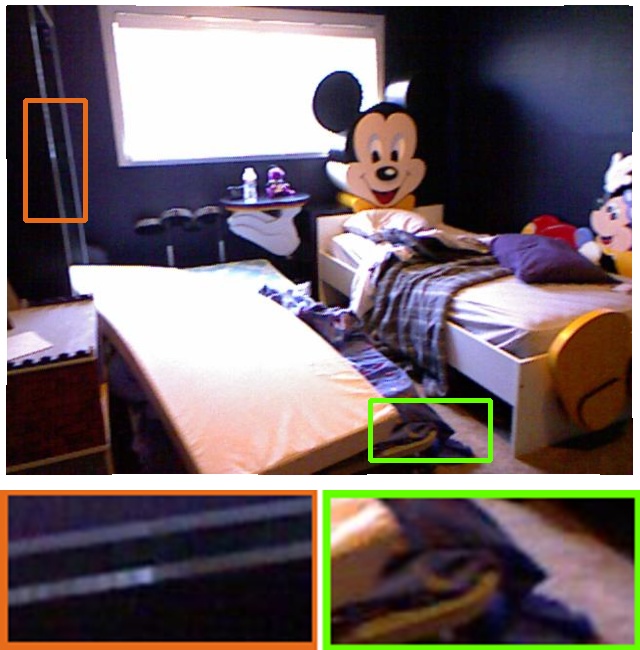} 
		\end{minipage}
		\hspace{-0.09in}
	}\subfigure[Target]{
    		\begin{minipage}[b]{0.195\linewidth}
  		 	\includegraphics[width=1\linewidth]{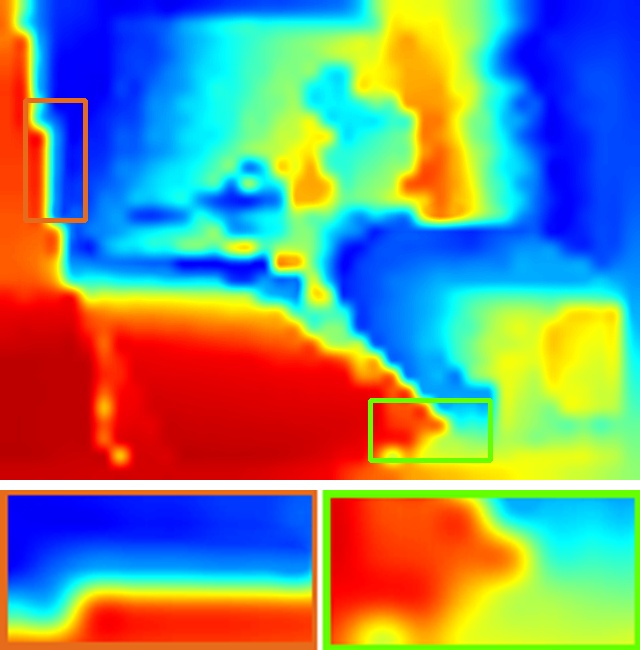}
    		\end{minipage}
    		\hspace{-0.09in}
    	}\subfigure[w/o $\mathcal{L}_{ba}$]{
		\begin{minipage}[b]{0.195\linewidth}
			\includegraphics[width=1\linewidth]{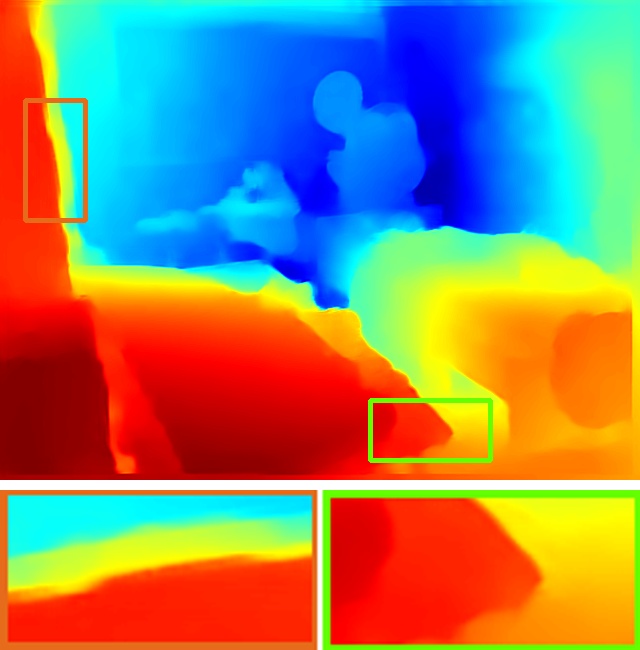} 
		\end{minipage}
		\hspace{-0.09in}
	}\subfigure[w/ $\mathcal{L}_{ba}$]{
		\begin{minipage}[b]{0.195\linewidth}
			\includegraphics[width=1\linewidth]{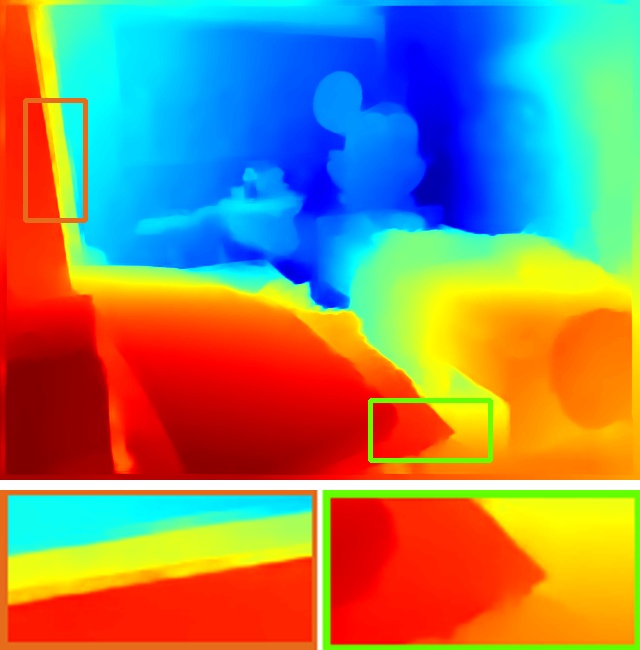} 
		\end{minipage}
		\hspace{-0.09in}
	}\subfigure[GT]{
		\begin{minipage}[b]{0.195\linewidth}
			\includegraphics[width=1\linewidth]{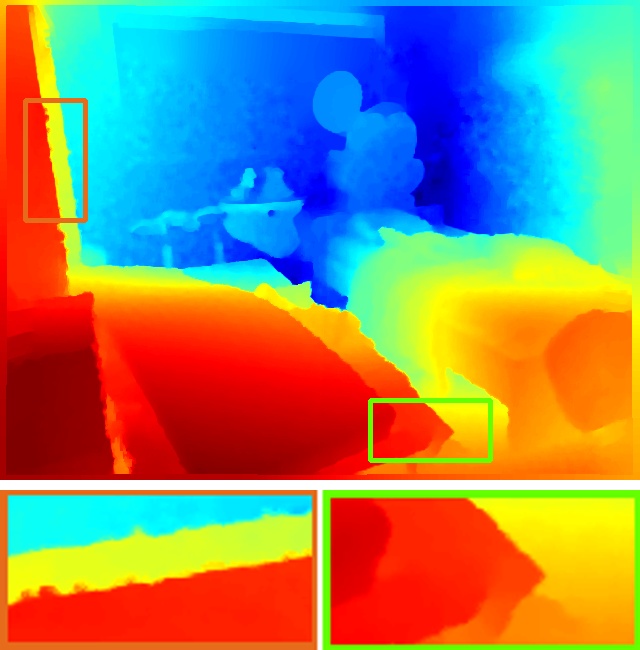} 
		\end{minipage}
	}
	\end{center}
	\vspace{-.1in}
	\caption{\textbf{Ablation Study}. Visual comparison of an example without and with the proposed boundary-aware loss for depth image super-resolution. Please enlarge the PDF for more details.}
	\label{fig:abl_baloss}
\end{figure*}
\begin{figure}[!tb]
	\begin{center}
		\subfigure[Guidance]{
		\begin{minipage}[b]{0.24\linewidth}
			\includegraphics[width=1\linewidth]{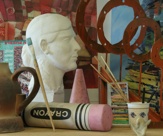} 
		\end{minipage}
		\hspace{-0.1in}
	}\subfigure[$\bm{A}_0$]{
    		\begin{minipage}[b]{0.24\linewidth}
  		 	\includegraphics[width=1\linewidth]{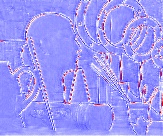}
    		\end{minipage}
    		\hspace{-0.1in}
    	}\subfigure[$\bm{A}_1$]{
		\begin{minipage}[b]{0.24\linewidth}
			\includegraphics[width=1\linewidth]{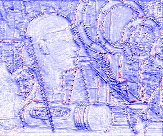} 
		\end{minipage}
		\hspace{-0.1in}
	}\subfigure[$\bm{A}_2$]{
		\begin{minipage}[b]{0.24\linewidth}
			\includegraphics[width=1\linewidth]{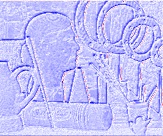} 
		\end{minipage}
	}
	\subfigure[Target]{
		\begin{minipage}[b]{0.24\linewidth}
			\includegraphics[width=1\linewidth]{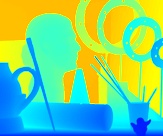} 
		\end{minipage}
		\hspace{-0.1in}
	}\subfigure[$\bm{1} - \bm{A}_0$]{
		\begin{minipage}[b]{0.24\linewidth}
			\includegraphics[width=1\linewidth]{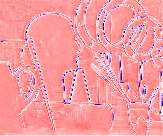} 
		\end{minipage}
		\hspace{-0.1in}
	}\subfigure[$\bm{1}- \bm{A}_1$]{
		\begin{minipage}[b]{0.24\linewidth}
			\includegraphics[width=1\linewidth]{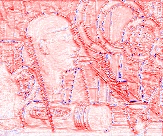} 
		\end{minipage}
		\hspace{-0.1in}
	}\subfigure[$\bm{1} - \bm{A}_2$]{
		\begin{minipage}[b]{0.24\linewidth}
			\includegraphics[width=1\linewidth]{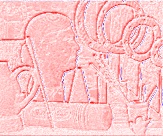} 
		\end{minipage}
	}
    \includegraphics[width=0.97\linewidth]{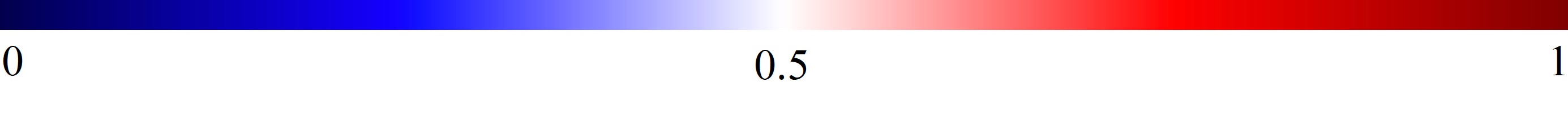} 
	\end{center}
% 	\vspace{-0.2in}
	 \caption{\textbf{Ablation Study}. Visualization of the learned multi-scale attention maps for kernel combination. We resize the attention map to the same size for better Visualization. Please enlarge the PDF for more details.}
	\label{fig:abl_map}
\end{figure}
\textbf{Effectiveness of Attentional Kernel Learning (AKL)}: In this paper, we propose to use AKL to generate filter kernels for guided image filtering. Specifically, it first generates dual sets of kernels by using the extracted guidance and target features respectively, and then adaptively combines the generated kernels by the learned attention maps. To demonstrate the effectiveness of AKL, we implement several variants (\eg, different inputs for kernel construction and different kernel fusion strategies) of the proposed method, including Model1--Model5. The quantitative results on the four testing datasets are reported in Table~\ref{tab_abl}. As can be seen from this table, Model1 generates kernels from target image only, thus the reconstruction accuracy is relatively low. With the assistance of guidance image, Model2 obtains a significant improvement compared with Model1, which implies that the guidance information is helpful for filter kernel generation. However, the guidance images are not always reliable, such as color images captured in bad weather or low-light conditions. In view of this, Model3 and Model4 generate dual sets of kernels from the guidance and target images, respectively, and the difference between the two models is the strategy of kernel combination. As shown in Table~\ref{tab_abl}, Model3 and Model4 can further improve the  accuracy over Model2 (The average RMSE is dropped from 8.45 to 8.28 and 8.23), which indicates that constructing kernels from both target and guidance images enjoys some benifits over using only the guidance. Nevertheless, using element-wise multiplication or summation to combine the generated kernels would limit the capacity of the network, since they ignore the inconsistency between guidance and target images. To solve this problem, we first learn an attention map, and then utilize the attention map to selectively combine the dual kernels as in Eq.~\ref{eq1}. As depicted in Table~\ref{tab_abl}, equipped with AKL, compared with Model3, Model5 reduces the average RMSE from 8.32 to 7.82.

To visually show the effect of AKL, we present in Fig.~\ref{fig:abl_res} the super-resolved depth images (first row) and error maps (last row) with different configurations. The error map is obtained by $\bm{I}^\text{h} - \bm{I}^{\text{out}}$. As shown in Fig.~\ref{fig:abl_res}, the result of Model1 is blur and lack of high-frequency details. For the error map of Model1, most of the values at the image boundaries are positive, which means that the boundaries generated by Model1 are weaker than the ones of ground truth. The reason is that the kernels generated from the target image only cannot produce the high-frequency details which are lost by the image degradation process. On the contrary, most values in the error map of Model2 are negative, although the depth boundaries are enhanced, the texture-copying artifacts seriously influence the super-resolved depth maps. Thanks to the proposed attentional kernel learning theme that constructs kernels by fully integrating complementary information contained in both guidance and target images, the visual effect and reconstruction accuracy of Model5 are substantially improved.

Moreover, we visualize the attention map in Fig.~\ref{fig:abl_map} to further validate the capability of the proposed AKL, from which we can see that the kernels generated from the target and guidance images are both important for the task of guided filtering as most of the pixel values in the attention maps are in the range of [0.4, 0.6]. In addition, as shown in the first row of Fig.~\ref{fig:abl_map}, the structure regions are lighter than texture regions, and this indicates that our model can adaptively select relevant information from the guidance image while avoiding texture over-transfer issues.

\begin{figure}[!tb]
	\begin{center}
		\subfigure[Train]{
		\begin{minipage}[b]{0.47\linewidth}
			\includegraphics[width=1\linewidth]{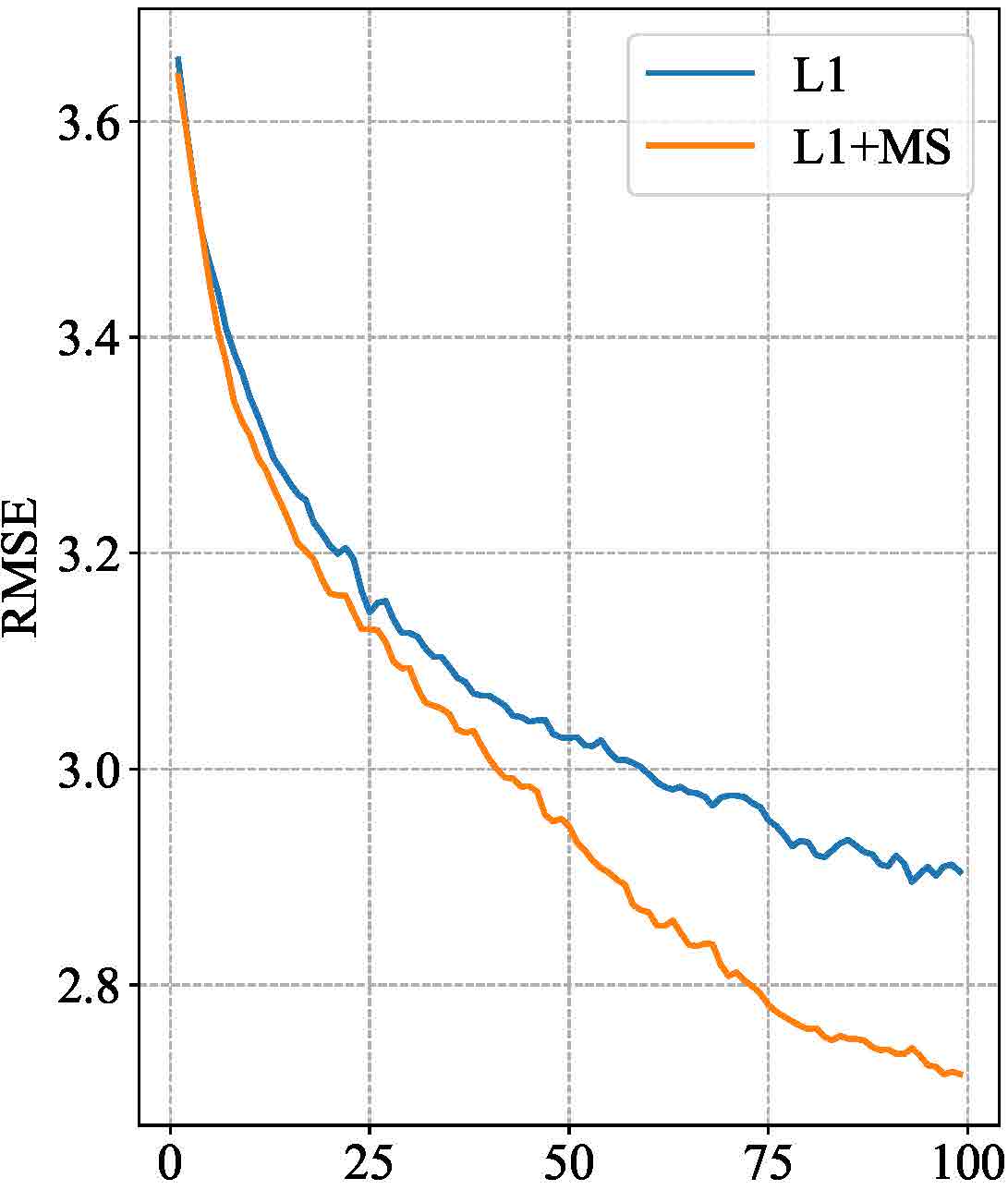} 
		\end{minipage}
		\hspace{-0.1in}
	}\subfigure[Test]{
    		\begin{minipage}[b]{0.458\linewidth}
  		 	\includegraphics[width=1\linewidth]{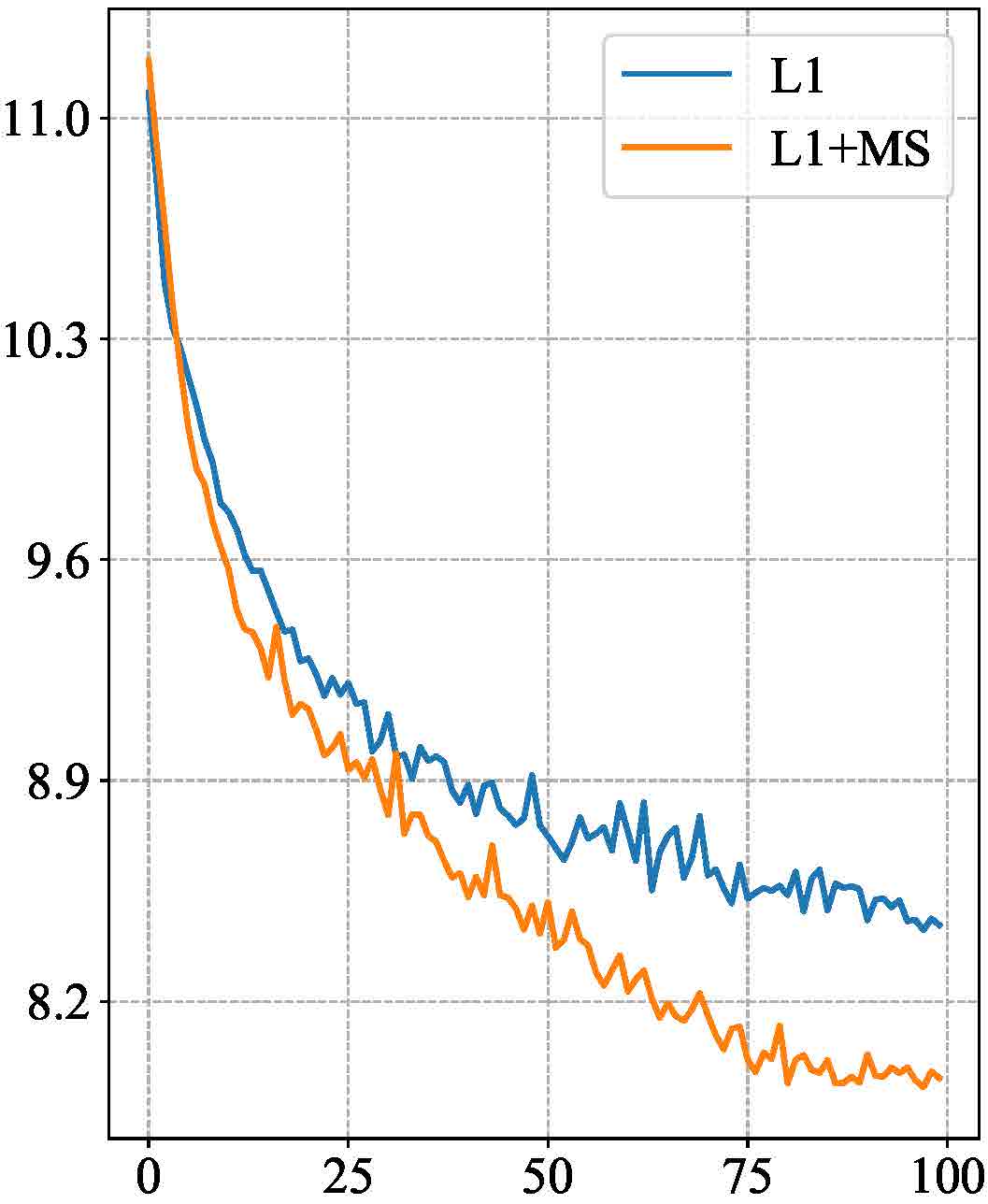}
    		\end{minipage}
    	}
	\end{center}
% 	\vspace{-.2in}
	\caption{\textbf{Ablation Study}. Training and testing RMSE values on NYU v2 dataset~(Silberman et al., \cite{NYU}) for $16\times$ depth image super-resolution. MS denotes the proposed multi-stage loss $\mathcal{L}_{ms}$.}
	\label{fig:ecl}
\end{figure}

\textbf{Effectiveness of Multi-scale Fusion and Deep Supervision}: In this paper, we propose a multi-scale framework for guided image filtering. Specifically, in order to obtain both high-level structure information and low-level details, we propose to fuse multi-level filtered outputs. Moreover, a multi-stage loss is introduced to enforce the intermediate results to be close to the ground-truth target image. The quantitative results are illustrated in Table~\ref{tab_abl}. As expected, Model6 trained with a hybrid loss of $\mathcal{L}_1$ and $\mathcal{L}_{ms}$ further improves the reconstruction accuracy. Fig.~\ref{fig:ecl} further shows the train (left) and test (right) RMSE plots. We observe that the multi-stage loss ($\mathcal{L}_{ms}$) is able to accelerate convergence velocity and produce results with lower RMSE values.

\textbf{Effectiveness of Boundary-aware Loss}: To encourage the network to pay more attention to high-frequency information, we propose to train our model with boundary-aware loss ($\mathcal{L}_{ba}$). Table~\ref{tab_abl} demonstrates that $\mathcal{L}_{ba}$ loss is helpful in improving the reconstruction accuracy (Model7). Fig.~\ref{fig:abl_baloss} presents an example visual comparison with and without the $\mathcal{L}_{ba}$ loss. Obviously, $\mathcal{L}_{ba}$ improves the visual quality further, yielding more precise edges. The boundaries on the doorframe and corner of the mattress are sharper and clearer, which verifies the effectiveness of the proposed boundary-aware loss.

\begin{figure}[!tb]
	\begin{center}
		\subfigure[~]{
		\begin{minipage}[b]{0.5\linewidth}
			\includegraphics[width=0.85\linewidth]{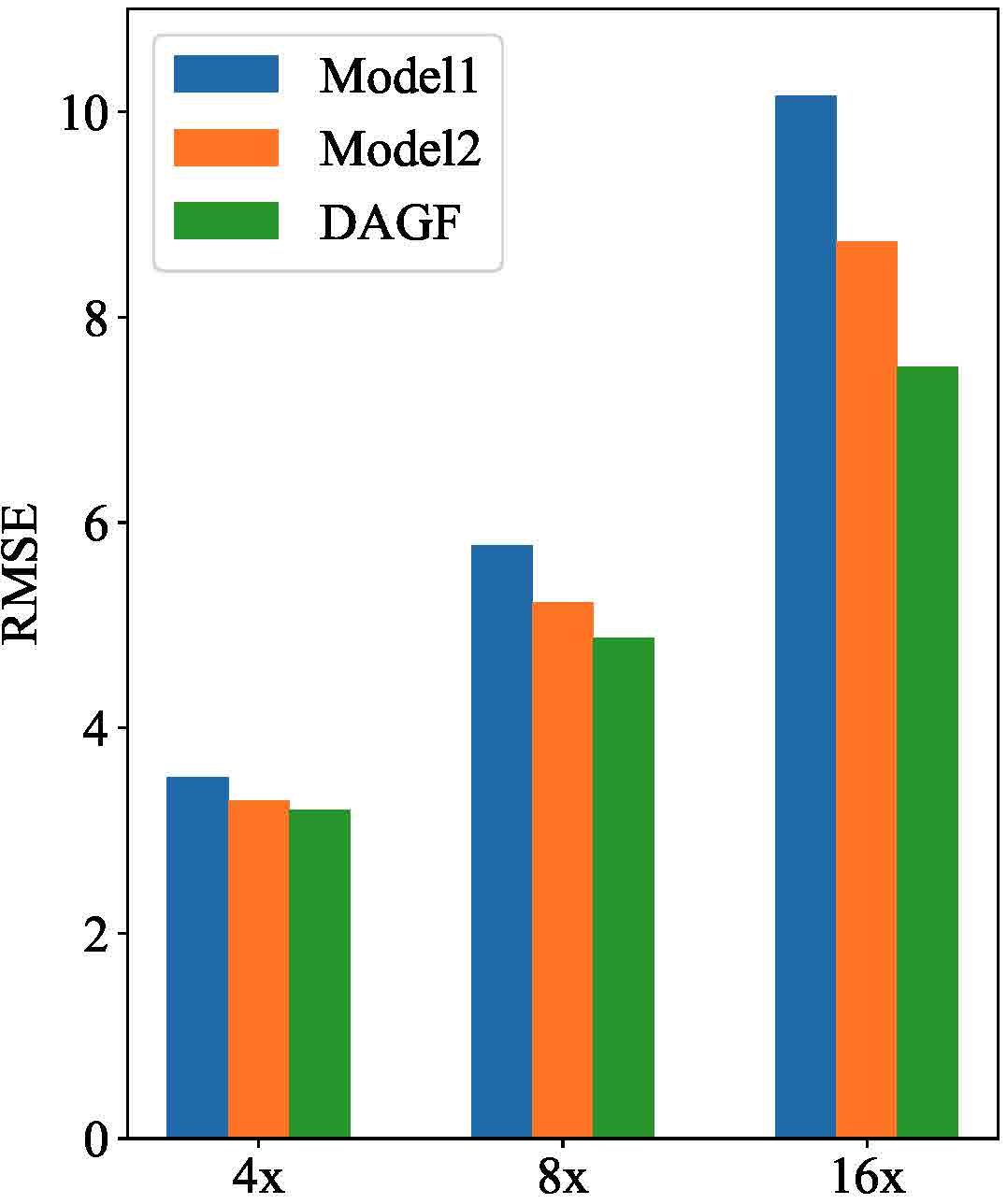} 
		\end{minipage}
		\hspace{-0.1in}
	}\subfigure[~]{
    		\begin{minipage}[b]{0.465\linewidth}
  		 	\includegraphics[width=0.85\linewidth]{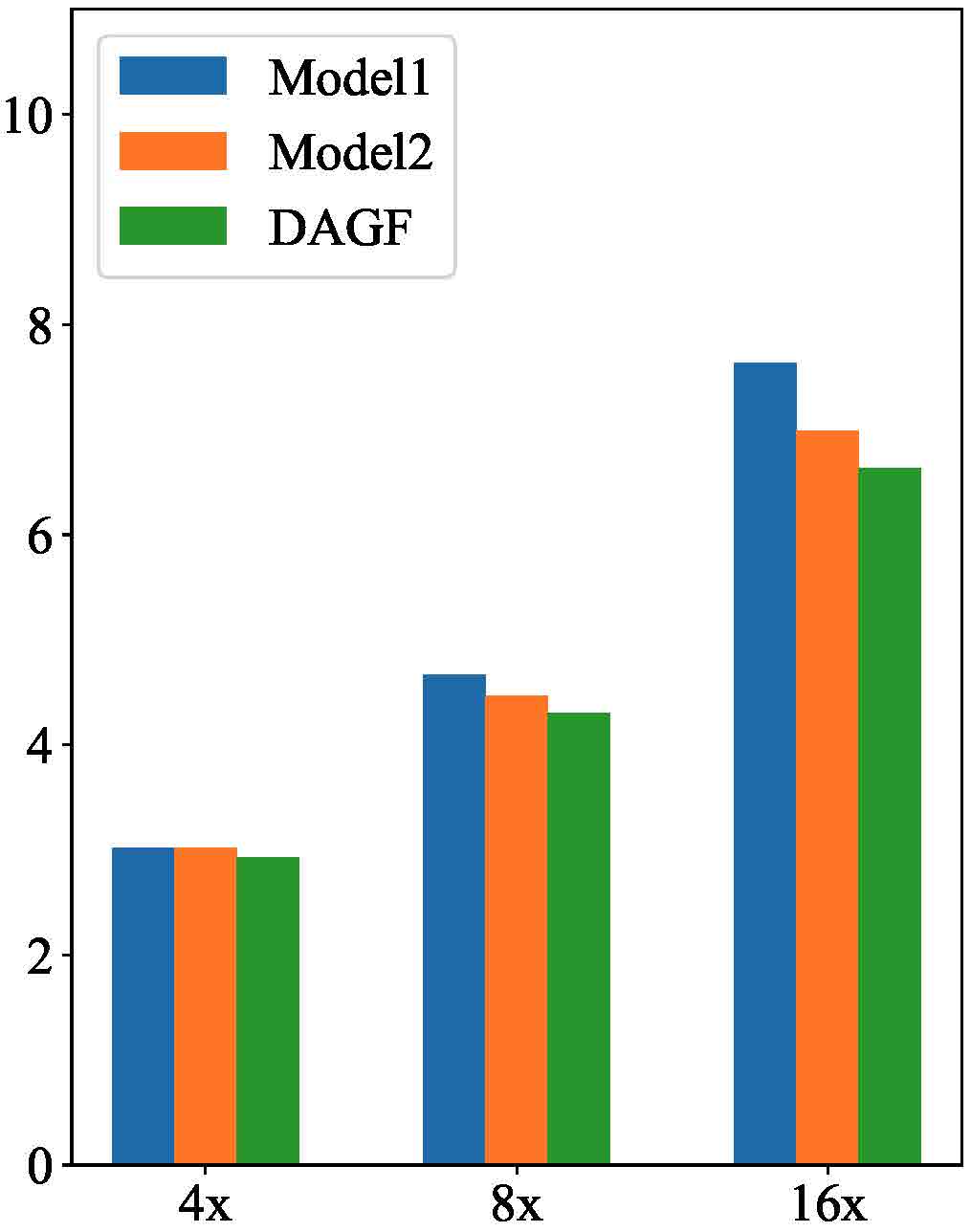}
    		\end{minipage}
    	}
	\end{center}
% 	\vspace{-.2in}
	\caption{\textbf{Ablation Study}. Average RMSE values for depth image super-resolution. The low-resolution depth image are obtained by (a): Nearest-neighbour downsampling, (b): Bicubic downsampling and Gaussian noise.}
	\label{fig:cd}
\end{figure}
\textbf{Effectiveness of Guidance Branch}: The general principle of guide image filtering is that we can transfer the valuable structure information contained in guidance image to the target image. Recently, various approaches have been proposed for guided image filtering. Nevertheless, most of them focus on designing advanced algorithm for efficiently transferring structures from the guidance to the target image, and the contributions of guidance images under different conditions are rarely explored. Here, we conduct experiments on several applications of guided image super-resolution, \eg, depth image super-resolution (nearest-neighbour downsampling) and noisy depth super-resolution (Bicubic downsampling and Gaussian noise). As shown in Fig.~\ref{fig:cd}, we evaluate the role of guidance image and compute the average RMSE value for each upsampling factor. Model1 takes the target image as input for kernel generation, while Model2 takes the guidance images as input for kernel generation. The results show that the guidance image can provide significant assistance for the $8 \times$ and $16 \times$ cases, and the model (DAGF) equipped with the proposed AKL can further improve the performance. However, for the  $4\times$ case, which is easy to recover, the guidance information has a negligible effect. The main reason is that the target image is not severely damaged by downsampling degradation, therefore, the target image can be easily recovered by Model1. For the more difficult cases ($8\times$ and $16\times$), the target image is badly polluted, the guidance image would play an important role in the reconstruction process. 

\begin{figure}[!t]
    \centering
    \includegraphics[width=\linewidth]{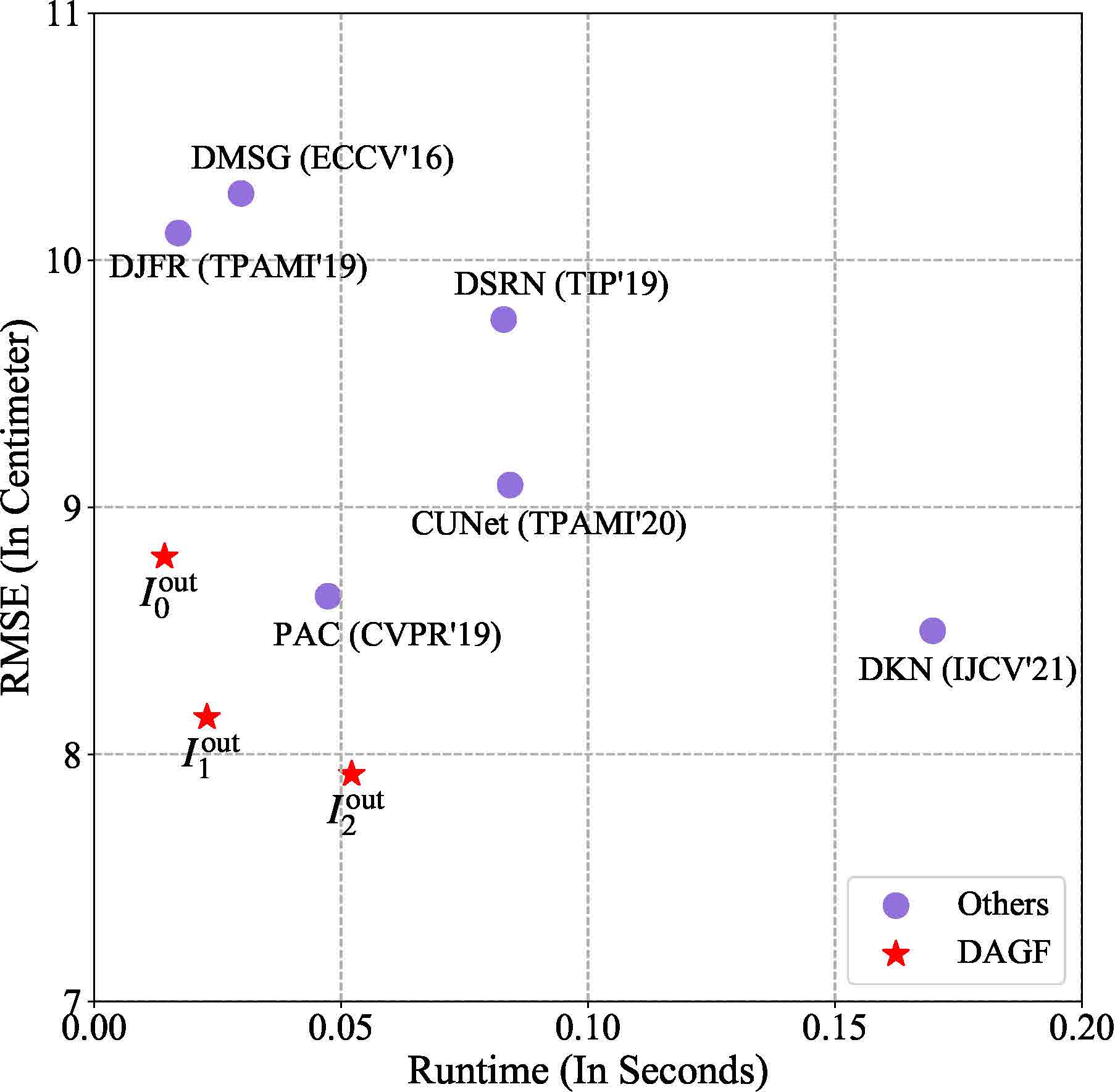}
    \caption{Average runtime (in seconds) and root mean square error (RMSE) comparison for $16 \times$ depth image super-resolution on NYU v2 dataset~\cite{NYU}. All the runtimes are evaluated on the same NVIDIA 1080Ti GPU with depth image size $480 \times 640$.}
    \label{fig:time_rmse}
\end{figure}

\textbf{Performance vs. Complexity Analysis}: In Fig.~\ref{fig:time_rmse}, we compare the running time among our method and other comparison methods on NYU v2 (Silberman et al., \cite{NYU}) for $16\times$ depth image super-resolution. For fair comparison, all the running times are obtained on the same machine by one NVIDIA 1080Ti GPU. As shown in Fig.~\ref{fig:dag}, our method produces multiple results $I_0^{\text{out}}, I_1^{\text{out}}, I_2^{\text{out}}$, and we first resize them to the same resolution as the ground truth target image by a simple bilinear interpolation method, and then calculate the RMSE values. As illustratedd in Fig.~\ref{fig:time_rmse}, the final result $I_2^{\text{out}}$ achieves the best RMSE result than DKN (Kim et al., \cite{DKN}) and DSRN (Guo et al., \cite{DepthSR}) but needs less time. The time cost for $I_0^{\text{out}}$ is the least, and the performance of $I_0^{\text{out}}$ is comparable to other methods. If the purpose is to achieve the performance as best as possible, we can increase the level of pyramid, otherwise reduce the level of pyramid. Overall, our method can achieve a better trade-off between the reconstruction performance and computational complexity.

\section{Conclusion}
\label{con}
In this paper, we present an effective network architecture for guided image filtering, which can automatically select and transfer important structures from the guidance to the target image. Specifically, an attentional kernel learning module (AKL) is proposed to generate dual sets of filter kernels from the guidance and target images, respectively, and then adaptively combine the learned kernels in a learning manner. Furthermore, a multi-scale guided image filtering framework is introduced, which takes the generated kernels and target image as inputs and progressively filters the target image in a coarse-to-fine manner. Moreover, to fully explore the intermediate results in the coarse-to-fine process, we propose a multi-scale fusion with deep supervision to regularize and combine multiple filtering results. Finally, boundary-aware loss is introduced to enhance the high-frequency details of guided filtering. Experimental results on various guided image filtering applications show the superiority and flexibility of the proposed model and the ablation experiments demonstrate the effectiveness of each component in our method.

\ifCLASSOPTIONcompsoc
  \section*{Acknowledgments}
\else
  \section*{Acknowledgment}
\fi


\ifCLASSOPTIONcaptionsoff
  \newpage
\fi



%\bibliographystyle{IEEEtran}


% \end{thebibliography}

\bibliographystyle{IEEEtran}
\bibliography{Mendeley}

 \begin{IEEEbiography}[{\includegraphics[width=1.0in,height=1.25in,clip,keepaspectratio]{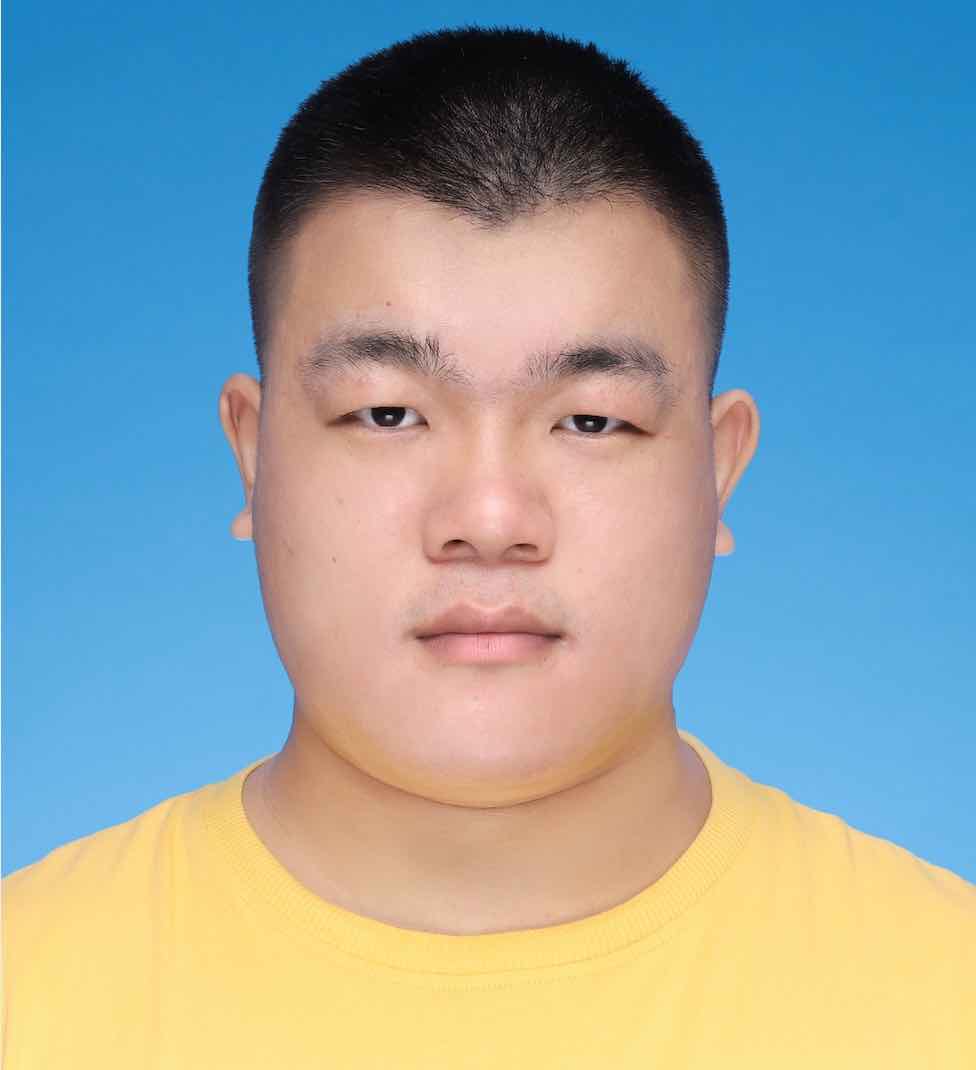}}]{Zhiwei Zhong}
received the B.S. degree in computer science from the Heilongjiang University, Harbin, China, in 2017. He is currently pursing the Ph.D. degree in computer science from the Harbin Institute of Technology (HIT), Harbin, China. His research interests include image processing, computer vision and deep learning.

 \end{IEEEbiography}

\begin{IEEEbiography}[{\includegraphics[width=1.0in,height=1.25in,clip,keepaspectratio]{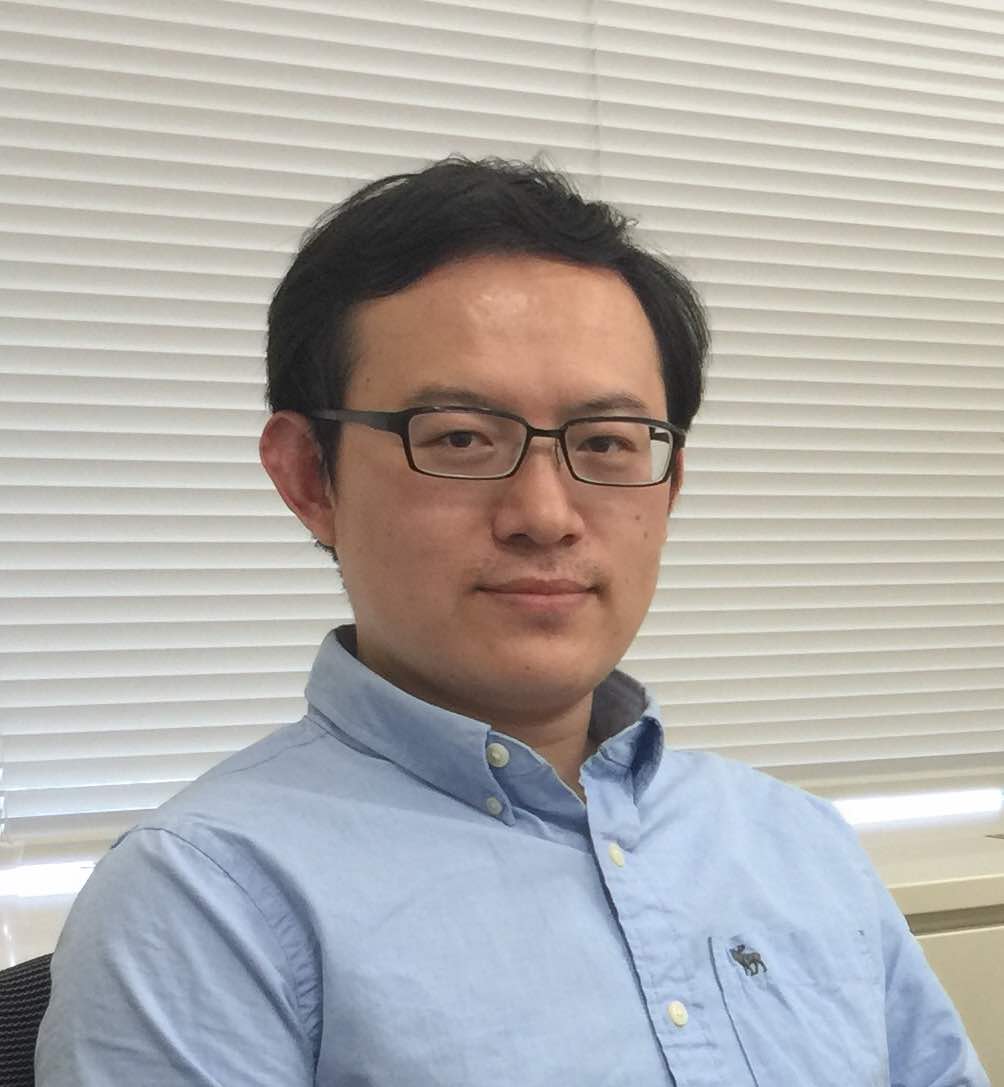}}]{Xianming Liu}
received the B.S., M.S., and Ph.D. degrees in computer science from the Harbin
Institute of Technology (HIT), Harbin, China,
in 2006, 2008, and 2012, respectively. In 2011, he
spent half a year at the Department of Electrical
and Computer Engineering, McMaster University,
Canada, as a Visiting Student, where he was a
Post-Doctoral Fellow from 2012 to 2013. He was
a Project Researcher with the National Institute of
Informatics (NII), Tokyo, Japan, from 2014 to 2017.
He is currently a Professor with the School of Computer
Science and Technology, HIT. He has published over 50 international
conference and journal publications, including top IEEE journals, such as T-IP,
T-CSVT, T-IFS, and T-MM, and top conferences, such as ICML, CVPR, IJCAI. He was a receipt of the IEEE ICME 2016 Best Student Paper Award.
\end{IEEEbiography}

 \begin{IEEEbiography}[{\includegraphics[width=1.0in,height=1.25in,clip,keepaspectratio]{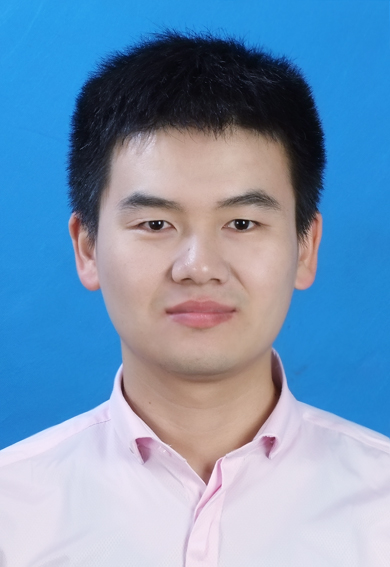}}]{Junjun Jiang}
received the B.S. degree from the Department of Mathematics, Huaqiao University, Quanzhou, China, in 2009, and the Ph.D. degree from the School of Computer, Wuhan University, Wuhan, China, in 2014.
From 2015 to 2018, he was an Associate Professor at China University of Geosciences, Wuhan. Since 2016, he has been a Project Researcher with the National Institute of Informatics, Tokyo, Japan. He is currently a Professor with the School of Computer Science and Technology, Harbin Institute of Technology, Harbin, China. He won the Finalist of the World's FIRST 10K Best Paper Award at ICME 2017, and the Best Student Paper Runner-up Award at MMM 2015. He received the 2016 China Computer Federation (CCF) Outstanding Doctoral Dissertation Award and 2015 ACM Wuhan Doctoral Dissertation Award. His research interests include image processing and computer vision.

 \end{IEEEbiography}
 
 \begin{IEEEbiography}[{\includegraphics[width=1in,height=1.25in,clip,keepaspectratio]{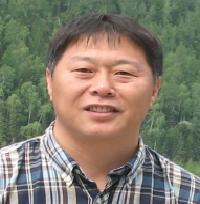}}]{Debin Zhao}
	received the B.S., M.S., and Ph.D. degrees in computer science from Harbin Institute of Technology, China in 1985, 1988, and 1998, respectively. He is now a professor in the Department of Computer Science, Harbin Institute of Technology. He has published over 200 technical articles in refereed journals and conference proceedings in the areas of image and video coding, video processing, video streaming and transmission, and pattern recognition.
\end{IEEEbiography}

\begin{IEEEbiography}[{\includegraphics[width=1.0in,height=1.25in,clip,keepaspectratio]{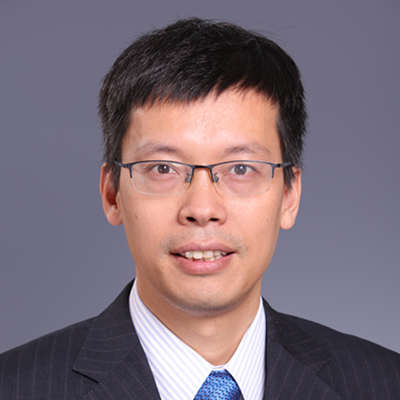}}]{Xiangyang Ji}
received the B.S. degree in materials science and the M.S. degree in computer science from the Harbin Institute of Technology, Harbin, China, in 1999 and 2001, respectively, and the Ph.D. degree in computer science from the Institute of Computing Technology, Chinese Academy of Sciences, Beijing, China. He joined Tsinghua University, Beijing, in 2008, where he is currently a Professor with the Department of Automation, School of Information Science and Technology. He has authored over 100 referred conference and journal papers. His current research interests include signal processing, image/video compressing, and intelligent imaging.

\end{IEEEbiography}


% \end{IEEEbiography}


% % insert where needed to balance the two columns on the last page with
% %\newpage



%\vfill


\end{document}